\documentclass[10pt,twocolumn,a4paper]{article}
\usepackage[cm]{raidium}     %

\usepackage{placeins}

\usepackage{xspace}
\usepackage{placeins}
\newcommand{\Jolia}{Jolia\xspace}
\newcommand{\JoliaTwo}{Jolia 2\xspace}   %
\newcommand{\Conquer}{ConQuer\xspace}  %
\newcommand{\Method}{\mbox{MaTIS}\xspace}      %
\newcommand{\ours}{\JoliaTwo}   %

\newcommand{\Merlin}{Merlin\xspace}
\newcommand{\PillarZero}{Pillar-0\xspace}
\newcommand{\CAPTOR}{CAPTOR\xspace}

\newcommand{\FlexiCT}{FlexiCT-VLM\xspace}
\newcommand{\COLIPRI}{COLIPRI\xspace}

\newcommand{\RadFinder}{RadFinder\xspace}

\newcommand{\dts}{patch-level segmentation\xspace}
\newcommand{\osl}{OSL\xspace}
\newcommand{\HieraCT}{Hiera-CT\xspace}
\newcommand{\SSL}{SSL\xspace}
\newcommand{\Coverage}{query grounding\xspace}
\newcommand{\OneMM}{1\,mm\xspace}

\newcommand{\CLIP}{CLIP\xspace}
\newcommand{\MAE}{MAE\xspace}
\newcommand{\CuriaMAE}{Curia-MAE\xspace}   %
\newcommand{\MAEBT}{MAE + BT\xspace}   %
\newcommand{\IJEPA}{I-JEPA\xspace}
\newcommand{\iBOT}{iBOT\xspace}
\newcommand{\DINOvTwo}{DINOv2\xspace}
\newcommand{\Franca}{Franca\xspace}
\newcommand{\VoCo}{VoCo\xspace}
\newcommand{\DenseBT}{DenseBT\xspace}
\newcommand{\ZSshort}{ZS short\xspace}
\newcommand{\ZSlong}{ZS long\xspace}
\newcommand{\Retrieval}{Retrieval\xspace}
\newcommand{\stanford}{Stanford\xspace}
\newcommand{\extabdo}{Ext-Abd\xspace}
\newcommand{\extchest}{Ext-Chest\xspace}
\newcommand{\ctrate}{CT-RATE\xspace}
\newcommand{\inspect}{INSPECT\xspace}
\newcommand{\Findings}{Findings\xspace}
\newcommand{\Seg}{Seg.\xspace}

\usepackage{array}
\usepackage{tikz}
\usetikzlibrary{arrows.meta}

\DeclareFontFamily{OT1}{jlm}{}
\DeclareFontShape{OT1}{jlm}{m}{n}{<-> rm-lmr10}{}
\DeclareFontShape{OT1}{jlm}{m}{it}{<-> rm-lmri10}{}
\DeclareFontShape{OT1}{jlm}{bx}{n}{<-> rm-lmbx10}{}
\DeclareFontShape{OT1}{jlm}{bx}{it}{<-> rm-lmbxi10}{}
\DeclareFontShape{OT1}{jlm}{b}{n}{<->ssub * jlm/bx/n}{}
\DeclareFontShape{OT1}{jlm}{b}{it}{<->ssub * jlm/bx/it}{}
\DeclareFontFamily{OML}{jcmm}{\skewchar\font'177}
\DeclareFontShape{OML}{jcmm}{m}{it}{<-> cmmi10}{}
\DeclareFontShape{OML}{jcmm}{b}{it}{<-> cmmib10}{}
\DeclareFontFamily{OMS}{jcmsy}{\skewchar\font'60}
\DeclareFontShape{OMS}{jcmsy}{m}{n}{<-> cmsy10}{}
\DeclareFontFamily{OMX}{jcmex}{}
\DeclareFontShape{OMX}{jcmex}{m}{n}{<-> cmex10}{}
\DeclareMathVersion{jfig}
\SetSymbolFont{operators}{jfig}{OT1}{jlm}{m}{n}
\SetSymbolFont{letters}{jfig}{OML}{jcmm}{m}{it}
\SetSymbolFont{symbols}{jfig}{OMS}{jcmsy}{m}{n}
\SetSymbolFont{largesymbols}{jfig}{OMX}{jcmex}{m}{n}

\definecolor{orgA}{HTML}{F2C14E}
\definecolor{orgB}{HTML}{5B8BD0}
\definecolor{orgC}{HTML}{892408}
\definecolor{pBlock}{HTML}{FBE9DC}\definecolor{pBlockLine}{HTML}{E3A57A}\definecolor{pPanel}{HTML}{FDF5EE}
\definecolor{pA}{HTML}{FFC20A}\definecolor{pB}{HTML}{4A90E2}\definecolor{pC}{HTML}{B3001B}
\definecolor{pLoss}{HTML}{FFFFFF}\definecolor{pBandA}{HTML}{EAF1F8}\definecolor{pBandALine}{HTML}{9DB7D1}
\definecolor{pBandV}{HTML}{EFEBF6}\definecolor{pBandVLine}{HTML}{A99BCB}\definecolor{pBandC}{HTML}{EAF3EC}
\definecolor{pBandCLine}{HTML}{93B8A1}\definecolor{pBandB}{HTML}{EDF0F4}\definecolor{pBandBLine}{HTML}{A7B3C4}
\definecolor{jLine}{HTML}{3D4352}
\definecolor{jNote}{HTML}{5F6675}
\definecolor{jSSL}{HTML}{2E7D5B}  %
\colorlet{jFill}{orgC!9!white}          %
\colorlet{jBlockLine}{orgC!55!white}
\colorlet{jCab}{orgA!12!white}          %
\colorlet{jCabLine}{orgA!75!white}
\colorlet{jCardSem}{orgB!8!white}\colorlet{jCardSemLine}{orgB!45!white}
\colorlet{jCardSpa}{orgA!12!white}\colorlet{jCardSpaLine}{orgA!70!white}
\colorlet{jCardSsl}{jSSL!13!white}\colorlet{jCardSslLine}{jSSL!50!white}
\colorlet{jCardDen}{orgC!8!white}\colorlet{jCardDenLine}{orgC!35!white}
\definecolor{jPanel}{HTML}{F7F8FA}
\definecolor{jPanelLine}{HTML}{B9BEC8}
\definecolor{jVoxEdge}{HTML}{8A90A0}
\definecolor{jVoxTop}{HTML}{ECEEF2}
\definecolor{jVoxSide}{HTML}{D2D6DD}
\definecolor{jFrozen}{HTML}{6F9FD8}
\definecolor{jCLS}{HTML}{7A4E9E}  %
\definecolor{jSc2}{HTML}{6FC1D4}\definecolor{jSc1}{HTML}{2C8FA6}\definecolor{jSc0}{HTML}{0B5A6B}  %

\tikzset{
  every picture/.append style={execute at begin picture={\fontencoding{OT1}\fontfamily{jlm}\selectfont\mathversion{jfig}}},
  jflow/.style={-{Stealth[length=2.4pt,width=2.2pt]},line width=0.5pt,draw=jLine},
  jlink/.style={-{Stealth[length=2.4pt,width=2.2pt]},line width=0.6pt},
  jnote/.style={font=\scriptsize,text=jNote},
  jblock/.style={draw=jBlockLine,fill=jFill,line width=0.6pt,rounded corners=2.5pt},
  jcab/.style={fill=jCab,draw=jCabLine,line width=0.6pt,rounded corners=3pt},
  jloss/.style={draw=jLine,fill=white,line width=0.6pt,rounded corners=2.5pt},
  jpanel/.style={fill=jPanel,draw=jPanelLine,line width=0.6pt,rounded corners=3pt},
  jtrain/.style={draw=jNote,line width=0.5pt,dash pattern=on 2pt off 1.5pt,rounded corners=3pt},
  jquery/.style={line width=0.5pt,rounded corners=1.5pt,minimum width=11pt,minimum height=7.5pt,inner sep=0,font=\scriptsize},
}

\newcommand\jflame[2]{%
  \begin{scope}[shift={(#1,#2)},scale=0.32]
    \shade[inner color=yellow!80!orange,outer color=red!85!orange]
      (0,0) .. controls (-8.2,0) and (-10.7,-8.3) .. (-6.7,-14.8) .. controls (-5.7,-11.8) and (-4.2,-10.8) .. (-3.2,-10.8)
      .. controls (-4.2,-17.3) and (-0.2,-22.3) .. (3.8,-24.1) .. controls (2.8,-19.3) and (5.8,-17.3) .. (7.8,-13.3)
      .. controls (10.8,-7.3) and (8.3,0) .. (0,0) -- cycle;
    \fill[yellow!70!orange] (0.4,-1) .. controls (-4.2,-1) and (-5.2,-6.3) .. (-2.2,-9.8) .. controls (-1.7,-7.8) and (-0.7,-7.3) .. (0.3,-7.3)
      .. controls (0.3,-10.3) and (1.8,-12.3) .. (3.3,-13.3) .. controls (3.3,-10.3) and (5.8,-8.3) .. (5.3,-5.3)
      .. controls (4.8,-2.3) and (2.8,-1) .. (0.4,-1) -- cycle;
  \end{scope}}

\DeclareRobustCommand\jbadge[1]{\tikz[baseline=(b.base)]\node[circle,draw=jLine,fill=white,line width=0.4pt,inner sep=0,minimum size=7pt,font=\scriptsize] (b) {#1};}
\DeclareRobustCommand\jflameinline{\tikz[x=0.85pt,y=-0.85pt,baseline=0pt]{\jflame{0}{0}}}
\newcommand\jnum[3]{\node[circle,draw=jLine,fill=white,line width=0.5pt,inner sep=0,minimum size=8.5pt,font=\scriptsize] at (#1,#2) {#3};}

\newcommand\jvoxgeom[4]{\def\jvp{#1}\def\jvs{#2}\def\jvd{#3}\def\jvdd{#4}}
\jvoxgeom{6}{5.4}{2.8}{2.4}
\def\jvn{2}
\def\jvface{white}  %
\makeatletter
\newcommand\jvox[4]{\def\jv@base{#3}\let\jv@colors\jv@levelcolors\jv@draw{#1}{#2}{#4}}
\newcommand\jvoxc[3]{\let\jv@colors\jv@namedcolors\jv@draw{#1}{#2}{#3}}
\def\jv@levelcolors#1{%
  \ifcase#1\relax
  \or\colorlet{vF}{\jv@base!22!white}\colorlet{vT}{\jv@base!12!white}\colorlet{vL}{\jv@base!35!white}%
  \or\colorlet{vF}{\jv@base!55!white}\colorlet{vT}{\jv@base!38!white}\colorlet{vL}{\jv@base!70!white}%
  \or\colorlet{vF}{\jv@base}\colorlet{vT}{\jv@base!65!white}\colorlet{vL}{\jv@base!75!black}%
  \fi}
\def\jv@namedcolors#1{\colorlet{vF}{#1}\colorlet{vT}{#1!65!white}\colorlet{vL}{#1!75!black}}
\def\jv@draw#1#2#3{%
  \foreach \i/\j/\k/\l in {#3} {\expandafter\xdef\csname jvx-\i-\j-\k\endcsname{\l}}%
  \foreach \k in {\jvn,...,0} {\foreach \j in {\jvn,...,0} {\foreach \i in {\jvn,...,0} {
    \pgfmathsetmacro\px{#1+\i*\jvp-\k*\jvd}\pgfmathsetmacro\py{#2+\j*\jvp-\k*\jvd}%
    \colorlet{vF}{\jvface}\colorlet{vT}{jVoxTop}\colorlet{vL}{jVoxSide}%
    \ifcsname jvx-\i-\j-\k\endcsname\expandafter\jv@colors\expandafter{\csname jvx-\i-\j-\k\endcsname}\fi
    \begin{scope}[draw=jVoxEdge,line width=0.25pt]
      \filldraw[fill=vT] (\px,\py) -- ++(\jvs,0) -- ++(-\jvdd,-\jvdd) -- ++(-\jvs,0) -- cycle;
      \filldraw[fill=vL] (\px,\py) -- ++(0,\jvs) -- ++(-\jvdd,-\jvdd) -- ++(0,-\jvs) -- cycle;
      \filldraw[fill=vF] (\px,\py) rectangle ++(\jvs,\jvs);
    \end{scope}
  }}}%
  \foreach \i/\j/\k/\l in {#3} {\expandafter\global\expandafter\let\csname jvx-\i-\j-\k\endcsname\jv@undefined}%
}
\makeatother

\newcommand\jct[4]{%
  \node[inner sep=0,anchor=north west] at (#1,#2) {\includegraphics[width=#3pt]{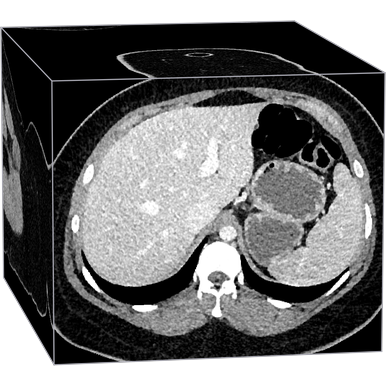}};
  \ifnum#4=1
    \begin{scope}[shift={(#1,#2)},scale=#3/386,opacity=0.72]
      \fill[orgA] plot[smooth cycle,tension=0.6] coordinates {(178,108) (201,110) (226,109) (249,108) (252,126) (248,148) (250,172) (240,188) (228,200) (217,214) (205,228) (192,240) (182,254) (168,266) (152,274) (127,274) (104,272) (92,260) (84,243) (72,229) (82,213) (84,191) (86,169) (100,159) (110,144) (125,134) (140,123) (156,114) (175,108)};
      \fill[orgB] plot[smooth cycle,tension=0.6] coordinates {(187,262) (194,265) (192,273) (190,282) (186,288) (180,292) (174,296) (168,299) (160,302) (151,302) (141,302) (132,304) (123,302) (116,300) (110,296) (104,292) (100,286) (104,280) (111,284) (118,286) (127,288) (137,288) (146,286) (154,284) (161,282) (168,278) (174,275) (180,270) (184,265)};
      \fill[orgB] plot[smooth cycle,tension=0.6] coordinates {(244,266) (250,268) (254,272) (258,276) (263,278) (269,280) (275,282) (281,284) (287,286) (294,288) (302,288) (310,288) (316,286) (323,284) (326,289) (322,293) (318,297) (313,300) (307,302) (300,302) (292,302) (287,300) (280,298) (274,296) (267,296) (261,294) (256,290) (251,288) (246,284) (246,277) (244,270)};
      \fill[orgC] plot[smooth cycle,tension=0.6] coordinates {(223,234) (230,235) (230,242) (232,246) (235,250) (237,256) (238,262) (238,268) (238,274) (233,276) (228,277) (225,272) (220,270) (214,271) (209,274) (208,280) (210,285) (212,291) (206,292) (200,291) (200,284) (199,278) (198,273) (198,267) (200,262) (202,256) (204,252) (206,247) (211,244) (216,242) (222,241) (222,235)};
    \end{scope}
  \fi}

\def\jobr{0.45}\def\jobimg{figures/ct_cube_ob}
\def\jorgA{orgA}\def\jorgB{orgB}\def\jorgC{orgC}
\newcommand\jcto[4]{%
  \pgfmathsetmacro\jobd{600*\jobr}\pgfmathsetmacro\jobw{600+\jobd}%
  \node[inner sep=0,anchor=north west] at (#1,#2) {\includegraphics[width=#3pt]{\jobimg.png}};
  \begin{scope}[shift={(#1,#2)},scale=#3/\jobw]
    \draw[jVoxEdge,line width=0.3pt] (\jobd,\jobd) rectangle (\jobw,\jobw) (0,0) -- (600,0) -- (\jobw,\jobd) (0,0) -- (0,600) -- (\jobd,\jobw) (0,0) -- (\jobd,\jobd);
    \ifnum#4=1
    \begin{scope}[opacity=0.72,shift={(\jobd-270,\jobd-270)}]
      \fill[\jorgA] plot[smooth cycle,tension=0.6] coordinates {(496,339) (537,344) (582,343) (624,342) (629,380) (622,426) (626,477) (608,511) (586,535) (566,565) (545,594) (521,619) (503,648) (478,673) (449,689) (404,688) (363,683) (342,657) (327,621) (306,591) (324,557) (327,511) (331,465) (356,444) (374,413) (401,392) (428,370) (456,351) (491,339)};
      \fill[\jorgB] plot[smooth cycle,tension=0.6] coordinates {(512,665) (525,671) (521,688) (518,707) (510,720) (500,728) (489,736) (478,742) (464,748) (447,748) (430,747) (413,751) (397,747) (385,742) (374,733) (363,725) (356,712) (363,699) (376,708) (388,713) (404,717) (422,718) (439,714) (453,710) (465,706) (478,698) (489,692) (500,681) (507,671)};
      \fill[\jorgB] plot[smooth cycle,tension=0.6] coordinates {(615,676) (626,680) (633,689) (640,697) (649,702) (660,706) (671,711) (682,715) (693,720) (705,724) (720,725) (734,725) (745,721) (758,717) (763,728) (756,736) (749,744) (740,750) (729,754) (716,754) (702,754) (693,749) (680,745) (669,740) (656,740) (646,735) (637,727) (628,722) (619,714) (619,699) (615,684)};
      \fill[\jorgC] plot[smooth cycle,tension=0.6] coordinates {(577,607) (590,610) (590,624) (593,633) (599,641) (602,654) (604,667) (604,680) (604,692) (595,696) (586,698) (581,688) (572,683) (561,685) (552,691) (550,704) (554,714) (557,727) (546,729) (536,727) (536,712) (534,699) (532,689) (532,676) (536,665) (539,653) (543,644) (546,634) (555,628) (564,624) (575,622) (575,609)};
    \end{scope}
    \fi
  \end{scope}}
\newcommand\jmaskedo[3]{%
  \jcto{#1}{#2}{#3}{0}
  \pgfmathsetmacro\jobd{600*\jobr}\pgfmathsetmacro\jobw{600+\jobd}%
  \begin{scope}[shift={(#1,#2)},scale=#3/\jobw,draw=jVoxEdge,line width=0.4pt]
    \filldraw[fill=jVoxTop] (300+\jobd,\jobd) -- (\jobw,\jobd) -- (600+\jobd/2,\jobd/2) -- (300+\jobd/2,\jobd/2) -- cycle;
    \filldraw[fill=jVoxSide] (300+\jobd,\jobd) rectangle (\jobw,300+\jobd);
  \end{scope}}
\newcommand\jcropboxvo[3]{%
  \pgfmathsetmacro\jobd{600*\jobr}\pgfmathsetmacro\jobw{600+\jobd}\pgfmathsetmacro\jobb{258*\jobr}%
  \begin{scope}[shift={(#1,#2)},scale=#3/\jobw]
    \begin{scope}[shift={(\jobd,\jobd)}]
    \fill[white,opacity=0.28] (47,92) rectangle (305,349);
    \begin{scope}[draw=jSSL!70!black,line width=1.3pt]\draw (47,92) rectangle (305,349) (47,92) -- (47-\jobb,92-\jobb) -- (305-\jobb,92-\jobb) -- (305,92) (47-\jobb,92-\jobb) -- (47-\jobb,349-\jobb) -- (47,349);\end{scope}
    \begin{scope}[draw=jSSL!22!white,line width=0.7pt,dash pattern=on 1.6pt off 1pt]\draw (47,92) rectangle (305,349) (47,92) -- (47-\jobb,92-\jobb) -- (305-\jobb,92-\jobb) -- (305,92) (47-\jobb,92-\jobb) -- (47-\jobb,349-\jobb) -- (47,349);\end{scope}
    \end{scope}
  \end{scope}}
\newcommand\jlocalo[3]{%
  \begin{scope}[shift={(#1,#2)}]
    \pgfmathsetmacro\jld{\jobr*#3}
    \filldraw[fill=black!88,draw=black!45,line width=0.3pt] (\jld,\jld) -- (\jld+#3,\jld) -- (#3,0) -- (0,0) -- cycle;
    \filldraw[fill=black!94,draw=black!45,line width=0.3pt] (\jld,\jld) -- (\jld,\jld+#3) -- (0,#3) -- (0,0) -- cycle;
    \node[inner sep=0,anchor=north west] at (\jld,\jld) {\includegraphics[width=#3pt,height=#3pt]{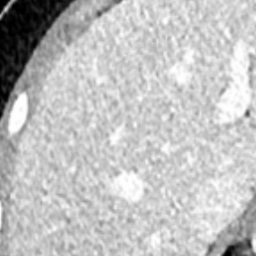}};
    \draw[black!45,line width=0.3pt] (\jld,\jld) rectangle +(#3,#3);
  \end{scope}}

\newcommand\jpyrsc[1]{\colorlet{jVoxTop}{#1!22!white}\colorlet{jVoxSide}{#1!42!white}\colorlet{jVoxEdge}{#1!65!jVoxEdge}\def\jvface{#1!10!white}}

\usepackage{pgfplots}
\pgfplotsset{compat=1.18}
\usepackage{pgfplotstable}
\usepgfplotslibrary{groupplots}

\renewcommand{\topfraction}{0.9}
\renewcommand{\floatpagefraction}{0.8}
\renewcommand{\textfraction}{0.08}

\usepackage[breaklinks]{hyperref}
\hypersetup{colorlinks=true, urlcolor=linkblue, linkcolor=linkblue, citecolor=black, anchorcolor=linkblue,
  filecolor=linkblue, pdftitle={Jolia 2: Reconciling Vision-Language Alignment and Dense Visual Learning for 3D CT Foundation Models}}

\newcommand{\PaperTitle}{Jolia 2: Reconciling Vision-Language Alignment and Dense Visual Learning for 3D CT Foundation Models}
\newcommand{\PaperDate}{November 26th, 2026}
\newcommand{\PaperKeywords}{foundation model, computed tomography, CT, vision-language pretraining, self-supervised learning,
segmentation}
\newcommand{\CorrespondingEmail}{corentin.dancette@raidium.eu}

\begin{document}
\RaidiumHeader{%
  \RaidiumTitle{\PaperTitle}
  \RaidiumAuthors{%
    Julien Khlaut\textsuperscript{1\,\textdagger*},
    L\'eo Butsanets\textsuperscript{1\,\textdagger},
    Baptiste Callard\textsuperscript{1},
    Charles Corbi\`ere\textsuperscript{1},
    Th\'eo Danielou\textsuperscript{1},
    Antoine Saporta\textsuperscript{1},
    Maxime Ronot\textsuperscript{2},
    Corentin Dancette\textsuperscript{1*}}
  \RaidiumAffiliations{%
    1*~Raidium, 27 rue du faubourg Saint-Jacques, Paris, 75014, France.
    2~Department of Radiology, FHU~MOSAIC, Beaujon Hospital, APHP.Nord, Clichy, France.}
  \RaidiumNotes{%
    *Corresponding authors. E-mail: \RaidiumURL{mailto:julien.khlaut@raidium.eu}{julien.khlaut@raidium.eu}, \RaidiumURL{mailto:\CorrespondingEmail}{\CorrespondingEmail};\\
    \textsuperscript{\textdagger}These authors contributed equally to this work.}
  \RaidiumAbstract{\begin{abstract}
A foundation model should serve many tasks with one encoder. 
In CT, finding classification and segmentation are the two most common tasks, yet no model leads on both. Image-text alignment helps to learn global representations adapted for classification, but leaves spatial features weakly structured. 
Image-only self-supervision and segmentation training give well-structured spatial features, but lag behind on classification tasks.
We argue that this trade-off comes from a missing spatial signal in report alignment, and that organ masks can supply it.
We introduce \JoliaTwo, a foundation model trained with a novel method, \Method (Masks, Text and Image Self-supervision), which adds organ masks to report alignment and image-only self-supervision.
It builds on \Conquer, which aligns organ queries with the matching sections of the report, and was used to train the first version of Jolia.
The masks now guide where each organ query looks and train the spatial features through a segmentation head.
Masks are generated by an automatic segmentation model and require no human annotation. Unlike most mask-based methods, \Method uses masks as training targets, and not at inference.
On five vision backbones, this improves both tasks by 1.8 AUC on finding classification, and 3.0 Dice for segmentation on average over \Conquer.
A coarse spatial signal is enough, as a point or a box per organ keeps most of the gain on finding classification.
\JoliaTwo, trained with \Method on public data, ranks first among about 30 CT models on finding classification (mean AUC 88.9 vs. 85.9), and leads zero-shot classification and retrieval. Additionally, \JoliaTwo helps to bridge the gap between findings classification and segmentation: with a frozen encoder, it is ahead of all self-supervised and vision-language models, only behind a supervised segmentation model. These results show that one frozen encoder can serve both tasks, once report alignment receives a spatial signal.
Weights will be released shortly.
\end{abstract}
}
  \RaidiumKeywords{\PaperKeywords}
  \RaidiumDate{\PaperDate}
  \RaidiumLogo{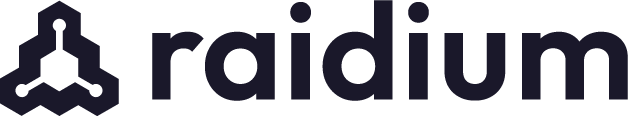}
}
\section{Introduction}
\label{sec:intro}
\begin{figure}[t]
  \centering
  \includegraphics[width=\columnwidth]{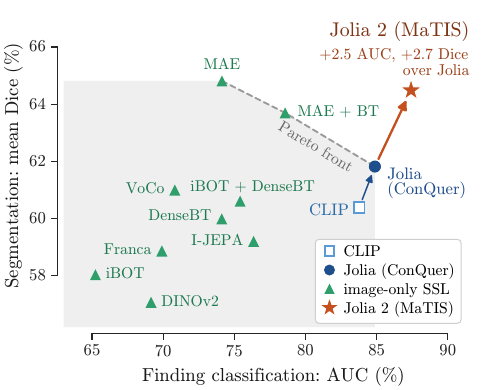}
  \caption{\textbf{\JoliaTwo, trained with \Method lies beyond the Pareto front of report alignment and image-only pretraining.} Each point is one pretraining method applied to the same \HieraCT encoder and data. The encoder is then frozen to measure the mean AUC on finding classification and the mean Dice over 13 segmentation datasets (ablation setting, \cref{sec:ablations}). Blue: report alignment. Green: image-only pretraining. Dashed line: Pareto front without \Method (grey: dominated). BT: Barlow Twins~\cite{barlowtwins}.}
  \label{fig:controlled_comparison}
\end{figure}

\begin{figure*}[t]
  \centering
  \input{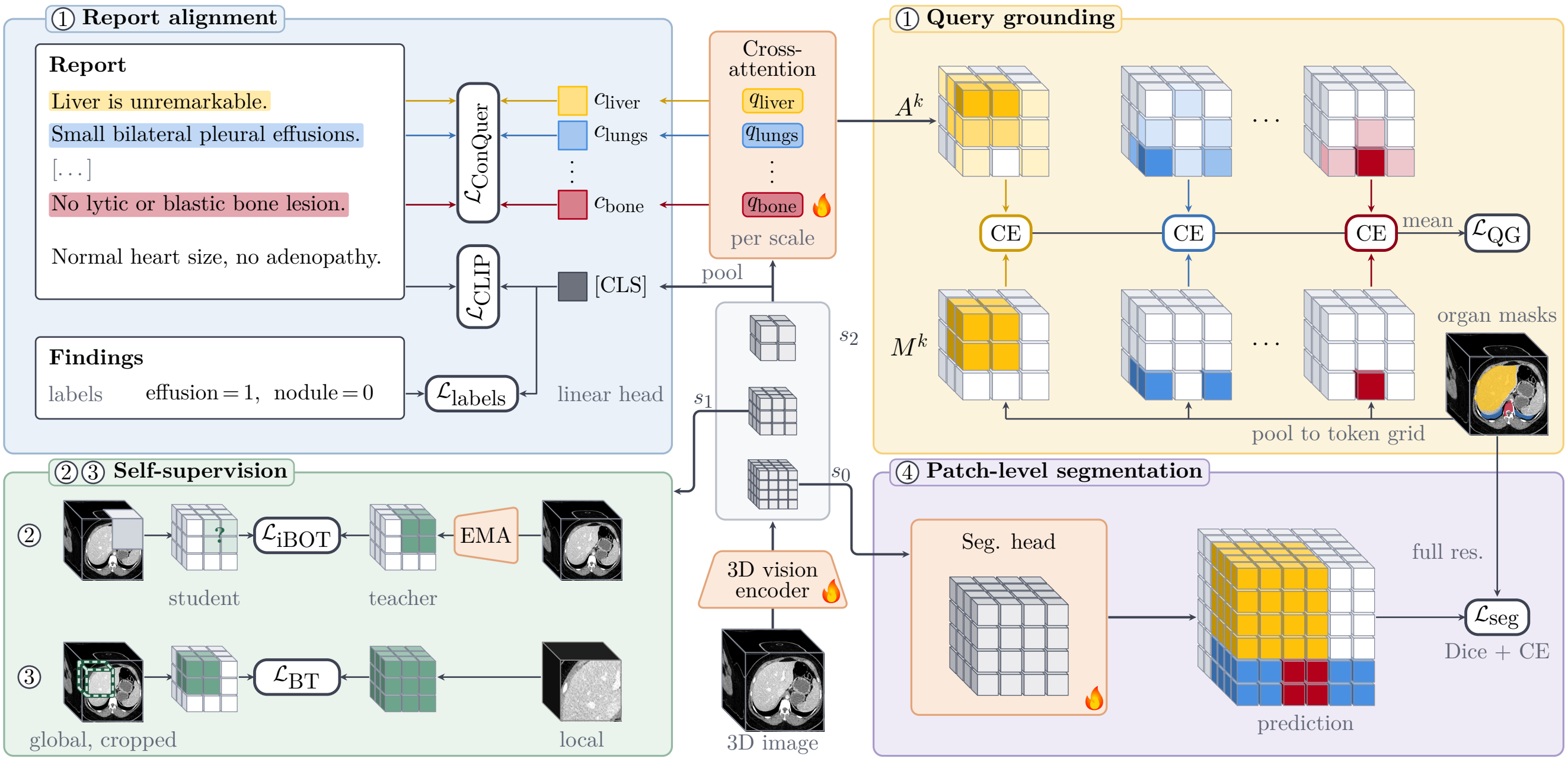}
    \caption{\textbf{Training method of \JoliaTwo (MaTIS)}
  The four phases alternate, one per training step, on the same hierarchical 3D encoder. Organ masks are used only in training, to guide the organ queries \jbadge{1} and to supervise the patch tokens \jbadge{4}. \jflameinline{} marks the trained modules.
  \jbadge{1} \textit{Report alignment, query grounding.} Each organ $k$ has one query per token scale (one scale shown). $\mathcal{L}_{\text{ConQuer}}$ aligns the organ token $c_k$ with the sentences about the organ. $\mathcal{L}_{\text{QG}}$ is the cross-entropy of the query attention $A^k$ against $M^k$, the share of the organ mask in each token. [CLS] is aligned with the report ($\mathcal{L}_{\text{CLIP}}$) and classified by a linear head ($\mathcal{L}_{\text{labels}}$).
  \jbadge{2}\,\jbadge{3} \textit{Self-supervision.} \iBOT~\cite{zhou2022ibot} matches masked student tokens to those of an EMA teacher. \DenseBT (\cref{sec:ssl}) aligns the tokens of a global and a local view.
  \jbadge{4} \textit{Patch-level segmentation.} A light head on $s_0$ predicts the TotalSegmentator classes. \JoliaTwo adds an opposite sentence loss $\mathcal{L}_{\text{OSL}}$ (\cref{app:osl}).}

  \label{fig:method}
\end{figure*}

A radiologist reading a CT scan both identifies the findings and locates them in the volume.
Most CT models address only one of these tasks, either finding classification or segmentation.
Findings are mostly classified rather than segmented, as their masks are costly to annotate. A foundation model is pretrained once without manual annotation and reused on many tasks, either zero-shot or with a small head~\cite{blankemeier2024merlin,ctfm,pillar0,curia}. Yet in CT, the best encoders for finding classification and segmentation still differ~\cite{flexictvlm,vista3d}.

Each CT scan comes with a report, in which a radiologist describes the findings. Aligning a volume with its report, in the spirit of \CLIP~\cite{radford2021clip}, gives recent models high performance on finding classification without any image annotation~\cite{blankemeier2024merlin,pillar0,flexictvlm}.
However, a report may omit the less salient findings and locate the others only coarsely. The alignment thus matches a sparse text with a global representation pooled over the whole volume.
This compression loses the patch-level information that segmentation needs.
Image-only self-supervision~\cite{voco,coralbay,curiamae} learns the local structure of the volume and keeps it in the patch tokens.
It receives no signal about findings, which remain implicit in the image and harder to classify.
Segmentation models supervised on masks~\cite{suprem,vista3d} learn the anatomy precisely but transfer poorly to findings.
\Cref{fig:controlled_comparison} shows this tension on a single backbone. Moving from report alignment to image-only pretraining trades one task for the other.

To reduce this trade-off, recent work adds local information to report alignment.
Some models~\cite{colipri,spectre,flexictvlm} combine report alignment with image-only self-supervision, but still segment worse than supervised and image-only models. Other models~\cite{fvlm,totalfm,arcct,captor,jolia} align parts of the report with the matching organs.
Most of them need a crop or a mask of each organ, often at inference too. The others learn where each organ is from the report alone. \Conquer~\cite{jolia} matches one learned query per organ with the report sentences about that organ. The attention of these queries rarely lands on their organ (\cref{app:localisation}).

Our hypothesis is that report alignment lacks a spatial signal, and that organ masks can provide one.
We propose \Method (Masks, Text and Image Self-supervision), a pretraining method that combines report alignment and image-only self-supervision with organ masks (\cref{fig:method}). It builds on \Conquer and adds to the organ-level report alignment a \emph{query grounding} loss that uses the masks to guide the attention of each organ query. Image-only self-supervision trains every patch token from the image alone, while \emph{patch-level segmentation} trains them to predict the masks through a light head.
Open segmenters such as TotalSegmentator~\cite{wasserthal2023totalsegmentator} generate the masks without new annotation. No mask is needed at inference.

On the five vision backbones we test, \Method lies beyond the Pareto front of earlier methods and gains 1.8 AUC and 3.0 Dice on average over \Conquer, the method it builds on, so the gap measures what the masks and self-supervision add.
The masks behind this gain can be coarse. With a box or a single point per organ instead of its mask, query grounding keeps about two thirds of its gain on finding classification, whereas boxes placed at random bring almost none. 
The gain even reaches the findings of organs that have no mask.
We use \Method to train \JoliaTwo on public data only.

Our contributions are:
\begin{enumerate}
  \item \textbf{A pretraining method for CT foundation models.} \Method combines report alignment, image-only self-supervision and organ masks. The masks act only as targets for the attention of each organ query and for the patch tokens. On five vision backbones, \Method lies beyond the Pareto front of \CLIP, \Conquer and image-only pretraining.
  \item \textbf{A state-of-the-art open-source model.} \JoliaTwo is a CT foundation model trained with \Method on public data only. Its encoder is \HieraCT, our variant of 3D Hiera~\cite{ryali2023hiera}, which improves finding classification and segmentation. It outperforms about 30 CT models on finding classification, zero-shot classification and retrieval. With a frozen encoder, it ranks second on segmentation, behind VISTA3D, which is supervised on several of our segmentation datasets (\cref{fig:competitors}).
  \item \textbf{A joint evaluation and controlled comparison.} We evaluate vision-language, image-only and supervised CT models on finding classification and segmentation, with frozen and fine-tuned encoders. We train 15 pretraining configurations on \HieraCT with the same data and five of them on four other backbones.

\end{enumerate}

\section{Related work}
\label{sec:related}

\paragraph{Foundation models for 3D CT.}
Most CT foundation models are pretrained without annotations made for training, in four families:
(1) global vision-language alignment~\cite{hamamci2024ctclip,m3d,blankemeier2024merlin,pillar0,radfinder};
(2) image-only self-supervision, by self-distillation~\cite{coralbay,tapct,flexictvlm}, crop-level contrast~\cite{ctfm,voco} or reconstruction~\cite{nnfoundation,curiamae};
(3) hybrids that add language supervision to such objectives~\cite{colipri,spectre,flexictvlm};
and (4) anatomical decomposition, which aligns organs or crops with parts of the report~\cite{fvlm,totalfm,arcct,captor,jolia}.
For dense tasks, supervised models~\cite{suprem,stunet,wasserthal2023totalsegmentator,vista3d} set the reference.

\paragraph{Organ masks in vision-language pretraining.}
Several works use organ masks to recover the anatomy lost by global alignment: they crop the volume per organ~\cite{totalfm}, pool the tokens of each organ with its mask~\cite{ctglip,okact,ocpct}, restrict the attention of each query to a provided mask~\cite{fvlm,arcct,burdenorder} or to a predicted one~\cite{radar,dcvlp}, or train query logits as mask predictors~\cite{scope}.
Closest to us, TotalFM~\cite{totalfm} pretrains with VideoMAE and then aligns organ crops with report sentences, SCOPE~\cite{scope} freezes the encoder during report alignment, and RADAR+~\cite{radar} trains a decoder on masks alongside the reports but keeps it at inference.
\Method differs in two ways: the mask is a soft target for the attention~\cite{liu2017attention} rather than a selection of tokens, and the head that trains the patch tokens is discarded after pretraining, so inference needs neither a mask nor a decoder.

\paragraph{Self-supervision alongside language.}
Language and self-supervision are combined to obtain global and dense features in natural images~\cite{tips,siglip2} and in CT~\cite{spectre,flexictvlm,colipri}, but none of these CT models uses organ-level structure.
We alternate \iBOT~\cite{zhou2022ibot} with organ-wise report alignment in a single training, and add \DenseBT, a redundancy reduction on tokens paired by their physical position, with no teacher or negative pair; earlier local criteria~\cite{vicregl,curiamae} pair or pool tokens differently.
These works combine masks, segmentation heads, self-supervision and language in different ways, but none tests which spatial signal report alignment needs. We show that a coarse signal per organ is enough, and that its gain reaches organs that never receive it.

\section{Method}
\label{sec:method}
\Method adds image-only self-supervision and organ masks to the organ-level report alignment of \Conquer. The masks guide the queries and train the patch tokens. Training cycles through four phases, one per step (\cref{fig:method}): (1) report alignment with query grounding, (2) \iBOT, (3) \DenseBT and (4) patch-level segmentation.
The phases alternate because their crops and batch sizes differ and the losses do not all fit in GPU memory in one step. The reports and the masks are not needed at inference.

\subsection{Backbone}
\label{sec:backbone}
\begin{figure}[t]
  \centering
  \begin{tikzpicture}[x=1pt,y=-1pt,line join=round,font=\scriptsize,
      hstage/.style={jblock,fill=pBlock,draw=pBlockLine,minimum width=36pt,minimum height=24pt,inner sep=0,align=center},
      hchange/.style={draw=pBlockLine!60!black,fill=pBlockLine!65!white,line width=0.8pt,rounded corners=2.5pt,inner sep=0,align=center},
      hflow/.style={jflow,line width=0.75pt}]
    \def\jsb#1{\text{\fontsize{6}{7}\selectfont #1}}%
    \def\jobr{0.30}\def\jobimg{figures/ct_cube_ob30}   %
    \colorlet{pSc}{jLine}\colorlet{pCLS}{jLine}        %
    \jcto{0}{2}{26}{0}
    \draw[hflow] (26.5,15) -- (34,15);
    \node[hchange,minimum height=24pt,minimum width=26pt] at (47.5,15) {Conv\\[-1pt]stem};
    \draw[hflow] (61.2,15) -- (66.5,15);
    \foreach \x/\n/\a in {85/0/windowed,149.5/1/global,214/2/global}
      \node[hstage] at (\x,15) {Stage \n\\[-1pt]\a};
    \foreach \l/\r/\x in {103.5/131.2/117.25,168/195.7/181.75} {
      \draw[hflow] (\l,15) -- (\r,15);
      \node[hchange,minimum width=16pt,minimum height=11pt] at (\x,15) {max};
    }
    \foreach \x/\s/\g in {85/0/48,149.5/1/24,214/2/12} {
      \draw[hflow] (\x,27.5) -- (\x,41.5);
      \node[anchor=west,inner sep=0,text=pSc] at (\x+2,34.5) {$s_{\jsb{\s}},\,\g^3$};
    }
    \node[hchange,minimum width=165pt,minimum height=11pt] at (149.5,47.5) {average + maximum of each scale};
    \draw[hflow] (149.5,53.5) -- (149.5,59.5);
    \node[draw=pCLS,fill=pCLS!15!white,line width=0.6pt,rounded corners=2pt,minimum width=34pt,minimum height=11pt,inner sep=0] at (149.5,65.5) {[CLS]};
  \end{tikzpicture}
  \caption{\textbf{\HieraCT.} Darker blocks: the three changes from the base Hiera. The stages give the token scales $s_0$, $s_1$ and $s_2$, with their grids. The pooled scales are concatenated and projected to the [CLS] token.}
  \label{fig:hiera}
\end{figure}
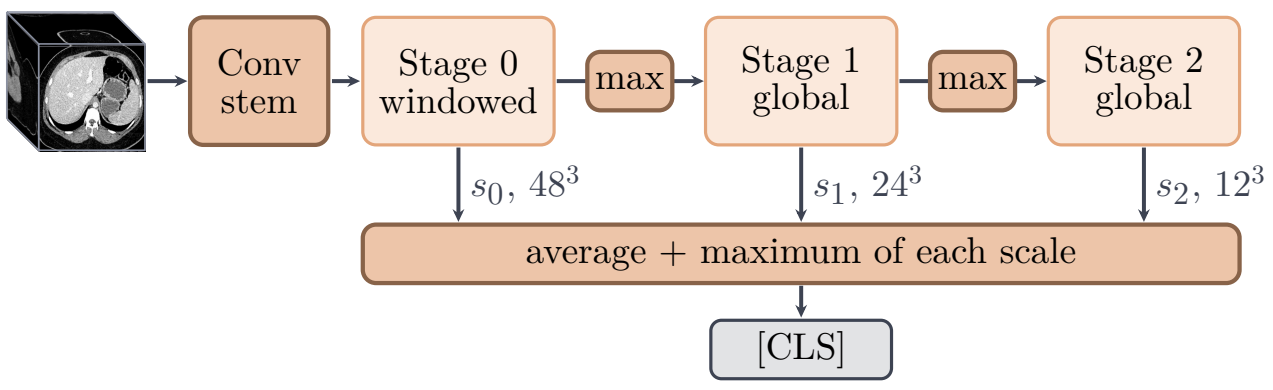
\begin{figure*}[t]
  \centering
  \includegraphics[width=\textwidth]{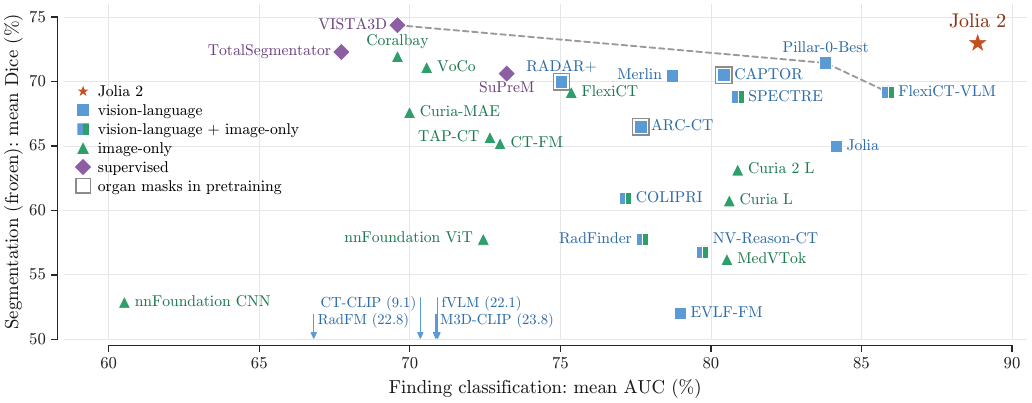}
   \caption{\textbf{Comparison with external CT foundation models, frozen encoder.} Mean Dice over the 13 segmentation datasets, all classes, against mean finding AUC (\cref{fig:controlled_comparison} uses the ablation setting). Half-and-half squares: models trained with both language and image-only self-supervision. Grey outline: models that use TotalSegmentator masks in pretraining. Dashed line: Pareto front of the external models. Arrows: models below the Dice axis, with their Dice. Pretraining data differ between models (\cref{tab:methods}). Per-model values in \cref{tab:competitors,tab:dense}.}
  \label{fig:competitors}
\end{figure*}
\HieraCT is a hierarchical 3D encoder derived from Hiera~\cite{ryali2023hiera} (\cref{fig:hiera}). Its three stages give the token scales $s_0$, $s_1$ and $s_2$, from fine to coarse (\cref{app:training}). The input has 11 channels, the intensity windows of Pillar-0~\cite{pillar0}.
It differs from a base 3D Hiera in three ways.
First, a convolutional stem with overlapping kernels replaces the patchify layer. Non-overlapping patches can split a small lesion or finding across several tokens, and convolutional stems also make vision transformers easier to optimise~\cite{xiao2021early}.
Then, max pooling replaces average pooling between stages.
Finally, the global token $z_\texttt{[CLS]}$ pools each scale by the sum of its average and its maximum, to see both diffuse and focal findings. Those three changes improve both classification and segmentation (\cref{tab:ablation}).

\subsection{Report alignment}
\label{sec:prelim}
The queries follow the $K$ organs of an anatomical taxonomy~\cite{jolia}.
Following \Conquer~\cite{jolia}, each organ $k$ has one learnable query per scale.
The query pools the patch tokens of its scale by cross-attention, which gives an attention distribution $A^k$ over these tokens.
The outputs of the scales are concatenated into an organ token $c_k$.
A frozen text encoder~\cite{zhang2025qwen3embedding} embeds the full report into $t$ and the section of the report about organ $k$ into $t_k$.
A symmetric InfoNCE loss~\cite{radford2021clip} aligns $z_\texttt{[CLS]}$ with $t$ ($\mathcal{L}_{\text{CLIP}}$) and, separately for each organ, $c_k$ with $t_k$ on the volumes whose report has such a section ($\mathcal{L}_{\text{ConQuer}}$).

Report alignment gives no explicit target for each finding. Like ARC-CT~\cite{arcct}, we add one with $\mathcal{L}_{\text{labels}}$, a binary cross-entropy on a linear head over $z_\texttt{[CLS]}$ with the finding labels extracted from the reports.

\subsection{Image self-supervision}
\label{sec:ssl}
A report describes only part of a volume. Textures and unreported findings are absent from it.
We add two image-only objectives that train each patch token on the local structure of the volume, with no additional data. Both act on the tokens of $s_1$, which are eight times fewer than those of $s_0$.
\iBOT learns from the masked context within one view, and \DenseBT from the agreement between two views.

\paragraph{\iBOT.}
The student encodes a volume in which 60\% of the blocks are masked, and matches the prototype assignments that an EMA teacher computes on the full volume~\cite{zhou2022ibot}, for all the tokens of scale $s_1$.

\paragraph{Dense Barlow Twins (\DenseBT).}
\DenseBT applies the redundancy reduction of Barlow Twins~\cite{barlowtwins} to tokens instead of global embeddings.
From one volume we draw a global view and a local view, a random sub-box with a side of 64 to 160\,mm, and encode both with the same encoder.
We resample the global tokens of $s_1$ at the positions of the local tokens, which gives pairs of tokens at the same physical position. After a projector, the cross-correlation matrix $C$ of these pairs is driven towards the identity:
\begin{equation}
  \mathcal{L}_{\text{BT}} = \sum_{a} (1 - C_{aa})^2 + \lambda \sum_{a \neq b} C_{ab}^2,
  \label{eq:bt}
\end{equation}
with $\lambda = 0.005$, and no teacher, stop-gradient or negatives.

\subsection{Query grounding}
\label{sec:coverage}
In \Conquer, the attention $A^k$ of a query is shaped by the report alone, so a query may attend to regions unrelated to its organ.
\emph{Query grounding} supervises this attention with an organ mask.
We run TotalSegmentator~\cite{wasserthal2023totalsegmentator} once on every training volume.
The queries follow the $K=102$ organs of the report taxonomy, not the classes of TotalSegmentator.
A fixed table gives each organ the union of the classes that cover it: the five lung lobes form the lungs, and the 24 ribs form the ribs (\cref{app:coverage}).
Fifty-five organs are covered this way, some of them through a neighbouring structure; the others, such as the abdominal wall, have no class and no query grounding (\cref{app:coverage}).

For each token scale, the target $M^k_i$ is the share of the mask of organ $k$ that falls in token $i$. The loss is the cross-entropy between this target and the attention $A^k$:
\begin{equation}
  \ell^{k} = - \sum_{i} M^k_{i} \log A^k_{i}.
  \label{eq:cov}
\end{equation}
$\mathcal{L}_{\text{QG}}$ is the mean of this term over the token scales and the organs with a mask in the crop.
It adds no parameters.

\subsection{Patch-level segmentation}
\label{sec:dense}
\emph{Patch-level segmentation} uses the TotalSegmentator masks directly, with all their classes, on the patch tokens.
A light convolutional head on the finest tokens $s_0$, which localise most precisely, predicts the 118 classes of TotalSegmentator (117 structures and the background) on a $96^3$ grid, with a Dice and a cross-entropy loss of equal weight, $\mathcal{L}_{\text{seg}}$.
Whereas $\mathcal{L}_{\text{QG}}$ acts on the attention of the queries, this loss acts on the tokens that a downstream decoder will read.
The head is discarded after pretraining.

\paragraph{\JoliaTwo.}
\JoliaTwo is trained with \Method and one more loss, $\mathcal{L}_{\text{OSL}}$, adapted from the opposite sentence loss of COLIPRI~\cite{colipri}. It teaches the embedding of $\mathcal{L}_{\text{CLIP}}$ to separate a positive and a negative prompt for each finding label (\cref{app:osl}).
This loss targets zero-shot classification with short prompts, not the two tasks that \Method addresses. We keep it out of \Method, and the comparisons of methods and backbones run without it.

\section{Experiments}
\label{sec:results}

\subsection{Setup}
\label{sec:setup}
\begin{figure*}[t]
  \centering
  \includegraphics[width=\textwidth]{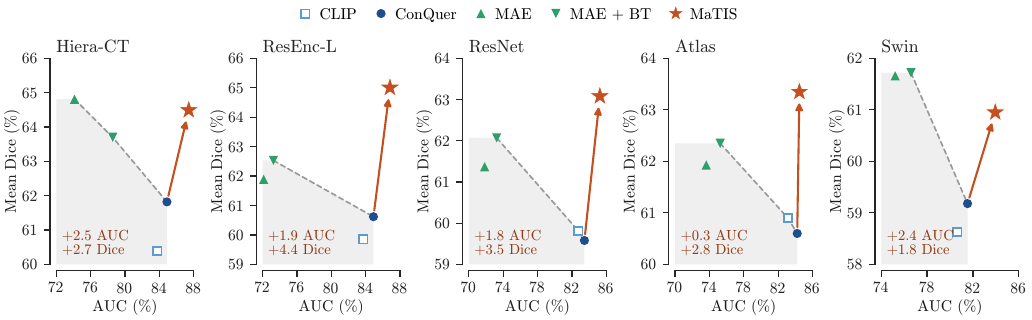}
  \caption{\textbf{\Method moves the Pareto front on every backbone.}
    Mean Dice over the 13 datasets (lesion class only for the lesion datasets) against AUC, with a frozen encoder. Dashed line: Pareto front of \CLIP, \Conquer, \MAE and \MAEBT on that backbone (grey: dominated). The arrow goes from \Conquer to \Method. Axes differ between panels.}
  \label{fig:arch_ablation}
\end{figure*}

\paragraph{Pretraining data.}
We pretrain on the training splits of three public datasets, \ctrate~\cite{hamamci2024ctclip} and \inspect~\cite{inspect2023} for the chest and \stanford~\cite{blankemeier2024merlin} for the abdomen, 58,379 CT--report pairs in total.
Reports are split into organ-tagged findings with the LLM pipeline of~\cite{jolia}, which also gives the 252 finding labels (\cref{app:data}).

\paragraph{Pretraining recipe.}
\label{sec:recipe}
\JoliaTwo is \HieraCT trained with \Method. Its final recipe adds three elements, measured in the last rows of \cref{tab:ablation}: (1) $\mathcal{L}_{\text{OSL}}$, (2) a finer resolution of 1\,mm, against 1.5\,mm in the ablations, and (3) a final cooldown of the learning rate.
Training lasts 180k steps on crops of $288^3$ voxels, on 48 GPUs (\cref{app:training}).

\paragraph{Evaluation data.}
We evaluate finding classification, zero-shot classification and image-text retrieval on four test sets.
Two are held-out splits of our pretraining datasets, the test split of \stanford (abdomen) and the validation split of \ctrate (chest), which releases no test split.
Two are private external sets, \extabdo and \extchest, never used in pretraining (\cref{app:data}).
We evaluate segmentation on 13 datasets of organs, lesions and vessels (\cref{tab:seg_targets}).

\paragraph{Baselines.}
We compare with the released checkpoints of about 30 external vision-language, image-only and supervised segmentation models (\cref{app:baselines}).
They differ from ours in pretraining data, size and resolution, and some saw our test distributions (\cref{tab:methods}). This comparison ranks models as they are; \cref{sec:ablations} compares methods on the same data.

\subsection{Finding classification}
\label{sec:res_findings}
\begin{table}[!htbp]
  \centering
  \footnotesize
  \caption{\textbf{Finding classification} AUC (\%) on the four test sets (Stanf.: Stanford) and their mean. \JoliaTwo: [CLS] token and organ queries; [CLS] only, queries only: global token or organ queries alone (one decimal). Bold: best per column.}
  \label{tab:competitors}
  \setlength{\tabcolsep}{2.2pt}
  \sbox0{\begin{tabular}{@{}lccccc@{}}
    \toprule
    Model & \extabdo & Stanf. & \extchest & CT-RATE & Mean \\
    \midrule
    \JoliaTwo & \textbf{85.65} & \textbf{88.25} & \textbf{92.85} & \textbf{88.68} & \textbf{88.86} \\
    \JoliaTwo, queries only & 84.6 & 87.9 & 92.2 & 88.1 & 88.2 \\
    \JoliaTwo, [CLS] only & 81.5 & 86.1 & 91.4 & 87.7 & 86.7 \\
    \midrule   %
    FlexiCT-VLM & 80.88 & 84.82 & 91.18 & 86.66 & 85.88 \\
    \Jolia & 77.61 & 83.59 & 89.02 & 86.44 & 84.16 \\
    Pillar-0-Best & 78.22 & 83.18 & 88.94 & 84.92 & 83.81 \\
    SPECTRE & 75.39 & 78.82 & 86.09 & 83.31 & 80.90 \\
    Curia 2 L & 74.85 & 79.08 & 87.34 & 82.30 & 80.89 \\
    Curia L & 72.43 & 79.16 & 88.55 & 82.31 & 80.61 \\
    MedVTok & 72.98 & 79.10 & 86.91 & 83.14 & 80.53 \\
    CAPTOR & 73.63 & 82.25 & 83.77 & 82.08 & 80.43 \\
    NV-Reason-CT & 69.34 & 78.68 & 85.06 & 85.80 & 79.72 \\
    EVLF-FM & 71.22 & 77.84 & 85.31 & 81.55 & 78.98 \\
    Merlin & 72.75 & 82.93 & 81.55 & 77.67 & 78.72 \\
    RadFinder & 71.48 & 78.27 & 81.60 & 79.55 & 77.72 \\
    ARC-CT & 64.56 & 72.20 & 86.62 & 87.32 & 77.67 \\
    COLIPRI & 66.46 & 75.28 & 82.80 & 84.10 & 77.16 \\
    FlexiCT & 68.11 & 74.39 & 81.25 & 77.68 & 75.36 \\
    RADAR+ & 63.34 & 82.90 & 79.55 & 74.37 & 75.04 \\
    SuPreM & 64.36 & 74.14 & 80.24 & 74.15 & 73.22 \\
    CT-FM & 64.02 & 70.41 & 80.97 & 76.62 & 73.00 \\
    TAP-CT & 65.38 & 69.50 & 80.76 & 74.98 & 72.66 \\
    nnFoundation ViT & 61.52 & 72.84 & 80.37 & 75.01 & 72.44 \\
    fVLM & 60.46 & 64.52 & 79.64 & 79.06 & 70.92 \\
    M3D-CLIP & 66.46 & 67.47 & 78.81 & 70.75 & 70.87 \\
    \VoCo & 63.34 & 70.04 & 78.21 & 70.64 & 70.56 \\
    CT-CLIP & 60.76 & 64.94 & 79.40 & 76.28 & 70.35 \\
    \CuriaMAE & 63.76 & 65.37 & 79.77 & 71.05 & 69.99 \\
    VISTA3D & 61.93 & 68.31 & 75.02 & 73.09 & 69.59 \\
    Coralbay & 61.04 & 67.81 & 78.45 & 71.04 & 69.59 \\
    TotalSegmentator & 62.64 & 66.05 & 74.57 & 67.65 & 67.73 \\
    RadFM & 59.93 & 62.47 & 76.79 & 68.04 & 66.81 \\
    nnFoundation CNN & 59.00 & 54.39 & 67.59 & 61.08 & 60.52 \\
    \bottomrule
  \end{tabular}}%
  \ifdim\wd0>\linewidth\resizebox{\linewidth}{!}{\usebox0}\else\usebox0\fi
\end{table}

\providecommand{\tbd}{\textcolor{orange}{--}}
\providecommand{\dup}[1]{{\scriptsize\textcolor{ForestGreen}{($+$#1)}}}
\providecommand{\ddown}[1]{{\scriptsize\textcolor{BrickRed}{($-$#1)}}}

\begin{table*}[!htbp]
    \centering
    \caption{\textbf{Cross-center transfer.} AUC (\%). A linear head trained on \stanford or \ctrate is tested on the same center (ID) and on \extabdo or \extchest (OOD), with the taxonomy of~\cite{jolia}, used by the external sets (\cref{tab:datasets}). {\color{gray}Grey}: evaluation outside the pretraining region of the model. In parentheses: OOD AUC minus that of a head trained on the external set itself (\cref{tab:competitors}), {\color{ForestGreen}gain} or {\color{BrickRed}drop}. Avg EXT: mean of the two OOD columns. Bold: best per column, and values within one standard deviation of it. \textsuperscript{\textdagger}Pillar-0-Best: each Pillar model on its own region. \textsuperscript{*}Value reported in~\cite{jolia}.}
    \label{tab:cross_transfert}
    \footnotesize
    \renewcommand{\arraystretch}{0.92}
    \begin{tabular*}{\textwidth}{@{\extracolsep{\fill}} l c c c c c}
        \toprule
        & \multicolumn{2}{c}{\textbf{Abdomen}} & \multicolumn{2}{c}{\textbf{Chest}} & \\
        \cmidrule(lr){2-3} \cmidrule(lr){4-5}
        \textbf{Model}  & \stanford & \extabdo & \ctrate & \extchest & \textbf{Avg EXT} \\
        \midrule

        CT-CLIP~\cite{hamamci2024ctclip}
        & \textcolor{gray}{62.56} {\scriptsize \textcolor{gray}{$\pm$0.10}}
        & \textcolor{gray}{61.34} {\scriptsize \textcolor{gray}{$\pm$0.33}} \dup{0.6}
        & 75.52 {\scriptsize \textcolor{gray}{$\pm$0.23}}
        & 60.02 {\scriptsize \textcolor{gray}{$\pm$1.00}} \ddown{19.4}
        & \textcolor{gray}{60.68} \\
        fVLM~\cite{fvlm}
        & \textcolor{gray}{58.99\textsuperscript{*}}
        & \textcolor{gray}{56.59\textsuperscript{*}} \ddown{3.9}
        & 77.03\textsuperscript{*}
        & 59.12\textsuperscript{*} \ddown{20.5}
        & \textcolor{gray}{57.86} \\
        COLIPRI~\cite{colipri}
        & \textcolor{gray}{69.62\textsuperscript{*}}
        & \textcolor{gray}{67.44\textsuperscript{*}} \dup{1.0}
        & 82.91\textsuperscript{*}
        & 75.05\textsuperscript{*} \ddown{7.8}
        & \textcolor{gray}{71.25} \\
        ARC-CT~\cite{arcct}
        & \textcolor{gray}{68.20} {\scriptsize \textcolor{gray}{$\pm$0.29}}
        & \textcolor{gray}{63.77} {\scriptsize \textcolor{gray}{$\pm$0.58}} \ddown{0.8}
        & 84.16 {\scriptsize \textcolor{gray}{$\pm$0.33}}
        & 75.47 {\scriptsize \textcolor{gray}{$\pm$0.54}} \ddown{11.2}
        & \textcolor{gray}{69.62} \\

        Merlin~\cite{blankemeier2024merlin}
        & 79.45 {\scriptsize \textcolor{gray}{$\pm$0.03}}
        & 71.98 {\scriptsize \textcolor{gray}{$\pm$0.11}} \ddown{0.8}
        & \textcolor{gray}{75.91} {\scriptsize \textcolor{gray}{$\pm$0.25}}
        & \textcolor{gray}{63.63} {\scriptsize \textcolor{gray}{$\pm$0.24}} \ddown{17.9}
        & \textcolor{gray}{67.81} \\
        RADAR+~\cite{radar} %
        & 67.14 {\scriptsize \textcolor{gray}{$\pm$0.16}}
        & 65.62 {\scriptsize \textcolor{gray}{$\pm$0.24}} \dup{2.3}
        & \textcolor{gray}{66.85} {\scriptsize \textcolor{gray}{$\pm$0.23}}
        & \textcolor{gray}{57.88} {\scriptsize \textcolor{gray}{$\pm$0.91}} \ddown{21.7}
        & \textcolor{gray}{61.75} \\

        CT-FM~\cite{ctfm}
        & 63.92\textsuperscript{*}
        & 60.57\textsuperscript{*} \ddown{3.4}
        & 71.87\textsuperscript{*}
        & 63.01\textsuperscript{*} \ddown{18.0}
        & 61.79 \\
        SPECTRE~\cite{spectre}
        & 78.77\textsuperscript{*}
        & 71.90\textsuperscript{*} \ddown{3.5}
        & 82.25\textsuperscript{*}
        & 72.12\textsuperscript{*} \ddown{14.0}
        & 72.01 \\
        M3D-CLIP~\cite{m3d}
        & 64.01 {\scriptsize \textcolor{gray}{$\pm$0.27}}
        & 61.35 {\scriptsize \textcolor{gray}{$\pm$1.38}} \ddown{5.1}
        & 68.52 {\scriptsize \textcolor{gray}{$\pm$0.76}}
        & 56.45 {\scriptsize \textcolor{gray}{$\pm$1.34}} \ddown{22.4}
        & 58.90 \\
        CAPTOR~\cite{captor}
        & 78.85 {\scriptsize \textcolor{gray}{$\pm$0.08}}
        & 72.01 {\scriptsize \textcolor{gray}{$\pm$0.31}} \ddown{1.6}
        & 82.26 {\scriptsize \textcolor{gray}{$\pm$0.13}}
        & 68.03 {\scriptsize \textcolor{gray}{$\pm$0.75}} \ddown{15.7}
        & 70.02 \\
        RadFinder~\cite{radfinder}
        & 81.14 {\scriptsize \textcolor{gray}{$\pm$0.30}}
        & 77.82 {\scriptsize \textcolor{gray}{$\pm$0.74}} \dup{6.3}
        & 83.17 {\scriptsize \textcolor{gray}{$\pm$0.27}}
        & 78.79 {\scriptsize \textcolor{gray}{$\pm$0.12}} \ddown{2.8}
        & 78.31 \\
        FlexiCT~\cite{flexictvlm}
        & 71.09 {\scriptsize \textcolor{gray}{$\pm$0.47}}
        & 69.77 {\scriptsize \textcolor{gray}{$\pm$1.13}} \dup{1.7}
        & 76.32 {\scriptsize \textcolor{gray}{$\pm$0.44}}
        & 61.04 {\scriptsize \textcolor{gray}{$\pm$1.25}} \ddown{20.2}
        & 65.41 \\
        FlexiCT-VLM~\cite{flexictvlm}
        & 83.46 {\scriptsize \textcolor{gray}{$\pm$0.15}}
        & 81.08 {\scriptsize \textcolor{gray}{$\pm$0.25}} \dup{0.2}
        & 86.92 {\scriptsize \textcolor{gray}{$\pm$0.53}}
        & 81.39 {\scriptsize \textcolor{gray}{$\pm$0.46}} \ddown{9.8}
        & 81.23 \\
        Pillar-0-Best\textsuperscript{\textdagger}~\cite{pillar0}
        & 80.49\textsuperscript{*}
        & 76.07\textsuperscript{*} \ddown{2.2}
        & 84.71\textsuperscript{*}
        & 73.92\textsuperscript{*} \ddown{15.0}
        & 75.00 \\

        \midrule
        \Jolia - [CLS]~\cite{jolia}
        & 82.28 {\scriptsize \textcolor{gray}{$\pm$0.11}}
        & 74.64 {\scriptsize \textcolor{gray}{$\pm$0.24}}
        & 85.92 {\scriptsize \textcolor{gray}{$\pm$0.04}}
        & 71.11 {\scriptsize \textcolor{gray}{$\pm$0.31}}
        & 72.88 \\
        \Jolia~\cite{jolia}
        & 83.16 {\scriptsize \textcolor{gray}{$\pm$0.01}}
        & 75.02 {\scriptsize \textcolor{gray}{$\pm$0.04}} \ddown{2.6}
        & 86.89 {\scriptsize \textcolor{gray}{$\pm$0.02}}
        & 75.83 {\scriptsize \textcolor{gray}{$\pm$0.06}} \ddown{13.2}
        & 75.42 \\
        \midrule
        \ours{} - [CLS]
        & 85.97 {\scriptsize \textcolor{gray}{$\pm$0.13}}
        & 82.41 {\scriptsize \textcolor{gray}{$\pm$0.15}}
        & 88.89 {\scriptsize \textcolor{gray}{$\pm$0.18}}
        & 79.41 {\scriptsize \textcolor{gray}{$\pm$0.40}}
        & 80.91 \\
        \ours{}
        & \textbf{89.13} {\scriptsize \textcolor{gray}{$\pm$0.01}}
        & \textbf{85.52} {\scriptsize \textcolor{gray}{$\pm$0.03}} \ddown{0.1}
        & \textbf{90.31} {\scriptsize \textcolor{gray}{$\pm$0.06}}
        & \textbf{84.13} {\scriptsize \textcolor{gray}{$\pm$0.47}} \ddown{8.7}
        & \textbf{84.82} \\
        \bottomrule
    \end{tabular*}
\end{table*}

Following~\cite{jolia}, we train a linear head on the frozen encoder with the training split of each dataset (\cref{tab:datasets}), and report the area under the ROC curve (AUC) averaged over the four test sets.
For every model, we report the best of several readouts (\cref{app:findings}). For \JoliaTwo, as for \Jolia~\cite{jolia}, the best readout combines the global token with the organ token (\cref{app:readout}).

\cref{fig:competitors} places every model on both tasks.
\JoliaTwo is first on each of the four test sets, with a mean AUC of 88.9 against 85.9 for the second best, FlexiCT-VLM (\cref{tab:competitors}).
The lead does not depend on the readout. With the [CLS] token alone, \JoliaTwo is still first (86.7 against 85.9, \cref{tab:competitors}).
The lead is larger on abdominal CT (4.1 points) than on chest CT (1.8), as the organ tokens add 3.2 points on the former and 1.2 on the latter (\cref{app:readout}).

No model sees \extabdo and \extchest in pretraining, and \JoliaTwo leads on both (85.7 and 92.9, against 80.9 and 91.2 for FlexiCT-VLM).
It also leads when the linear head is trained on another center, \stanford or \ctrate. The mean AUC on the two external sets is then 84.8, against 81.2 for FlexiCT-VLM and 75.4 for \Jolia (\cref{tab:cross_transfert}).
The lead holds on every test set, readout and center.

\subsection{Organ and lesion segmentation}
\label{sec:res_dense}
\begin{table*}[t]   %
  \centering
  \scriptsize
  \caption{\textbf{Segmentation with a frozen encoder, comparison with external CT foundation models.} Dice (\%) on the 13 datasets.
    MSD: MSD Liver; Liv.\ v.: liver vessels and lesions; WAW: WAW HCC; TS liv.: TotalSegmentator liver lesions; ImCAS: ImageCAS.
    VISTA3D and TotalSegmentator are supervised segmentation models. Bold: best per column.}
  \label{tab:dense}
  \setlength{\tabcolsep}{2.6pt}
  \begin{adjustbox}{max width=\linewidth, max totalheight=0.46\textheight}   %
  \begin{tabular}{@{}lcccccccccccccc@{}}
    \toprule
    Model & MSD & AMOS & TS & Lung & Colon & Liv.\ v. & KiTS & CRLM & WAW & TS liv. & LIDC & ImCAS & SegTHOR & Mean \\
    \midrule
    \JoliaTwo & 77.87 & 86.79 & \textbf{86.32} & \textbf{74.22} & 49.12 & 59.75 & 80.79 & 69.18 & 84.80 & 73.98 & 59.82 & 60.01 & 85.96 & 72.97 \\
    \Jolia & 74.00 & 83.88 & 78.99 & 65.30 & 28.88 & 55.32 & 77.93 & 65.35 & 76.33 & 64.64 & 47.56 & 41.44 & 85.06 & 64.98 \\
    \cmidrule{1-15}   %
    VISTA3D & \textbf{81.84} & \textbf{88.09} & 77.85 & 71.98 & 48.27 & \textbf{65.41} & 79.34 & \textbf{72.18} & 86.20 & \textbf{79.31} & \textbf{63.26} & 65.75 & 87.37 & \textbf{74.37} \\
    TotalSegmentator & 78.06 & 87.74 & 78.23 & 65.95 & 42.07 & 61.39 & 80.06 & 71.19 & 85.52 & 76.44 & 59.52 & \textbf{65.86} & \textbf{87.75} & 72.29 \\
    Coralbay & 77.40 & 87.18 & 79.00 & 70.58 & 40.85 & 61.30 & 80.67 & 71.62 & 85.92 & 72.55 & 58.15 & 62.92 & 87.37 & 71.96 \\
    Pillar-0-Best & 78.92 & 87.09 & 76.13 & 60.64 & 45.75 & 59.71 & 80.73 & 69.98 & 85.49 & 75.27 & 56.65 & 65.02 & 87.32 & 71.44 \\
    \VoCo & 77.95 & 87.07 & 75.22 & 64.89 & 45.80 & 61.74 & \textbf{81.40} & 70.76 & 85.43 & 75.75 & 47.37 & 63.77 & 87.21 & 71.10 \\
    SuPreM & 79.07 & 86.79 & 73.77 & 66.94 & 47.01 & 58.33 & 80.25 & 70.44 & 81.34 & 74.73 & 50.22 & 63.56 & 85.56 & 70.62 \\
    CAPTOR & 78.59 & 85.72 & 72.82 & 62.92 & 41.22 & 58.95 & 78.83 & 70.37 & 84.61 & 71.84 & 61.63 & 62.61 & 86.37 & 70.50 \\
    Merlin & 77.66 & 86.10 & 72.08 & 61.98 & 39.32 & 60.66 & 78.01 & 71.12 & 85.61 & 71.00 & 61.94 & 63.33 & 86.85 & 70.44 \\
    RADAR+ & 79.95 & 85.10 & 70.28 & 62.88 & 35.62 & 60.70 & 78.25 & 71.05 & \textbf{86.29} & 74.75 & 53.91 & 63.90 & 86.92 & 69.97 \\
    FlexiCT & 79.31 & 85.10 & 76.39 & 63.87 & 47.32 & 58.71 & 76.16 & 68.46 & 82.40 & 68.05 & 51.90 & 54.94 & 86.64 & 69.17 \\
    FlexiCT-VLM & 79.58 & 85.10 & 76.24 & 64.15 & \textbf{49.41} & 57.70 & 75.26 & 68.23 & 83.11 & 68.66 & 52.27 & 52.75 & 86.69 & 69.17 \\
    SPECTRE & 78.28 & 82.72 & 67.41 & 66.65 & 44.95 & 59.84 & 74.06 & 69.83 & 78.44 & 69.08 & 54.69 & 62.25 & 86.11 & 68.79 \\
    \CuriaMAE & 79.74 & 82.37 & 66.87 & 65.83 & 27.61 & 56.35 & 80.67 & 68.75 & 72.81 & 66.22 & 61.60 & 64.12 & 85.94 & 67.61 \\
    ARC-CT & 72.81 & 82.58 & 64.48 & 68.43 & 30.31 & 55.29 & 69.74 & 68.50 & 74.65 & 67.40 & 62.74 & 62.60 & 84.67 & 66.48 \\
    TAP-CT & 75.56 & 79.94 & 65.21 & 64.14 & 36.39 & 56.98 & 68.23 & 69.31 & 74.88 & 62.26 & 51.51 & 62.60 & 86.65 & 65.67 \\
    CT-FM & 74.53 & 83.39 & 69.00 & 39.60 & 25.33 & 55.77 & 72.57 & 68.98 & 79.63 & 73.13 & 58.61 & 61.67 & 85.34 & 65.20 \\
    Curia 2 L & 75.65 & 82.05 & 72.31 & 56.70 & 37.98 & 54.96 & 70.70 & 65.56 & 72.73 & 59.81 & 34.82 & 52.24 & 85.47 & 63.15 \\
    COLIPRI & 69.45 & 73.52 & 58.00 & 66.24 & 24.26 & 50.70 & 61.48 & 64.69 & 60.63 & 53.46 & 61.89 & 64.25 & 83.67 & 60.94 \\
    Curia L & 73.32 & 80.54 & 69.46 & 54.30 & 30.80 & 53.11 & 67.44 & 65.71 & 73.58 & 58.49 & 32.08 & 46.37 & 84.85 & 60.77 \\
    nnFoundation ViT & 65.14 & 67.87 & 53.91 & 66.59 & 27.15 & 49.65 & 60.30 & 58.09 & 51.41 & 51.35 & 55.27 & 62.51 & 81.92 & 57.78 \\
    RadFinder & 71.96 & 76.88 & 61.03 & 53.70 & 38.78 & 46.93 & 68.03 & 57.65 & 76.34 & 57.99 & 22.26 & 36.85 & 82.49 & 57.76 \\
    NV-Reason-CT & 67.87 & 73.53 & 59.60 & 61.65 & 25.40 & 49.22 & 66.29 & 58.18 & 69.67 & 53.04 & 39.63 & 32.40 & 81.43 & 56.76 \\
    MedVTok & 68.44 & 70.65 & 50.83 & 55.47 & 23.70 & 48.65 & 63.10 & 62.36 & 67.91 & 51.95 & 36.39 & 50.96 & 80.26 & 56.21 \\
    nnFoundation CNN & 61.20 & 74.04 & 53.75 & 56.35 & 7.04 & 38.75 & 53.95 & 57.34 & 47.57 & 41.52 & 50.52 & 63.68 & 81.87 & 52.89 \\
    EVLF-FM & 63.44 & 66.83 & 45.46 & 45.17 & 23.23 & 46.35 & 54.38 & 61.39 & 56.85 & 45.88 & 37.38 & 50.72 & 79.16 & 52.02 \\
    M3D-CLIP & 47.87 & 29.91 & 18.86 & 8.29 & 2.96 & 16.41 & 27.47 & 34.98 & 40.77 & 17.80 & 3.41 & 8.47 & 51.51 & 23.75 \\
    RadFM & 38.79 & 33.93 & 24.95 & 3.91 & 1.78 & 22.28 & 18.48 & 38.59 & 38.24 & 14.03 & 1.23 & 16.29 & 44.24 & 22.83 \\
    fVLM & 44.47 & 31.07 & 18.48 & 8.15 & 2.66 & 12.26 & 20.40 & 28.88 & 40.03 & 14.88 & 2.15 & 11.82 & 51.62 & 22.07 \\
    CT-CLIP & 18.42 & 9.31 & 1.97 & 2.71 & 1.48 & 3.83 & 5.01 & 16.82 & 22.08 & 3.01 & 0.15 & 7.75 & 25.31 & 9.07 \\
    \bottomrule
  \end{tabular}
  \end{adjustbox}
\end{table*}
We train one decoder per dataset, with a frozen and a fine-tuned encoder, and report the mean Dice over the 13 datasets, with all the classes of each dataset.
The decoder is pyramidal for hierarchical encoders, including ours, and single-scale for ViT encoders (\cref{app:dense}, \cref{tab:seg_training}).

With a frozen encoder, \JoliaTwo has a mean Dice of 73.0 (\cref{tab:dense}). Only VISTA3D, a supervised segmentation model, is higher (74.4), and \JoliaTwo is ahead of TotalSegmentator (72.3), of the best image-only model, Coralbay (72.0), and of the best vision-language model, Pillar-0-Best (71.4).
VISTA3D and Coralbay pretrain on several of our segmentation datasets (\cref{tab:methods}).
On both tasks together, no external model is better (\cref{fig:competitors}), and VISTA3D is 19.3 points behind on findings.
This result does not depend on the datasets that our masks cover. Without TotalSegmentator and AMOS, whose organs are TotalSegmentator classes, \JoliaTwo stays on par with TotalSegmentator (70.5 against 70.4) and ahead of Pillar-0-Best (69.6).
It has the best Dice on TotalSegmentator and on the lung tumours, and is second on the colon tumours and on KiTS, all classes together (\cref{tab:dense}).

With a fine-tuned encoder, \JoliaTwo ranks sixth (74.3), 1.2 points behind VISTA3D (75.5, \cref{app:dense}).
We target a frozen encoder shared across tasks, and fine-tuning brings the models within a narrower range.
A single frozen encoder comes within 2 points of a ResEnc-L~\cite{isensee2024nnunetrevisited} trained from scratch (73.0 against 74.9).

\subsection{Zero-shot classification and retrieval}
\label{sec:zs}

\begin{figure}[t]
  \centering
  \includegraphics[width=\columnwidth]{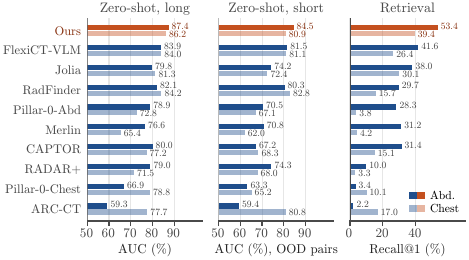}
  \caption{\textbf{Zero-shot classification and retrieval.} The ten best models by mean rank on zero-shot long and retrieval. Zero-shot short uses the 29 out-of-distribution (OOD) pairs (\cref{tab:zs_summary}). Merlin, Pillar-0-Abd and RADAR+ are trained on abdominal CT only, Pillar-0-Chest and ARC-CT on chest CT only (\cref{app:global}).}
  \label{fig:global_summary}
\end{figure}

\begin{table*}[t]
\centering
\footnotesize
\caption{\textbf{Zero-shot finding classification.} AUC (\%) on the four test sets. S-ID: the 8 short prompt pairs seen in training by \JoliaTwo (aggregated); S-OOD: the 29 other pairs (\cref{app:zs_short}); Long: prototypes from $r=50$ reports. Grey: model not trained on that region, not ranked; its mean covers its own region only. \JoliaTwo, queries only and [CLS] only: the organ queries or the global token alone (\JoliaTwo: both). Best in bold, second underlined. \extabdo and \extchest: external sets, unseen in pretraining.}
\label{tab:zs_summary}
\setlength{\tabcolsep}{2.4pt}
\begin{adjustbox}{max width=\linewidth}   %
\begin{tabular}{l ccc@{\hspace{5pt}}ccc@{\hspace{5pt}}ccc@{\hspace{5pt}}ccc@{\hspace{5pt}}ccc}
\toprule
 & \multicolumn{6}{c}{\textbf{Abdomen}} & \multicolumn{6}{c}{\textbf{Chest}} & \multicolumn{3}{c}{} \\
\cmidrule(lr){2-7}\cmidrule(lr){8-13}
 & \multicolumn{3}{c}{\stanford} & \multicolumn{3}{c}{\extabdo} & \multicolumn{3}{c}{CT-RATE} & \multicolumn{3}{c}{\extchest} & \multicolumn{3}{c}{\textbf{Average}} \\
\cmidrule(lr){2-4}\cmidrule(lr){5-7}\cmidrule(lr){8-10}\cmidrule(lr){11-13}\cmidrule(lr){14-16}
\textbf{Model} & {\scriptsize S-ID} & {\scriptsize S-OOD} & {\scriptsize Long} & {\scriptsize S-ID} & {\scriptsize S-OOD} & {\scriptsize Long} & {\scriptsize S-ID} & {\scriptsize S-OOD} & {\scriptsize Long} & {\scriptsize S-ID} & {\scriptsize S-OOD} & {\scriptsize Long} & {\scriptsize S-ID} & {\scriptsize S-OOD} & {\scriptsize Long} \\
\midrule
CT-CLIP~\cite{hamamci2024ctclip} & \textcolor[gray]{0.5}{60.6} & \textcolor[gray]{0.5}{60.5} & \textcolor[gray]{0.5}{60.4} & \textcolor[gray]{0.5}{51.7} & \textcolor[gray]{0.5}{51.7} & \textcolor[gray]{0.5}{53.8} & 71.1 & 71.8 & 74.2 & 59.3 & 61.6 & 70.3 & 65.2 & 66.7 & 72.3 \\
fVLM~\cite{fvlm} & \textcolor[gray]{0.5}{59.5} & \textcolor[gray]{0.5}{58.7} & \textcolor[gray]{0.5}{57.2} & \textcolor[gray]{0.5}{55.7} & \textcolor[gray]{0.5}{56.2} & \textcolor[gray]{0.5}{55.6} & 68.3 & 69.9 & 68.5 & 64.3 & 69.3 & 54.2 & 66.3 & 69.6 & 61.3 \\
COLIPRI~\cite{colipri} & \textcolor[gray]{0.5}{73.0} & \textcolor[gray]{0.5}{73.5} & \textcolor[gray]{0.5}{75.0} & \textcolor[gray]{0.5}{61.2} & \textcolor[gray]{0.5}{61.3} & \textcolor[gray]{0.5}{63.1} & 77.2 & 77.5 & 79.0 & 72.5 & 72.6 & 71.2 & 74.8 & 75.1 & 75.1 \\
SPECTRE~\cite{spectre} & 59.7 & 59.0 & 79.7 & 60.5 & 61.0 & 71.3 & 57.0 & 58.3 & 72.5 & 61.0 & 62.3 & 80.2 & 59.5 & 60.2 & 75.9 \\
Merlin~\cite{blankemeier2024merlin} & 77.7 & 78.0 & 84.3 & 63.5 & 63.6 & 68.9 & \textcolor[gray]{0.5}{65.0} & \textcolor[gray]{0.5}{65.7} & \textcolor[gray]{0.5}{70.3} & \textcolor[gray]{0.5}{57.1} & \textcolor[gray]{0.5}{58.4} & \textcolor[gray]{0.5}{60.5} & 70.6 & 70.8 & 76.6 \\
Pillar-0-Abd~\cite{pillar0} & 71.1 & 74.9 & 81.6 & 63.2 & 66.0 & 76.3 & \textcolor[gray]{0.5}{63.3} & \textcolor[gray]{0.5}{64.9} & \textcolor[gray]{0.5}{66.8} & \textcolor[gray]{0.5}{62.2} & \textcolor[gray]{0.5}{69.2} & \textcolor[gray]{0.5}{78.8} & 67.2 & 70.5 & 79.0 \\
Pillar-0-Chest~\cite{pillar0} & \textcolor[gray]{0.5}{67.2} & \textcolor[gray]{0.5}{68.6} & \textcolor[gray]{0.5}{71.9} & \textcolor[gray]{0.5}{56.3} & \textcolor[gray]{0.5}{58.0} & \textcolor[gray]{0.5}{61.9} & 70.9 & 72.6 & 78.6 & 47.8 & 57.8 & 79.1 & 59.3 & 65.2 & 78.9 \\
FlexiCT-VLM~\cite{flexictvlm} & 86.5 & \underline{86.5} & 86.6 & 76.2 & 76.6 & 81.3 & 79.5 & 81.5 & 82.6 & 80.5 & 80.7 & 85.4 & 80.7 & 81.4 & 84.0 \\
ARC-CT~\cite{arcct} & \textcolor[gray]{0.5}{64.2} & \textcolor[gray]{0.5}{64.4} & \textcolor[gray]{0.5}{62.6} & \textcolor[gray]{0.5}{53.5} & \textcolor[gray]{0.5}{54.4} & \textcolor[gray]{0.5}{56.0} & \textbf{84.1} & \textbf{84.3} & 77.2 & 78.0 & 77.4 & 78.1 & 81.1 & 80.8 & 77.7 \\
CAPTOR~\cite{captor} & 70.2 & 72.7 & 85.3 & 58.4 & 61.7 & 74.8 & 63.5 & 64.5 & 76.2 & 67.2 & 72.2 & 78.3 & 64.8 & 67.7 & 78.7 \\
RadFinder~\cite{radfinder} & 81.3 & 82.6 & 82.9 & 77.8 & 78.1 & 81.3 & 82.2 & 82.9 & 81.4 & \underline{82.8} & \underline{82.7} & 86.9 & 81.0 & 81.5 & 83.1 \\
RADAR+~\cite{radar} & 77.3 & 80.2 & 85.3 & 67.7 & 68.4 & 72.8 & \textcolor[gray]{0.5}{71.4} & \textcolor[gray]{0.5}{72.3} & \textcolor[gray]{0.5}{72.8} & \textcolor[gray]{0.5}{61.6} & \textcolor[gray]{0.5}{63.6} & \textcolor[gray]{0.5}{70.3} & 72.5 & 74.3 & 79.0 \\
\Jolia~\cite{jolia} & 78.2 & 80.9 & 85.4 & 62.7 & 67.6 & 74.2 & 74.8 & 76.6 & 82.4 & 63.9 & 68.3 & 80.3 & 69.9 & 73.3 & 80.6 \\
\midrule
\textbf{\JoliaTwo} & \textbf{88.1} & \textbf{88.2} & \textbf{89.2} & \textbf{81.4} & \textbf{80.8} & \textbf{85.6} & 80.9 & 80.9 & \textbf{84.8} & 82.6 & 80.9 & \underline{87.6} & \underline{83.2} & \underline{82.7} & \textbf{86.8} \\
\JoliaTwo, queries only & 83.8 & 85.4 & \underline{88.5} & 75.5 & 76.9 & 82.3 & 72.4 & 74.2 & 82.3 & 68.1 & 68.5 & 78.5 & 75.0 & 76.2 & 82.9 \\
\JoliaTwo, [CLS] only & \underline{87.1} & \underline{86.5} & 87.2 & \underline{80.2} & \underline{79.4} & \underline{84.0} & \underline{83.5} & \underline{83.2} & \underline{84.0} & \textbf{85.1} & \textbf{84.3} & \textbf{88.3} & \textbf{84.0} & \textbf{83.3} & \underline{85.9} \\
\bottomrule
\end{tabular}
\end{adjustbox}
\end{table*}
Zero-shot classification scores each finding by the difference between the cosine similarities of the image with a positive and a negative text.
In \emph{zero-shot long}, the two texts are prototypes, the mean embeddings of $r=50$ reports of positive studies and of 50 reports of negative studies.
In \emph{zero-shot short}, the texts are 29 pairs of positive and negative prompts, and we average the similarities over the pairs. These pairs are out-of-distribution (OOD) for \JoliaTwo, whose $\mathcal{L}_{\text{OSL}}$ trains on eight other pairs (\cref{app:zs_short}).

On zero-shot long, \JoliaTwo is the best model on all four test sets (\cref{fig:global_summary,tab:zs_summary}).
On zero-shot short, \JoliaTwo has the best mean (82.7 against 81.5 for RadFinder), but ARC-CT and RadFinder lead on \ctrate and \extchest respectively, where the organ tokens do not help (\cref{tab:zs_summary}, \cref{app:global}).

Image-text retrieval matches each volume with the Findings section of its report within disjoint pools of 100 test studies. We report the Recall@1 averaged over image-to-text and text-to-image.
The mean Recall@1 of \JoliaTwo is 46.4, against 34.0 for \Jolia and FlexiCT-VLM.
The organ tokens add to the global token in retrieval and in zero-shot long (\cref{app:readout}).
The lead extends to tasks with no training on top of the encoder, except in zero-shot short on chest CT.

\subsection{Qualitative results}
\label{sec:res_qualitative}
\begin{figure*}[t]
  \centering
  \includegraphics[width=\textwidth]{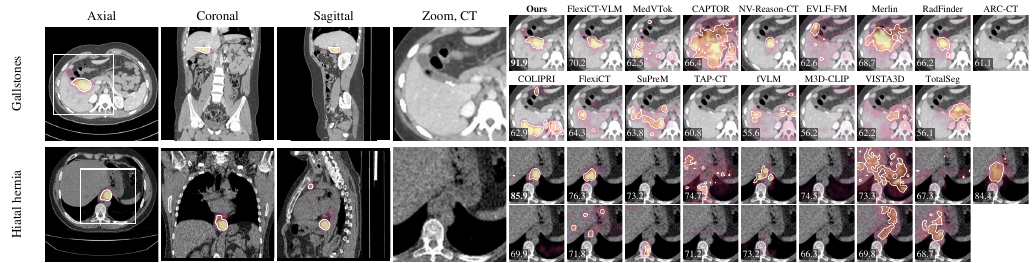}\\[2pt]
  \includegraphics[width=\textwidth]{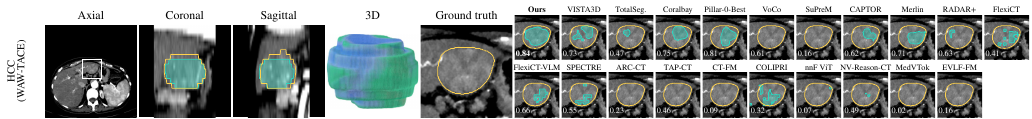}\\[2pt]
  \includegraphics[width=\textwidth]{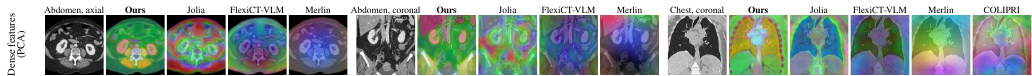}
  \caption{\textbf{Selected cases.} Top, findings: gradient $\times$ input of the finding logit, each model on its own scale, white line at 50\% of the maximum; number: AUC of the model on this finding over the test set, not a score of the case. Middle, a lesion, frozen encoder: prediction in teal, ground truth in yellow, 3D Dice of the case. Left: \JoliaTwo (``Ours'') in three planes, then the CT of the white box, shown in every tile, or its 3D surface (blue inside, green within 1\,mm, red outside the ground truth). Bottom: PCA of patch tokens (PC2--4 as RGB).}
  \label{fig:qualitative_main}
\end{figure*}

\cref{fig:qualitative_main} shows a few selected cases.
The top rows show attribution maps, the voxels that contribute most to each model's score for a finding (\cref{app:qualitative}).
The map of \JoliaTwo falls on the calcified gallstones. Most external maps are more dispersed.
\JoliaTwo and a few other models localise the hiatal hernia.
In the middle row, on a hepatocellular carcinoma, \JoliaTwo has the highest Dice and many models miss most of the tumour.
In the bottom row, a PCA of the patch tokens of \JoliaTwo gives each organ a uniform colour and a sharp edge, unlike the other models (\cref{app:pca}).

\section{Ablation study}
\label{sec:ablations}

To reduce the compute cost, ablations use \HieraCT at 1.5\,mm, with the Dice of lesion datasets on the lesion class only, so values are lower than in \cref{sec:results}. Run lengths are in \cref{app:ablation_setup}.

\subsection{Pretraining methods}
\label{sec:res_pretraining}
Report alignment and image-only pretraining trade one task for the other (\cref{fig:controlled_comparison}).
\Conquer has a high AUC and a low Dice. Image-only methods~\cite{he2022mae,oquab2024dinov2,zhou2022ibot,assran2023ijepa,voco,venkataramanan2025franca} have a higher Dice at best and a much lower AUC: \MAE has the best Dice, with an AUC 11 points below \Conquer, and self-distillation (\DINOvTwo, \iBOT, \Franca) is lowest on both.
nnFoundation~\cite{nnfoundation} also finds \MAE ahead of \DINOvTwo in 3D, and DINO-based encoders trained by their authors on 11k to 150k volumes stay below \Conquer (\cref{tab:competitors}).

Adding image-only self-supervision to \Conquer does not move the front: \MAE or \MAEBT lower the AUC by 4 points, and \iBOT with \DenseBT changes both scores by less than 0.2.

\Method moves the front. It raises the AUC by 2.5 and the Dice by 2.7 over \Conquer, and is the only configuration beyond the front of the others.
Its Dice nearly reaches that of \MAE, with an AUC 13 points higher, so one encoder serves both tasks (\cref{app:ablation_detail}).

\subsection{Backbones}
\label{sec:res_pareto}
The gain is consistent across backbones.
On five backbones, convolutional (ResNet~\cite{he2016resnet}, ResEnc-L~\cite{isensee2024nnunetrevisited}) and transformer (\HieraCT, Atlas~\cite{agrawal2025atlas}, Swin~\cite{liu2021swin}), \Method lies beyond the Pareto front of \CLIP, \Conquer, \MAE and \MAEBT, with average gains over \Conquer of 1.8 AUC and 3.0 Dice (\cref{fig:arch_ablation}).
The Dice gain holds on all five, the AUC gain on four; on Atlas it is 0.3, and two of the four test sets go down (\cref{tab:ablation_detail_findings}).
\MAEBT is the recipe of Curia-MAE~\cite{curiamae}, retrained on each backbone.
We keep \HieraCT, with the best AUC and a near-best Dice.

\begin{figure*}[t]
  \centering
  \includegraphics[width=\textwidth]{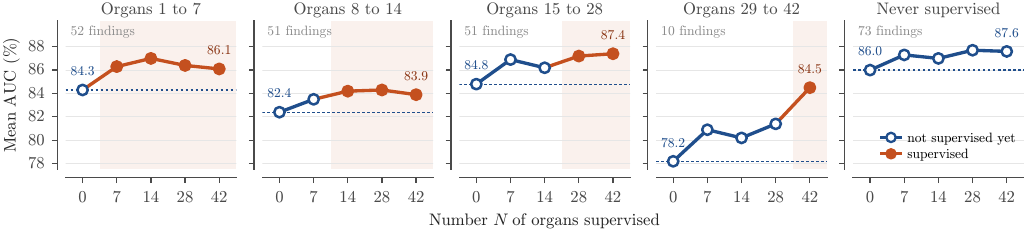}
  \caption{\textbf{Query grounding helps beyond the supervised organs.} Each panel follows the findings of one group of organs as the number $N$ of supervised organs grows. Blue: the organs of the group have no mask yet; orange: they have one. The dotted line is the value without query grounding. Mean linear-probe AUC over the four test sets (\HieraCT, 30k steps).}
  \label{fig:cov_nb_organs}
\end{figure*}

\begin{figure}[!ht]
  \centering
  \includegraphics[width=\columnwidth]{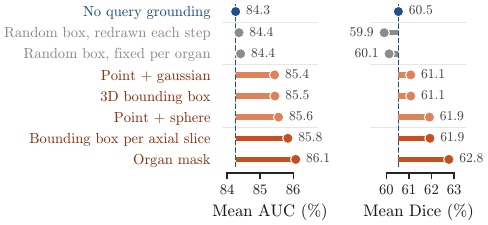}
  \caption{\textbf{The target of query grounding must sit on the organ.} Mean AUC over the four test sets (left) and mean Dice over the 13 datasets (right), $N=42$ (\HieraCT, 30k steps).}
  \label{fig:cov_target}
\end{figure}

\subsection{Incremental ablation}
\label{sec:res_ladder}
\begin{table}[!h]
  \centering
  \footnotesize
  \setlength{\tabcolsep}{2.5pt}
  \caption{\textbf{Incremental ablation.} Each row adds a component to the previous one (+) or replaces its backbone ($\rightarrow$). Findings: mean AUC. Seg.: mean Dice over the 13 datasets, lesion class only for the lesion datasets. ZS short, ZS long: AUC in zero-shot short and zero-shot long (\cref{sec:zs}). Retrieval: Recall@1. Findings, ZS and retrieval averaged over the four test sets. All with a frozen encoder, in \%.}
  \label{tab:ablation}
  \resizebox{\columnwidth}{!}{%
  \begin{tabular}{@{}llccccc@{}}
    \toprule
     & Model & \Findings & \Seg & \ZSshort & \ZSlong & \Retrieval \\
    \midrule
    \textit{Baseline} & \CLIP (Atlas) & 83.14 & 60.90 & 72.90 & 80.63 & 28.13 \\
     & + \Conquer & 84.22 & 60.60 & 72.78 & 80.79 & 28.30 \\
    \midrule
    \textit{Architecture} & $\rightarrow$ Hiera & 83.15 & 60.35 & 69.83 & 79.35 & 26.06 \\
     & $\rightarrow$ \HieraCT & 84.88 & 61.82 & 74.08 & 82.12 & 31.53 \\
    \midrule
    \textit{\Method} & + finding labels & 85.27 & 60.90 & 70.41 & 80.02 & 25.92 \\
     & + \SSL & 85.43 & 61.00 & 72.58 & 81.76 & 32.20 \\
     & + \Coverage & 86.68 & 62.26 & 73.88 & 83.92 & 37.05 \\
     & + patch-level seg. & 87.42 & 64.49 & 76.70 & 84.31 & 38.41 \\
    \midrule
    \textit{\JoliaTwo} & + \osl & 87.03 & 65.15 & 79.90 & 84.72 & 37.89 \\
     & + \OneMM & 88.48 & 69.41 & 82.60 & 86.28 & 44.28 \\
     & + cooldown & \textbf{88.86} & \textbf{69.56} & \textbf{82.70} & \textbf{86.81} & \textbf{46.41} \\
    \bottomrule
  \end{tabular}}
\end{table}

\cref{tab:ablation} adds the components one at a time from \CLIP.
\Conquer raises the AUC but not the Dice and \HieraCT raises both tasks. Finding-label supervision raises the AUC by 0.4, mostly on the external sets (\cref{tab:ablation_detail_findings}), but lowers the Dice by 0.9 and the zero-shot and retrieval scores.
\SSL brings little (+0.2 AUC, +0.1 Dice).
Query grounding and patch-level segmentation together raise the AUC by 2.0 and the Dice by 3.5.
Patch-level segmentation alone adds only 0.4 Dice to \Conquer, but 2.2 Dice after query grounding, so the head does not suffice alone (\cref{app:ablation_setup}).
1\,mm is part of the recipe and brings the largest jump in segmentation (+4.3 Dice); its gain adds to that of the masks (\cref{app:ablation_setup}).
$\mathcal{L}_{\text{OSL}}$ costs 0.4 AUC but adds 0.7 Dice and 3.2 AUC in zero-shot short, so we keep it.
The cooldown adds 0.4 AUC and 2.1 Recall@1.

\subsection{Analysis of query grounding}
\label{sec:res_mask}
\begin{figure*}[!p]
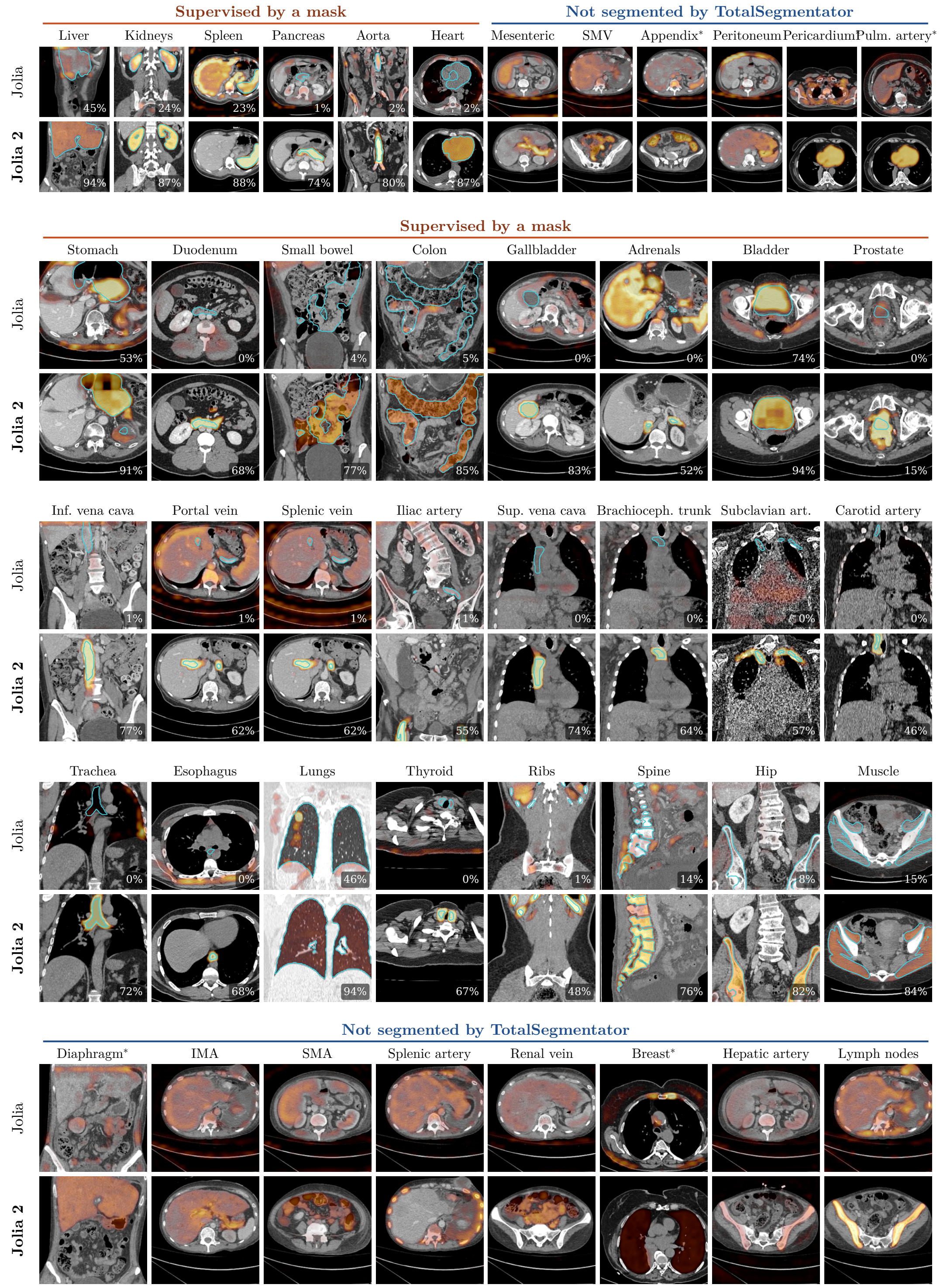

  \centering
  \providecommand{\attnGroupUnmapped}{}\renewcommand{\attnGroupUnmapped}{Not segmented by TotalSegmentator}
  \scalebox{0.84}{\input{figures/attention_maps_body}}\\[6pt]
  \scalebox{0.84}{\input{figures/attention_maps_more_body}}
  \caption{\textbf{Where the organ queries look.}
    Attention of each query relative to a uniform map, on a log scale from 1$\times$ (transparent) to 100$\times$ (saturated), each model on its own best slice.
    Cyan: TotalSegmentator mask; number: share of the attention inside it.
    Top: 12 queries, on the Stanford or CT-RATE volume at the median for \JoliaTwo. Bottom: 24 other organs with a mask, and 8 other organs that TotalSegmentator does not segment, which share one abdominal and one chest volume.
    $^{\ast}$: query supervised by the mask of a neighbouring structure (\cref{app:coverage}). The other queries of this group are never supervised.
    These queries are among those that \JoliaTwo concentrates more than \Jolia (\cref{app:localisation}). Without any mask, a query can land on an unrelated structure, such as the iliac bones for the hepatic artery and the lymph nodes.
    SMV: superior mesenteric vein; IMA, SMA: inferior, superior mesenteric artery.}
  \label{fig:attention_maps}
\end{figure*}
Without query grounding, the attention maximum of an organ query of \Jolia falls inside its organ in 5.2\% of the volumes. Query grounding, whose loss optimises this attention, brings it to 91.5\% in \JoliaTwo, and its attention fills the organ (\cref{fig:attention_maps}, \cref{app:localisation}). The runs below add query grounding alone to \Conquer on \HieraCT, and vary the supervised organs and the target (\cref{app:mask_supervision}).

\noindent\textbf{Localisation.}
\cref{fig:attention_maps} shows where each organ query looks. The attention of \Jolia is diffuse and mostly outside the organ, whereas that of \JoliaTwo is much more precise.
Queries that never receive a mask are less reliable, but still better localised than those of \Jolia.

\noindent\textbf{Unsupervised organs.}

The gain of query grounding is not limited to the supervised organs.
As the number $N$ of supervised organs grows (\cref{fig:cov_nb_organs}), the findings of an organ gain before that organ is supervised: the 73 findings of organs that never receive a mask gain 1.6 AUC from $N=0$ to 42.
Supervising an organ adds a further gain, mostly for the organs supervised last.

\noindent\textbf{Target.}
The target must sit on the organ, but it can be coarse (\cref{fig:cov_target}).
Random boxes bring at most 0.2 AUC, whereas a point or a 3D box recovers about two thirds of the gain of the organ mask, which remains the best target (\cref{app:mask_supervision}).
On the Dice, random boxes lower it, and coarse targets recover 0.6 to 1.4 of the 2.2 points of the mask.
Each token scale gives a similar gain (1.8 to 2.1 AUC). Supervising only the small, medium or large organs gives 1.3 to 1.5, below the 1.8 of all 42 organs, so the gain depends more on the number of supervised organs than on their size (\cref{tab:cov_scale_size}).

\section{Conclusion}
\label{sec:conclusion}

We presented \Method, a pretraining method for CT foundation models that leverages images, reports and organ masks. %
The masks guide the organ queries and train the patch tokens through a head discarded after pretraining.
A coarse signal is enough: a point or a box per organ recovers most of the findings gain of a full mask, and organs that never receive a mask gain too.
\Method moves the Pareto front on five backbones. Trained with it on public data, \JoliaTwo ranks first among about 30 CT models on finding classification, zero-shot classification and retrieval, and second on frozen segmentation. 
More broadly, our results suggest that the choice between a classification and a segmentation encoder is not necessary: with the right spatial signal, one frozen encoder can serve both finding classification and segmentation.
The masks come from an off-the-shelf model and add no annotation cost, so \Method can scale with the data. A natural next step is to train on larger and more diverse corpora, and to extend \Method to other modalities such as MRI and PET.
Weights will be released shortly.

\section*{Acknowledgements}
This work was granted access to the HPC resources of IDRIS (Jean Zay) under the allocation A0201017510 made by GENCI.
We acknowledge the EuroHPC Joint Undertaking for awarding the project ID EHPC-EXT-2025E02-048 access to JUPITER Booster hosted by JSC, Germany.
\clearpage

\makeatletter\global\@colht\textheight\global\@colroom\textheight\global\vsize\textheight\makeatother

{
    \small
    \bibliographystyle{unsrtnat}
    \bibliography{main}
}

\clearpage
\onecolumn
\renewcommand{\floatpagefraction}{0.5}\renewcommand{\topfraction}{0.9}\renewcommand{\bottomfraction}{0.5}\renewcommand{\textfraction}{0.08}
\appendix
\begin{center}
  {\Large\textbf{Appendix}}\\[1em]
\end{center}

\section{Data}
\label{app:data}
\begin{table}[!htbp]
  \centering
  \footnotesize
  \setlength{\tabcolsep}{3pt}
  \caption{\textbf{Datasets.} CT volumes used for pretraining (Pretrain) and for finding classification (Train: linear-probe training, Val: choice of its learning rate, Test: evaluation), and the size of each finding taxonomy. \ctrate releases no test split, so its Test set is its public validation split, in every table; with no separate validation split, the learning rate of its linear probe is selected on this set, for every model. \ctrate and \stanford use their native taxonomies; \inspect and the external sets use the taxonomy of~\cite{jolia}. \textsuperscript{$\dagger$}External: institutions absent from the pretraining data.}
  \label{tab:datasets}
    \begin{adjustbox}{max width=\linewidth}   %
    \begin{tabular}{@{}llrrrrr@{}}
    \toprule
    Dataset & Region & Pretrain & Train & Val & Test & Findings \\
    \midrule
    \ctrate~\cite{hamamci2024ctclip}        & Chest   & 24{,}128 & 24{,}128 & --      & 1{,}564 & 18  \\
    \inspect~\cite{inspect2023}             & Chest   & 18{,}937 & --       & --      & --      & --  \\
    \stanford~\cite{blankemeier2024merlin}  & Abdomen & 15{,}314 & 15{,}314 & 5{,}055 & 5{,}137 & 30  \\
    \extchest\textsuperscript{$\dagger$}    & Chest   & --       & 30{,}873 & 3{,}815 & 3{,}851 & 80  \\
    \extabdo\textsuperscript{$\dagger$}     & Abdomen & --       & 6{,}503  & 782     & 811     & 172 \\
    \bottomrule
  \end{tabular}
  \end{adjustbox}
\end{table}

We use the corpus, the LLM structuring pipeline and the taxonomies of~\cite{jolia}, and \cref{tab:datasets} gives the datasets.
The pipeline splits each report into findings tagged with their organ and gives the 252 finding labels used by the finding-label supervision and by $\mathcal{L}_{\text{OSL}}$ (\cref{app:osl}).
Images are resampled to 1\,mm and cropped to $288^3$ voxels, with the 11 intensity windows of Pillar-0~\cite{pillar0} as input channels.
TotalSegmentator~\cite{wasserthal2023totalsegmentator} masks are computed once on every training volume, and a fixed table maps organs to their classes (\cref{app:coverage}).

\paragraph{Ethics.}
\stanford, \ctrate and \inspect are public de-identified datasets, used under their licences or data use agreements.
\extabdo and \extchest are retrospective studies acquired in routine care at a clinical center in France. They were de-identified by the hospital before analysis and are used only for evaluation, to train and test the linear heads, never for pretraining. None of their images appears in the paper.

\section{Method and training details}
\label{app:method}

\subsection{Query grounding}
\label{app:coverage}
\begin{figure}[!htbp]
  \centering
  \begin{tikzpicture}[x=1pt,y=-1pt,line join=round,font=\footnotesize]
  \colorlet{pBandV}{orgA!20!white}\colorlet{pBandVLine}{orgA!75!white}   %
  \def\jorgA{pA}\def\jorgB{pB}\def\jorgC{pC}
  \def\jobr{0.30}\def\jobimg{figures/ct_cube_ob30}
  \jvoxgeom{6}{5.4}{1.87}{1.6}   %
  \filldraw[fill=pBandV,draw=pBandVLine,line width=0.6pt,rounded corners=3pt] (73,0.5) rectangle (233.5,57);
  \node[jnote,anchor=north,inner sep=0] at (35,0.5) {CT volume};   %
  \jcto{11.6}{9.5}{46.8}{0}   %
  \draw[jflow] (35,56.8) -- (35,66.5);
  \draw[jflow,rounded corners=3pt] (35,86) -- (69,86) -- (69,60) -- (116,60) -- (116,49.5);
  \draw[jblock,fill=pBlock,draw=pBlockLine,rounded corners=2pt] (3,67) -- (67,67) -- (58,104) -- (12,104) -- cycle;
  \node[align=center,font=\footnotesize,inner sep=0] at (35,84) {3D vision\\encoder};
  \jflame{56}{101}
  
  \foreach \y/\col/\q in {17/pA/1,27/pB/2,46.5/pC/K}
    \node[jquery,draw=\col,fill=\col!50!white] at (84.5,\y) {$q_{\q}$};
  \node[font=\scriptsize,inner sep=0,rotate=90] at (84.5,35.5) {$\cdots$};
  \draw[jflow] (90.5,31) -- (97.5,31);
  \draw[jblock,fill=pBlock,draw=pBlockLine] (98,14) rectangle (134,49);
  \node[align=center,font=\scriptsize,inner sep=0] at (116,29.5) {Cross-\\[-1pt]attention};
  \jflame{129}{47}
  \draw[jflow] (134,31) -- (141,31);

  \jvox{147.5}{23}{pA}{0/0/0/3,1/0/0/3,0/1/0/2,1/1/0/2,0/0/1/2,1/0/1/2,0/0/2/1,2/0/0/1,2/1/0/1,0/2/0/1,2/2/0/1}
  \jvox{173.5}{23}{pB}{0/2/0/3,2/2/0/2,1/2/0/1,0/2/1/2,0/1/0/1,2/1/0/1,1/0/0/1}
  \jvox{211}{23}{pC}{1/2/0/3,1/1/0/2,1/2/1/2,0/2/0/1,2/2/0/1,1/0/0/1}
  \node at (198.3,30) {$\cdots$};
  \foreach \x/\s in {156.2/1,182.2/2,219.7/K} \node[anchor=base] at (\x,14) {$A^{\s}$};

  \draw[jtrain] (73,63) rectangle (233.5,139);
  \node[jnote,anchor=north west,inner sep=1pt,font=\scriptsize\itshape] at (75.5,64.5) {training only};   %
  \jcto{80.8}{78.5}{46.8}{1}   %
  \node[jnote,anchor=north,inner sep=0] at (104.2,127.4) {organ masks};
  \draw[jflow] (128.1,106) -- (141.5,106);
  \jvox{147.5}{98}{pA}{0/0/0/3,1/0/0/3,0/1/0/3,1/1/0/3,0/0/1/3,1/0/1/3,0/0/2/3}
  \jvox{173.5}{98}{pB}{0/2/0/3,2/2/0/3,0/2/1/3}
  \jvox{211}{98}{pC}{1/2/0/3,1/1/0/3,1/2/1/3}
  \node at (198.3,105) {$\cdots$};
  \foreach \x/\s in {156.2/1,182.2/2,219.7/K} \node[anchor=base] at (\x,128) {$M^{\s}$};

  \node[draw=jLine,fill=pLoss,line width=0.9pt,rounded corners=4pt,inner sep=1.5pt,minimum height=12pt,minimum width=87.5pt] at (184.75,75) {$\mathcal{L}_{\text{QG}} = \mathrm{CE}\big(M^k, A^k\big)$};
  \foreach \x in {156.2,182.2,219.7} {
    \draw[jflow] (\x,43) -- (\x,68);
    \draw[jflow] (\x,90) -- (\x,82);
  }
\end{tikzpicture}%
  \caption{\textbf{Query grounding.}
    Each learnable query $q_k$ attends over the patch tokens of the volume, giving an attention map $A^k$.
    During training only, the organ mask is projected onto the token grid as a target $M^k$, the normalised number of mask voxels in each token, and the query grounding loss is the cross-entropy between $M^k$ and $A^k$.
    One token scale is drawn: each organ has one query per scale, and $\mathcal{L}_{\text{QG}}$ is the mean of this term over the scales and the organs with a mask in the crop.}
  \label{fig:coverage}
\end{figure}
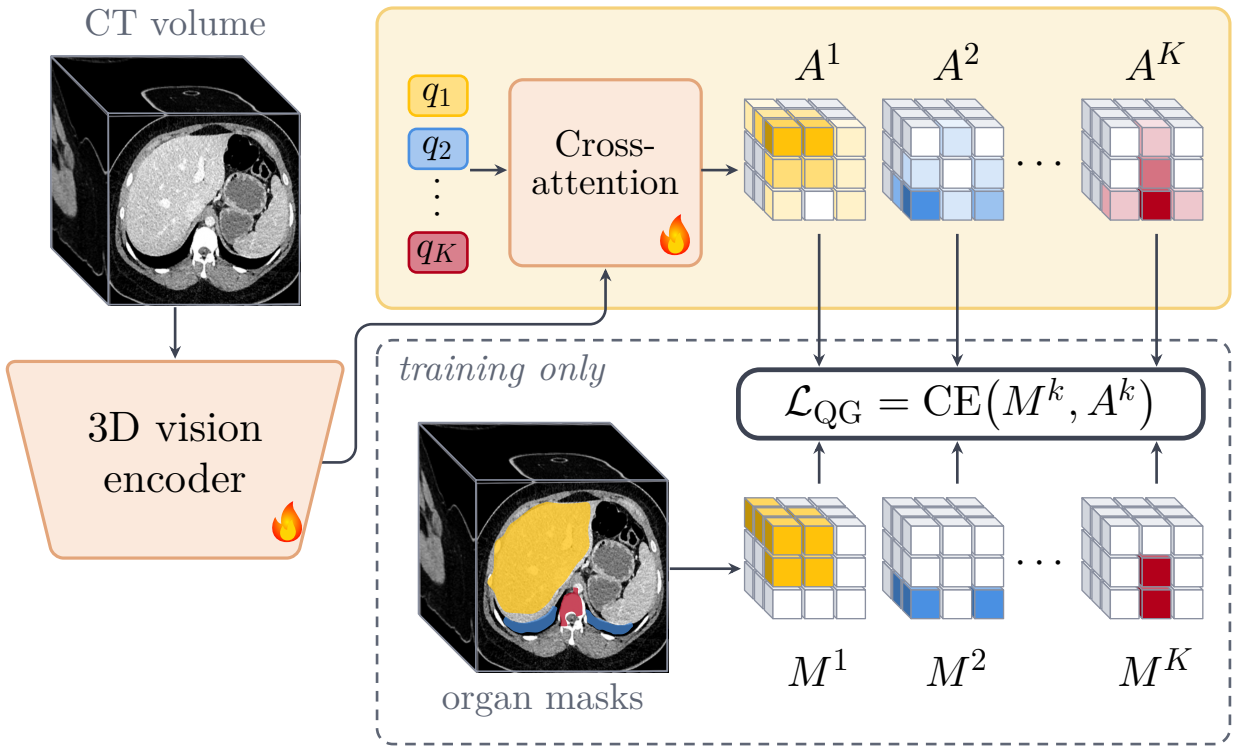

\cref{fig:coverage} details the query grounding loss of \cref{sec:coverage}.
A fixed table maps each organ of the taxonomy to its TotalSegmentator classes. {A few organs have no class of their own and are supervised with the mask of a neighbouring structure.
The ablations of \cref{sec:res_mask} use a simpler table of at most 42 organs, each with its own class.}

\subsection{Opposite sentence loss}
\label{app:osl}
COLIPRI~\cite{colipri} contrasts positive sentences of the report with their negation, built with a single template (``No \{sentence\}.''), through a two-way softmax with a temperature.
$\mathcal{L}_{\text{OSL}}$ applies the same idea to the 252 finding labels, wicyth $n$ pairs of positive and negative prompts per label, such as ``[label]'' / ``no [label]''.
We use $n=8$, the in-distribution pairs of the zero-shot evaluation (\cref{tab:zs_templates}). The 29 other pairs of that evaluation are never seen in training.

For a volume $v$ and a label $f$, $\bar{c}^{+}_{v,f}$ and $\bar{c}^{-}_{v,f}$ are the mean cosine similarities between the [CLS] token of the volume and the $n$ positive or the $n$ negative prompts, both projected as in $\mathcal{L}_{\text{CLIP}}$:
$\bar{c}^{\pm}_{v,f} = \frac{1}{n}\sum_{j=1}^{n}\cos\big(z_{\texttt{[CLS]},v}, t^{\pm}_{f,j}\big)$.
Their difference is the score of the zero-shot evaluation. With $y_{v,f}=1$ if finding $f$ is present in volume $v$ and $0$ otherwise,
\begin{equation*}
  \mathcal{L}_{\text{OSL}} = \frac{1}{\lvert\mathcal{A}\rvert}\sum_{(v,f) \in \mathcal{A}} \mathrm{BCE}\Big(\sigma\big((\bar{c}^{+}_{v,f} - \bar{c}^{-}_{v,f})/\tau + b\big),\, y_{v,f}\Big),
\end{equation*}
where $\mathcal{A}$ holds the pairs $(v,f)$ of the batch for which $f$ is annotated, $\tau$ is the temperature of $\mathcal{L}_{\text{CLIP}}$ and $b$ a learned bias.
With $n=1$ and $b=0$, this is the loss of COLIPRI, applied to labels instead of sentences.
$\mathcal{L}_{\text{OSL}}$ is added with weight 1 to the phase of $\mathcal{L}_{\text{CLIP}}$.

\subsection{Training recipe}
\label{app:training}
{%
\JoliaTwo uses \HieraCT with about 77M parameters
Training is run for 180k steps on 48 H100 gpus.
The loss weights are 1 for $\mathcal{L}_{\text{CLIP}}$, $\mathcal{L}_{\text{labels}}$, $\mathcal{L}_{\text{OSL}}$ and \DenseBT, and 0.5 for $\mathcal{L}_{\text{ConQuer}}$, $\mathcal{L}_{\text{QG}}$, \iBOT and $\mathcal{L}_{\text{seg}}$.
We use AdamW with a weight decay of 0.1, gradient clipping at 1.
The learning rate warms up over 1k steps to $3\times10^{-5}$, decays to $10^{-6}$, and ends with a cooldown over the last 5k steps.
Inputs are crops of $288^3$ voxels at 1\,mm, taken from $336^3$, with the 11 intensity windows of Pillar-0~\cite{pillar0}.
Augmentations are random rotations up to $20^\circ$ and random intensity shifts.
The frozen text encoder is Qwen3-Embedding-8B~\cite{zhang2025qwen3embedding}. The global projection has 1152 dimensions and each organ token $c_k$ 576, from single-head queries on tokens projected to 192.
\iBOT masks 60\% of the $2^3$ blocks, with 4096 prototypes, a head of 3 layers (2048, bottleneck 256), a teacher momentum of 0.996 and temperatures of 0.07 and 0.1.
The segmentation head reads $s_0$, with a hidden dimension of 128, and predicts the 118 classes on a $96^3$ grid.\par}

The ablation runs differ on a few points, given in \cref{app:ablation_setup}.

\section{Baselines}
\label{app:baselines}
\cref{tab:pretraining_data} lists the pretraining data of every model we compare with, for the checkpoint we evaluate rather than for the largest model of each paper. Pillar-0-Best takes, on each test set and segmentation dataset, the better of Pillar-0-Abd and Pillar-0-Chest.
Our corpus has 58k studies, compared with 159k for RadFinder and 92k for CAPTOR, which add a private hospital corpus to public datasets, and 150k to 570k for the image-only models (CT-FM, VoCo, FlexiCT, Curia, nnFoundation).
SPECTRE, FlexiCT-VLM, RadFinder, CAPTOR, \Jolia and \JoliaTwo all train on \ctrate and \stanford, so only the two external sets are unseen by every model.
\begin{table*}[!htbp]
  \centering
  \scriptsize
  \setlength{\tabcolsep}{4pt}
  \caption{\textbf{Pretraining data of the compared models}, for the checkpoint we evaluate: anatomical region, number of CT scans (in the unit of each paper: reconstructions, series or exams), number of these scans with a paired report, and data sources. Counts are rounded.}
  \label{tab:pretraining_data}
  \begin{tabularx}{\textwidth}{@{}l l r r >{\raggedright\arraybackslash}X@{}}   %
    \toprule
    Model & Region & Scans & Reports & Data \\
    \midrule
    \multicolumn{5}{@{}l}{\textit{Vision--language}} \\
    CT-CLIP~\cite{hamamci2024ctclip}      & chest          & 47k  & 47k  & \ctrate \\
    fVLM~\cite{fvlm}                      & chest          & 47k  & 47k  & \ctrate \\
    ARC-CT~\cite{arcct}                   & chest          & 47k  & 47k  & \ctrate \\
    COLIPRI~\cite{colipri}                & chest          & 95k  & 23k  & \ctrate, NLST~\cite{nlst2011} \\
    Pillar-0-Chest~\cite{pillar0}         & chest          & 86k  & 86k  & private (UCSF) \\
    Pillar-0-Abd~\cite{pillar0}           & abdomen        & 43k  & 43k  & private (UCSF) \\
    RADAR+~\cite{radar}                   & abdomen        & 15k  & 15k  & \stanford \\
    Merlin~\cite{blankemeier2024merlin}   & abdomen        & 15k  & 15k  & \stanford \\
    NV-Reason-CT~\cite{nvreason}          & chest, abdomen & 70k  & 63k  & \ctrate, CancerVerse, private (NIH) \\
    \Jolia~\cite{jolia}                   & chest, abdomen & 58k  & 58k  & \ctrate, \stanford, \inspect \\
    CAPTOR~\cite{captor}                  & chest, abdomen & 92k  & 92k  & \ctrate, \stanford, \inspect, BIMCV-R~\cite{chen2024bimcvr}, private \\
    RadFinder~\cite{radfinder}            & chest, abdomen & 159k & 159k & \ctrate, \stanford, \inspect, private \\
    SPECTRE~\cite{spectre}                & chest, abdomen & 230k & 86k  & \ctrate, \stanford, \inspect, 5 public sets \\
    FlexiCT-VLM~\cite{flexictvlm}         & whole body     & 266k & 64k  & \ctrate, \stanford, \inspect, 53 public sets \\
    M3D-CLIP~\cite{m3d}                   & whole body     & 120k & 42k  & Radiopaedia (captions) \\
    RadFM~\cite{radfm}                    & whole body     & 500k & 500k & Radiopaedia, MedPix, PMC (captions; CT and MRI) \\
    EVLF-FM~\cite{evlffm}                 & 2D slices      & 31k  & --   & CT slices of MedMNIST; 18 sets and 10 modalities in all (1.3M 2D images) \\
    \midrule
    \multicolumn{5}{@{}l}{\textit{Image only}} \\
    Coralbay~\cite{coralbay}              & chest, abdomen, head and neck & 11k & -- & 7 public sets \\
    MedVTok~\cite{medvtok}                & whole body     & 35k  & --   & 28 public sets, plus 8.7M 2D slices and 8 other modalities \\
    TAP-CT~\cite{tapct}                   & not stated     & 104k & --   & private (NKI) \\
    CT-FM~\cite{ctfm}                     & chest, abdomen & 148k & --   & Imaging Data Commons (69 collections, NLST~\cite{nlst2011} 86\%) \\
    Curia L~\cite{curia}                  & whole body     & 150k & --   & private (CT and MRI) \\
    Curia 2 L~\cite{curia2}               & whole body     & 150k & --   & same as Curia \\
    \CuriaMAE~\cite{curiamae}             & whole body     & 300k & --   & corpus of Curia, counted in volumes \\
    VoCo~\cite{voco,voco2}                & whole body     & 160k & --   & PreCT-160K (30 public sets) \\
    FlexiCT~\cite{flexictvlm}             & whole body     & 266k & --   & 56 public sets \\
    nnFoundation ViT, CNN~\cite{nnfoundation} & whole body & 570k & --   & 125 sources, 61\% from clinical partners; CT share of 2.1M CT, MRI and PET volumes \\
    \midrule
    \multicolumn{5}{@{}l}{\textit{Supervised segmentation}} \\
    TotalSegmentator~\cite{wasserthal2023totalsegmentator} & whole body & 1.6k & -- & Basel (partly public) \\
    SuPreM~\cite{suprem}                  & abdomen        & 2.1k & --   & AbdomenAtlas (public) \\
    VISTA3D~\cite{vista3d}                & whole body     & 11k  & --   & 14 public sets, private bone lesions \\
    \midrule
    \textbf{\JoliaTwo}                    & chest, abdomen & 58k  & 58k  & \ctrate, \stanford, \inspect \\
    \bottomrule
  \end{tabularx}
\end{table*}

\cref{tab:methods} gives the method of each model and what it saw of segmentation during pretraining.
Among the other vision--language models, only RADAR+ trains a decoder on masks. fVLM, ARC-CT and CAPTOR use them to select tokens or crops, and fVLM and CAPTOR also need them at test time.
Image-only models have no segmentation in their loss, but several train on our segmentation datasets, as the supervised segmentation models do; \JoliaTwo trains on none of them, although its masks come from the TotalSegmentator tool.
\begin{table*}[!htbp]
  \centering
  \scriptsize
  \setlength{\tabcolsep}{2pt}
  \caption{\textbf{Pretraining method of the compared models}, for the checkpoint we evaluate. Glob.: global image-report alignment. Region: image-text alignment below the volume. Image SSL: image-only objective and the tokens it acts on. Seg.\ in training: segmentation in pretraining; TS: pseudo-masks of the TotalSegmentator tool, distinct from its dataset. Test masks: masks needed to read findings at test time. Seen datasets: our segmentation datasets in the pretraining data, at the dataset level, as most papers do not publish their splits; $\dagger$: through an aggregate dataset. ``None listed'' does not prove that private data are disjoint from public ones.}
  \label{tab:methods}
  \begin{adjustbox}{max totalheight=0.86\textheight}   %
  \begin{tabularx}{\textwidth}{@{}>{\raggedright\arraybackslash}p{70pt}>{\raggedright\arraybackslash}X>{\raggedright\arraybackslash}X>{\raggedright\arraybackslash}p{13pt}>{\raggedright\arraybackslash}p{34pt}>{\raggedright\arraybackslash}p{48pt}>{\raggedright\arraybackslash}X>{\raggedright\arraybackslash}p{15pt}>{\raggedright\arraybackslash}X@{}}
    \toprule
    Model & Method & Data & Glob. & Region & Image SSL & Seg.\ in training & Test masks & Seen datasets \\
    \midrule
    \multicolumn{9}{@{}l}{\textit{Vision--language}} \\
    CT-CLIP~\cite{hamamci2024ctclip} & CLIP & \ctrate & yes & -- & -- & -- & no & none listed \\
    fVLM~\cite{fvlm} & organ-wise CLIP, false-negative correction & \ctrate & -- & organ & -- & TS: hard token scope & yes & none listed \\
    ARC-CT~\cite{arcct} & CLIP, organ-wise CLIP, finding labels & \ctrate & yes & organ & -- & TS: hard attention scope, mask pooling & no & none listed \\
    COLIPRI~\cite{colipri} & CLIP, report generation, MAE & \ctrate, NLST~\cite{nlst2011} & yes & -- & MAE (dense) & -- & no & none listed \\
    Pillar-0~\cite{pillar0} & CLIP & private (UCSF) & yes & -- & -- & -- & no & none listed \\
    RADAR+~\cite{radar} & CLIP, organ-wise CLIP, segmentation head & \stanford & yes & organ & -- & TS: Dice on a decoder; predicted mask as token scope & no & none listed; U-Net init.\ not documented \\
    Merlin~\cite{blankemeier2024merlin} & CLIP, EHR phenotypes & \stanford & yes & text only & -- & -- & no & none listed \\
    CAPTOR~\cite{captor} & CLIP on organ-centred crops & \ctrate, \stanford, \inspect, BIMCV-R~\cite{chen2024bimcvr}, private & -- & crop & -- & TS: crop centring only & yes & none listed \\
    NV-Reason-CT~\cite{nvreason} & report generation, reinforcement learning, COLIPRI init. & \ctrate, CancerVerse, private (NIH) & -- & -- & inherited & -- & no & none named \\
    \Jolia~\cite{jolia} & CLIP, organ-wise CLIP & \ctrate, \stanford, \inspect & yes & organ & -- & -- & no & none listed \\
    RadFinder~\cite{radfinder} & SigLIP, disease prompts, depth localisation, frozen SPECTRE local backbone & \ctrate, \stanford, \inspect, private & yes & disease, depth & inherited & -- & no & AMOS, MSD$\dagger$, KiTS$\dagger$ (backbone) \\
    SPECTRE~\cite{spectre} & DINO-style SSL, then SigLIP & \ctrate, \stanford, \inspect, 5 public sets & yes & -- & DINO (global), iBOT (dense) & -- (TS for quality control) & no & AMOS, MSD$\dagger$, KiTS$\dagger$ \\
    FlexiCT-VLM~\cite{flexictvlm} & DINOv3-style SSL, then CLIP and iBOT & \ctrate, \stanford, \inspect, 53 public sets & yes & -- & iBOT (dense) & -- & no & AMOS, TotalSeg., CRLM, LIDC, MSD$\dagger$, KiTS$\dagger$ \\
    M3D-CLIP~\cite{m3d} & CLIP & Radiopaedia & yes & -- & -- & -- & no & none listed \\
    RadFM~\cite{radfm} & report generation & Radiopaedia, MedPix, PMC & -- & -- & -- & -- & no & none listed \\
    EVLF-FM~\cite{evlffm} & 2D language model, supervised and reinforcement & MedMNIST and 17 other sets, 10 modalities & -- & boxes & -- & expert boxes as a reward & no & not stated \\
    \midrule
    \multicolumn{9}{@{}l}{\textit{Image only}} \\
    Coralbay~\cite{coralbay} & DINO & 7 public sets & -- & -- & DINO (global) & -- & no & LIDC, AMOS$\dagger$, MSD$\dagger$, KiTS$\dagger$ \\
    MedVTok~\cite{medvtok} & autoencoder, alignment to expert encoders & 28 public CT sets, 3D volumes and 2D slices & -- & -- & reconstruction (dense) & -- & no & TotalSeg., LIDC, KiTS, MSD \\
    TAP-CT~\cite{tapct} & 3D DINOv2 & private (NKI) & -- & -- & DINO (global), iBOT (dense) & -- & no & none listed \\
    CT-FM~\cite{ctfm} & intra-sample SimCLR & Imaging Data Commons & -- & -- & SimCLR (pooled crops) & -- & no & LIDC, KiTS \\
    Curia L~\cite{curia} & 2D DINOv2 & private & -- & -- & DINO (global), iBOT (dense) & -- & no & none listed \\
    Curia 2 L~\cite{curia2} & 2D DINOv2, SigReg & private & -- & -- & DINO (global), iBOT (dense) & -- & no & none listed \\
    \CuriaMAE~\cite{curiamae} & MAE, Barlow Twins & private & -- & -- & MAE (dense), pooled crops & -- & no & none listed \\
    VoCo~\cite{voco,voco2} & volume contrast & PreCT-160K & -- & -- & position contrast (pooled crops) & -- & no & AMOS, TotalSeg., MSD, KiTS, LIDC \\
    FlexiCT~\cite{flexictvlm} & DINOv3-style SSL & 56 public sets & -- & -- & DINO (global), iBOT (dense) & -- & no & AMOS, TotalSeg., CRLM, LIDC, MSD$\dagger$, KiTS$\dagger$ \\
    nnFoundation~\cite{nnfoundation} & MAE & 125 sources, 61\% from clinical partners & -- & -- & MAE (dense) & -- & no & KiTS, CRLM \\
    \midrule
    \multicolumn{9}{@{}l}{\textit{Supervised segmentation}} \\
    TotalSegmentator~\cite{wasserthal2023totalsegmentator} & supervised nnU-Net & Basel & -- & -- & -- & manual, model-assisted & no & TotalSeg.; rest not stated \\
    SuPreM~\cite{suprem} & supervised segmentation & AbdomenAtlas & -- & -- & -- & manual, pseudo tumours; BCE and Dice & no & AMOS, MSD, KiTS (pool; list not published) \\
    VISTA3D~\cite{vista3d} & promptable supervised segmentation & 14 public sets, private & -- & -- & -- & manual, TS, SAM supervoxels; Dice and CE & no & AMOS, TotalSeg., MSD, CRLM, LIDC \\
    \midrule
    \textbf{\JoliaTwo} & CLIP, organ-wise CLIP, finding labels, segmentation head, \iBOT, \DenseBT & \ctrate, \stanford, \inspect & yes & organ & \iBOT, \DenseBT (dense) & TS: loss on query attention; Dice and CE on a discarded head & no & none \\
    \bottomrule
  \end{tabularx}
  \end{adjustbox}
\end{table*}

\section{Finding classification}
\label{app:findings}
\cref{tab:competitors} gives the AUC of every model on each test set (\cref{sec:res_findings}).
\JoliaTwo is first on the four test sets, 3.0 points above FlexiCT-VLM on the mean (88.9 compared with 85.9). With the [CLS] token alone, it reaches 86.7, still above every other model.
For every model, we test several readouts of the linear probe, including the class token, the average of the patch tokens and their concatenation, and report the best (\cref{sec:res_findings}).
The readouts of \JoliaTwo are compared in \cref{app:readout}.

\subsection{Cross-center transfer}
\label{app:transfer}
\cref{tab:cross_transfert} follows the cross-center protocol of~\cite{jolia}. A linear head is trained on the frozen features of \stanford or \ctrate, labelled with the taxonomy of~\cite{jolia} used by the external sets (\cref{tab:datasets}), and tested on the same center (ID) and on \extabdo or \extchest (OOD), over 5 seeds.
\JoliaTwo leads on both external sets with a head trained on \stanford or on \ctrate, and \cref{app:readout} gives the transfer of each readout.
Compared with a head trained on the target center, \JoliaTwo loses 0.1 point on \extabdo and 8.7 on \extchest. On \extchest, every model loses 2.8 to 22.4 points.

\section{Segmentation}
\label{app:dense}
\cref{tab:seg_targets} lists the 13 datasets.
\cref{tab:seg_training} gives the decoder and the optimiser of each group of encoders.
All runs see 500k samples, or 62.5k steps of 8 samples, with a Dice and cross-entropy loss. SGD uses a momentum of 0.99 with Nesterov and a weight decay of $3\times10^{-5}$, and AdamW a weight decay of 0.05.
With a frozen encoder, the learning rate warms up over the first 5\% of training, then decays polynomially.
Fine-tuning warms up the decoder alone over the first 5\%, then the whole network over the next 5\%, then decays polynomially, with the same learning rate for the encoder and the decoder.
Inference uses a sliding window with an overlap of 0.5 and Gaussian weighting.
\begin{table}[!htbp]
  \centering
  \footnotesize
  \caption{\textbf{The 13 segmentation datasets.} Lesion class: the class reported in the ablation tables; organ datasets report the mean over their classes.}
  \label{tab:seg_targets}
  \begin{adjustbox}{max width=\linewidth}   %
  \begin{tabular}{@{}lllcc@{}}
    \toprule
    Dataset & Classes & Reported in the ablations & Train & Eval. \\
    \midrule
    MSD Liver~\cite{antonelli2022msd} & liver, lesion & lesion & 91 & 40 \\
    AMOS~\cite{ji2022amos} & 15 organs & mean of the organs & 200 & 100 \\
    TotalSegmentator v2~\cite{wasserthal2023totalsegmentator} & 115 of its 117 structures & mean of the structures & 1082 & 57 \\
    MSD Lung~\cite{antonelli2022msd} & lesion & lesion & 44 & 19 \\
    MSD Colon~\cite{antonelli2022msd} & lesion & lesion & 88 & 38 \\
    MSD Hepatic Vessel~\cite{antonelli2022msd} & vessels, tumour & tumour & 272 & 31 \\
    KiTS23~\cite{heller2023kits21} & left and right kidney, cyst, lesion & lesion & 348 & 42 \\
    CRLM~\cite{simpson2024crlm} & liver, tumour, hepatic vein, portal vein & tumour & 118 & 41 \\
    WAW-TACE~\cite{bartnik2024wawtace} & liver, HCC & HCC & 162 & 23 \\
    TotalSegmentator Liver Lesions & lesion & lesion & 674 & 82 \\
    LIDC~\cite{armato2011lidc} & high-consensus nodules & nodules & 801 & 92 \\
    ImageCAS~\cite{zeng2023imagecas} & coronary arteries & coronary arteries & 683 & 61 \\
    SegTHOR~\cite{lambert2020segthor} & heart, aorta, trachea, oesophagus & mean of the organs & 21 & 15 \\
    \bottomrule
  \end{tabular}
  \end{adjustbox}
\end{table}

\begin{table}[!htbp]
  \centering
  \footnotesize
  \caption{\textbf{Decoder and optimiser of the segmentation evaluation}, for each group of encoders. LR: learning rate. AMOS, TS, lung: these datasets only.}
  \label{tab:seg_training}
  \begin{adjustbox}{max width=\linewidth}   %
  \begin{tabular}{@{}p{0.31\linewidth}p{0.31\linewidth}p{0.32\linewidth}@{}}
    \toprule
    Encoders & Decoder & Optimiser, LR \\
    \midrule
    \JoliaTwo & pyramid (320, 320, 256), full-resolution 32 & SGD; 0.02 on AMOS, TS, lung; $10^{-3}$ otherwise \\
    Pillar-0-Best & pyramid (320, 256, 128), full-resolution 64 & AdamW, $3\times10^{-4}$ \\
    ARC-CT, CT-FM, VISTA3D & pyramid (320, 256, 128, 64), full-resolution 32 & SGD; 0.02 on AMOS, TS, lung; $10^{-3}$ otherwise \\
    CAPTOR, Coralbay, Merlin, RADAR+, SuPreM, TotalSegmentator, VoCo & pyramid (320, 320, 256, 128, 64), full-resolution 32 & SGD; 0.02 on AMOS, TS, lung; $10^{-3}$ otherwise \\
    COLIPRI, CT-CLIP, Curia L, Curia 2 L, EVLF-FM, FlexiCT, FlexiCT-VLM, fVLM, M3D-CLIP, MedVTok, nnFoundation ViT, NV-Reason-CT, RadFinder, RadFM, SPECTRE, TAP-CT & single-scale patch decoder & SGD, 0.02 on AMOS, TS; AdamW, $3\times10^{-4}$ otherwise \\
    nnFoundation CNN & its own U-Net decoder & SGD, $10^{-3}$ \\
    \bottomrule
  \end{tabular}
  \end{adjustbox}
\end{table}

\begin{figure*}[!htbp]
  \centering
  \includegraphics[width=\textwidth]{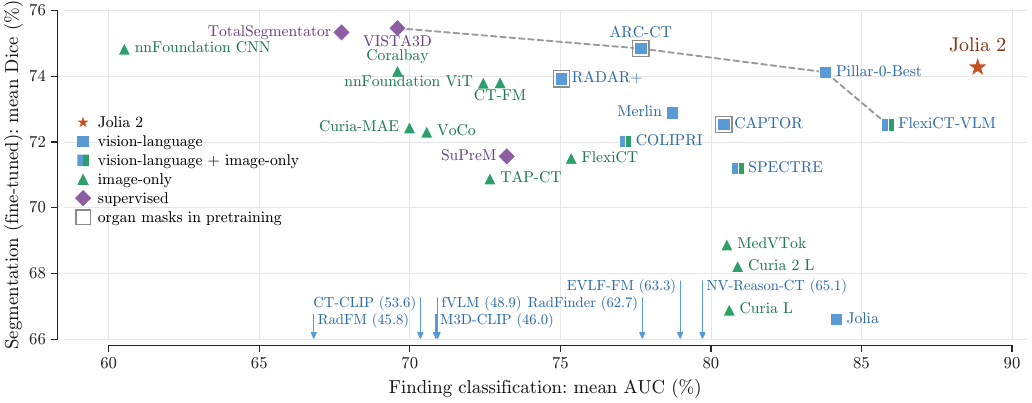}
  \caption{\textbf{Comparison with external CT foundation models, fine-tuned encoder.} As \cref{fig:competitors}, with the encoder fine-tuned on each dataset. Arrows: models below the Dice axis, with their Dice.}
  \label{fig:competitors_ft}
\end{figure*}

\cref{tab:dense,tab:dense_ft} give the Dice with all the classes of each dataset, with a frozen and a fine-tuned encoder (\cref{sec:res_dense}).
With a frozen encoder, \JoliaTwo is second on the mean Dice, and first on TotalSegmentator and on the lung tumours.
Without TotalSegmentator and AMOS, whose organs are TotalSegmentator classes, it stays second on the mean of the 11 other datasets (70.5), behind VISTA3D (72.8) and ahead of TotalSegmentator (70.4), Coralbay (69.9) and Pillar-0-Best (69.6).
Fine-tuning narrows the gaps between models, and the six best lie within 1.2 points. With a fine-tuned encoder too, no external model is better than \JoliaTwo on both tasks (\cref{fig:competitors_ft}).
\begin{table*}[!htbp]
  \centering
  \scriptsize
  \caption{\textbf{Segmentation with a fine-tuned encoder, comparison with external CT foundation models.} Dice (\%) on the 13 datasets of \cref{tab:dense}, same abbreviations.
    Scratch: ResEnc-L trained from random initialisation. Scratch, VISTA3D and TotalSegmentator are supervised segmentation models. Bold: best per column.}
  \label{tab:dense_ft}
  \setlength{\tabcolsep}{2.6pt}
  \begin{adjustbox}{max width=\linewidth, max totalheight=0.46\textheight}   %
  \begin{tabular}{@{}lcccccccccccccc@{}}
    \toprule
    Model & MSD & AMOS & TS & Lung & Colon & Liv.\ v. & KiTS & CRLM & WAW & TS liv. & LIDC & ImCAS & SegTHOR & Mean \\
    \midrule
    \JoliaTwo & 80.64 & 87.78 & \textbf{87.65} & 74.69 & 49.51 & 61.30 & \textbf{84.89} & 70.30 & 86.85 & 77.61 & 58.92 & 58.29 & 87.06 & 74.27 \\
    \Jolia & 74.52 & 85.39 & 80.12 & 66.08 & 29.84 & 56.18 & 78.75 & 66.38 & 80.43 & 71.84 & 46.58 & 44.48 & 85.23 & 66.60 \\
    \cmidrule{1-15}   %
    VISTA3D & 81.69 & 88.56 & 82.80 & 67.32 & \textbf{55.85} & 63.72 & 83.92 & \textbf{72.60} & 87.62 & 79.31 & 63.84 & 65.60 & 88.20 & \textbf{75.46} \\
    TotalSegmentator & 81.68 & 88.26 & 81.49 & \textbf{75.72} & 49.36 & 65.75 & 83.03 & 71.82 & 87.81 & 79.59 & 61.76 & 64.84 & 88.24 & 75.33 \\
    Scratch & 81.43 & 88.75 & 83.23 & 73.38 & 42.13 & \textbf{65.87} & 83.55 & 72.56 & 86.35 & 79.04 & 65.59 & 63.98 & 88.26 & 74.93 \\
    ARC-CT & 80.32 & 88.61 & 82.81 & 73.66 & 47.33 & 63.04 & 84.30 & 71.32 & 86.38 & 77.90 & 63.93 & 66.41 & 86.85 & 74.84 \\
    nnFoundation CNN & \textbf{83.81} & 88.63 & 82.34 & 72.22 & 37.43 & 65.15 & 82.66 & 72.24 & \textbf{88.21} & \textbf{81.12} & \textbf{66.99} & 63.73 & 88.25 & 74.83 \\
    Coralbay & 80.05 & 88.48 & 82.36 & 71.36 & 46.37 & 63.83 & 82.15 & 72.03 & 87.32 & 78.83 & 60.86 & 63.03 & 87.36 & 74.16 \\
    Pillar-0-Best & 80.79 & 88.21 & 84.89 & 71.86 & 44.32 & 64.36 & 81.96 & 71.03 & 87.41 & 76.44 & 61.71 & 62.83 & 87.72 & 74.12 \\
    RADAR+ & 80.65 & 87.94 & 80.10 & 66.45 & 48.58 & 63.97 & 84.47 & 71.91 & 87.75 & 79.50 & 58.19 & 63.62 & 87.82 & 73.92 \\
    CT-FM & 77.92 & 88.52 & 81.45 & 72.18 & 42.93 & 61.69 & 82.53 & 71.09 & 84.64 & 81.08 & 62.86 & 66.07 & 86.58 & 73.81 \\
    nnFoundation ViT & 76.49 & \textbf{89.31} & 87.50 & 64.49 & 51.79 & 63.90 & 80.94 & 71.45 & 83.47 & 78.59 & 59.60 & 63.38 & 88.55 & 73.80 \\
    Merlin & 81.34 & 88.07 & 83.00 & 62.93 & 37.19 & 61.62 & 81.56 & 71.13 & 87.36 & 77.24 & 63.66 & 64.59 & 87.83 & 72.89 \\
    CAPTOR & 80.27 & 88.12 & 82.51 & 65.30 & 41.64 & 61.89 & 81.58 & 71.34 & 86.12 & 78.39 & 54.86 & 63.15 & 87.70 & 72.53 \\
    FlexiCT-VLM & 79.03 & 87.72 & 82.69 & 70.78 & 52.96 & 61.76 & 81.42 & 71.30 & 82.77 & 76.62 & 52.89 & 54.16 & 88.62 & 72.52 \\
    \CuriaMAE & 78.43 & 88.56 & 82.09 & 68.96 & 36.62 & 62.69 & 82.76 & 72.32 & 86.19 & 76.51 & 56.94 & 63.22 & 86.40 & 72.44 \\
    \VoCo & 79.55 & 88.08 & 81.45 & 62.83 & 46.98 & 61.49 & 82.56 & 70.74 & 87.00 & 78.40 & 49.99 & 63.45 & 87.48 & 72.31 \\
    COLIPRI & 79.39 & 88.94 & 84.97 & 65.45 & 40.00 & 60.66 & 81.68 & 72.43 & 86.54 & 76.28 & 55.81 & 54.91 & \textbf{89.17} & 72.02 \\
    SuPreM & 76.81 & 87.86 & 79.85 & 59.46 & 46.19 & 60.38 & 81.94 & 70.23 & 84.25 & 78.15 & 54.71 & 64.45 & 86.05 & 71.56 \\
    FlexiCT & 79.42 & 88.03 & 83.86 & 67.69 & 44.81 & 61.65 & 80.98 & 71.45 & 82.89 & 69.97 & 57.75 & 52.49 & 88.61 & 71.51 \\
    SPECTRE & 77.24 & 87.63 & 79.80 & 67.37 & 48.26 & 57.59 & 78.17 & 69.72 & 82.24 & 70.87 & 52.15 & 67.01 & 87.45 & 71.19 \\
    TAP-CT & 78.34 & 87.68 & 79.79 & 66.08 & 32.69 & 58.08 & 78.29 & 70.25 & 83.87 & 73.54 & 57.10 & \textbf{67.68} & 88.19 & 70.89 \\
    MedVTok & 79.47 & 86.37 & 81.94 & 46.09 & 35.24 & 58.83 & 79.55 & 71.17 & 84.74 & 73.18 & 44.38 & 66.76 & 87.73 & 68.88 \\
    Curia 2 L & 78.58 & 85.43 & 78.46 & 64.79 & 41.17 & 59.47 & 78.40 & 69.42 & 75.31 & 67.51 & 46.74 & 54.56 & 87.05 & 68.22 \\
    Curia L & 73.67 & 84.78 & 77.73 & 60.52 & 38.34 & 58.10 & 76.01 & 68.08 & 75.05 & 67.77 & 47.27 & 55.86 & 86.48 & 66.90 \\
    NV-Reason-CT & 70.95 & 82.79 & 78.45 & 66.44 & 42.01 & 54.19 & 72.76 & 62.59 & 84.39 & 66.45 & 39.48 & 40.69 & 85.61 & 65.14 \\
    EVLF-FM & 70.66 & 82.40 & 73.24 & 56.16 & 24.15 & 51.43 & 70.46 & 68.97 & 73.12 & 62.41 & 38.60 & 64.00 & 87.66 & 63.33 \\
    RadFinder & 69.24 & 84.50 & 74.96 & 54.15 & 35.74 & 49.34 & 72.72 & 65.76 & 76.36 & 62.00 & 29.04 & 54.41 & 86.38 & 62.66 \\
    CT-CLIP & 67.02 & 74.06 & 57.95 & 49.27 & 21.97 & 40.33 & 66.48 & 54.74 & 77.11 & 60.78 & 19.38 & 27.98 & 79.57 & 53.59 \\
    fVLM & 61.16 & 77.92 & 70.66 & 40.42 & 6.28 & 32.86 & 61.48 & 58.38 & 69.26 & 30.84 & 8.55 & 34.80 & 82.46 & 48.85 \\
    M3D-CLIP & 57.43 & 77.36 & 73.34 & 11.16 & 4.86 & 25.98 & 63.23 & 59.42 & 62.45 & 27.05 & 6.44 & 45.97 & 82.61 & 45.95 \\
    RadFM & 66.83 & 79.81 & 67.92 & 5.52 & 1.49 & 52.29 & 66.80 & 63.70 & 73.47 & 24.73 & 0.03 & 7.90 & 84.29 & 45.75 \\
    \bottomrule
  \end{tabular}
  \end{adjustbox}
\end{table*}

\section{Zero-shot classification and retrieval}
\label{app:global}

\subsection{Zero-shot classification}
\label{app:zs_short}
\begin{table*}[!htbp]
\centering
\footnotesize
\caption{\textbf{Prompt pairs of the short zero-shot protocol.} The 37 pairs of \cref{fig:zs_pairs_abd,fig:zs_pairs_chest}, in the same order. The score of a finding is its mean similarity with the positive prompts minus that with the negative ones. The eight pairs of the first family are in-distribution (ID): \JoliaTwo is trained on them through $\mathcal{L}_{\text{OSL}}$. The 29 others are out-of-distribution (OOD). A French pair was also run and is not reported.}
\label{tab:zs_templates}
\renewcommand{\arraystretch}{1.05}
\begin{adjustbox}{max width=\linewidth}   %
\begin{tabular}{@{}p{2.9cm} c p{5.4cm} p{7.2cm}@{}}
\toprule
\textbf{Family} & \textbf{Pairs} & \textbf{Positive prompt} & \textbf{Negative prompt} \\
\midrule
In-distribution (ID) & 8 & [label] & no [label] \\
 &  & there is evidence of [label] & there is no evidence of [label] \\
 &  & [label] present & [label] not present \\
 &  & findings consistent with [label] & no findings consistent with [label] \\
 &  & The CT scan shows [label] & The CT scan does not show [label] \\
 &  & a CT showing [label] & a CT without [label] \\
 &  & Impression: [label] & Impression: no [label] \\
 &  & this is an image of a [label] & this is an image with no [label] \\
\midrule
CT-CLIP templates & 5 & Findings are compatible with [label]. & Findings are not compatible with [label]. \\
 &  & There is [label]. & There is no [label]. \\
 &  & [label] is seen. & [label] is not seen. \\
 &  & [label]. & Not [label]. \\
 &  & [label] is observed. & [label] is not observed. \\
\midrule
Report style & 6 & [label] is seen. & No [label] is seen. \\
 &  & [label] is identified. & No [label] is identified. \\
 &  & The study demonstrates [label]. & The study demonstrates no [label]. \\
 &  & Radiology report: [label] was found. & Radiology report: [label] was not found. \\
 &  & FINDINGS: [label]. & FINDINGS: No [label]. \\
 &  & Compared to the prior exam, there is [label]. & Compared to the prior exam, there is no [label]. \\
\midrule
Antonyms & 6 & presence of [label] & absence of [label] \\
 &  & positive for [label] & negative for [label] \\
 &  & [label] confirmed & [label] ruled out \\
 &  & the patient suffers from [label] & the patient is free of [label] \\
 &  & [label] established & [label] excluded \\
 &  & [label] (+) & [label] (-) \\
\midrule
Form-like & 5 & [label]: yes & [label]: no \\
 &  & The patient has [label]. & The patient does not have [label]. \\
 &  & Is there [label]? Yes. & Is there [label]? No. \\
 &  & [label]: true & [label]: false \\
 &  & Diagnosis: [label] & Diagnosis: [label] excluded \\
\midrule
Uncertain or qualified & 5 & findings suggestive of [label] & findings not suggestive of [label] \\
 &  & possible [label] & no [label] \\
 &  & [label] cannot be excluded & [label] is excluded \\
 &  & mild [label] & no [label] \\
 &  & [label], unchanged & no [label] \\
\midrule
Generic negative & 2 & [label] & normal study \\
 &  & [label] & no acute abnormality \\
\bottomrule
\end{tabular}
\end{adjustbox}
\end{table*}

\begin{figure*}[!htbp]
  \centering
  \includegraphics[width=\textwidth]{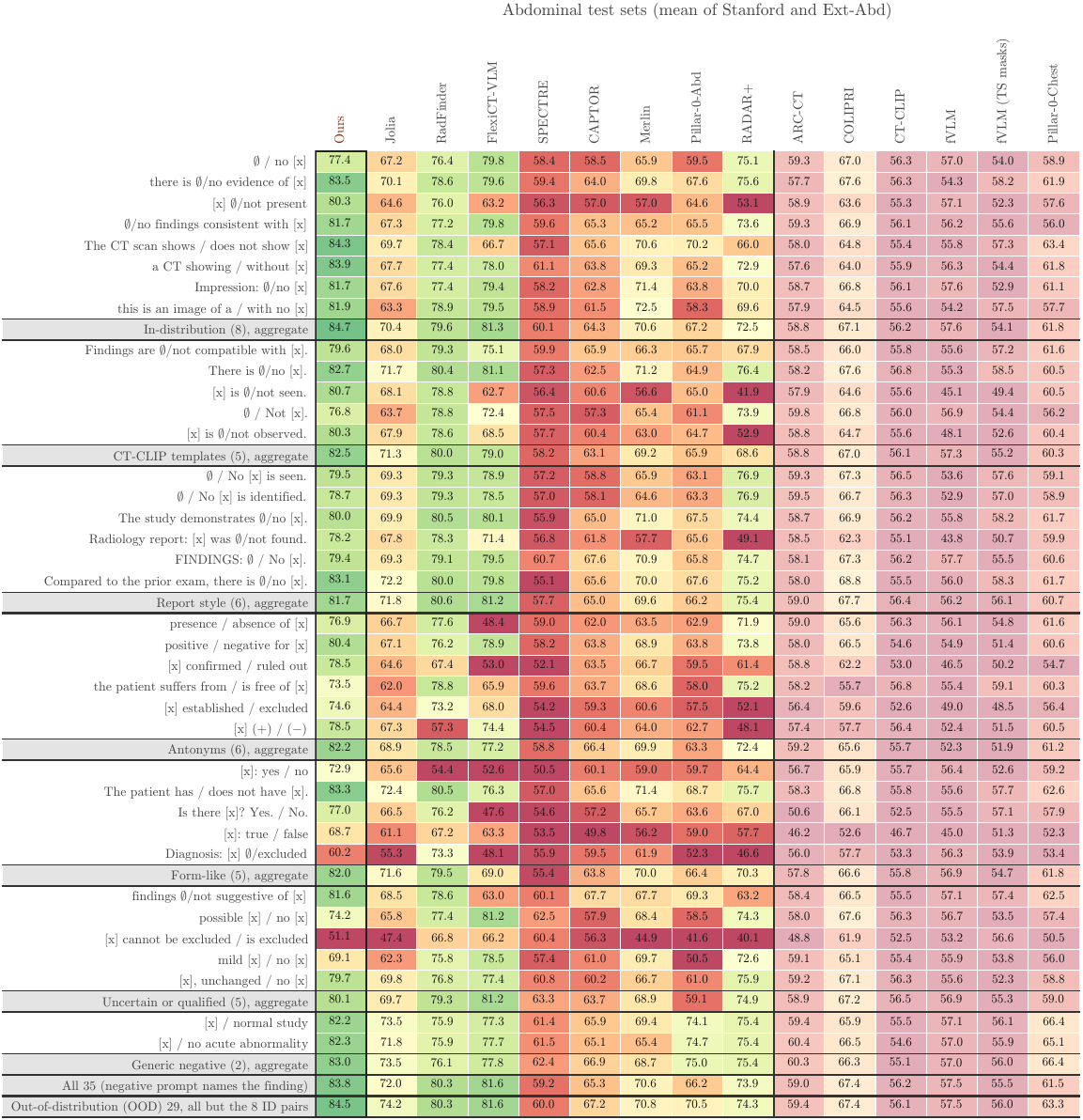}
  \caption{\textbf{Zero-shot classification with short prompts, abdominal test sets.} AUC (\%), mean of \stanford and \extabdo. Rows without a band are single prompt pairs (\cref{app:zs_short}), grouped by family and giving the positive and the negative prompt, with $\emptyset$ standing for the positive prompt when it is the negative one without its negation. Banded and framed rows are aggregates: the aggregate over the pairs of the family above them, then over the 35 pairs of the six families whose negative prompt names the finding, and over the 29 out-of-distribution (OOD) pairs that \JoliaTwo never sees in training (S-OOD in \cref{tab:zs_summary}); the eight pairs of the first family are in-distribution (ID). ``Ours'' is boxed. Right of the vertical rule, in paler colours: models whose released checkpoint was trained on the other anatomical region only, which are not ranked. fVLM (TS masks): fVLM run as in its paper, sliding windows and organ-mask-gated attention with TotalSegmentator masks.}
  \label{fig:zs_pairs_abd}
\end{figure*}

\begin{figure*}[!htbp]
  \centering
  \includegraphics[width=\textwidth]{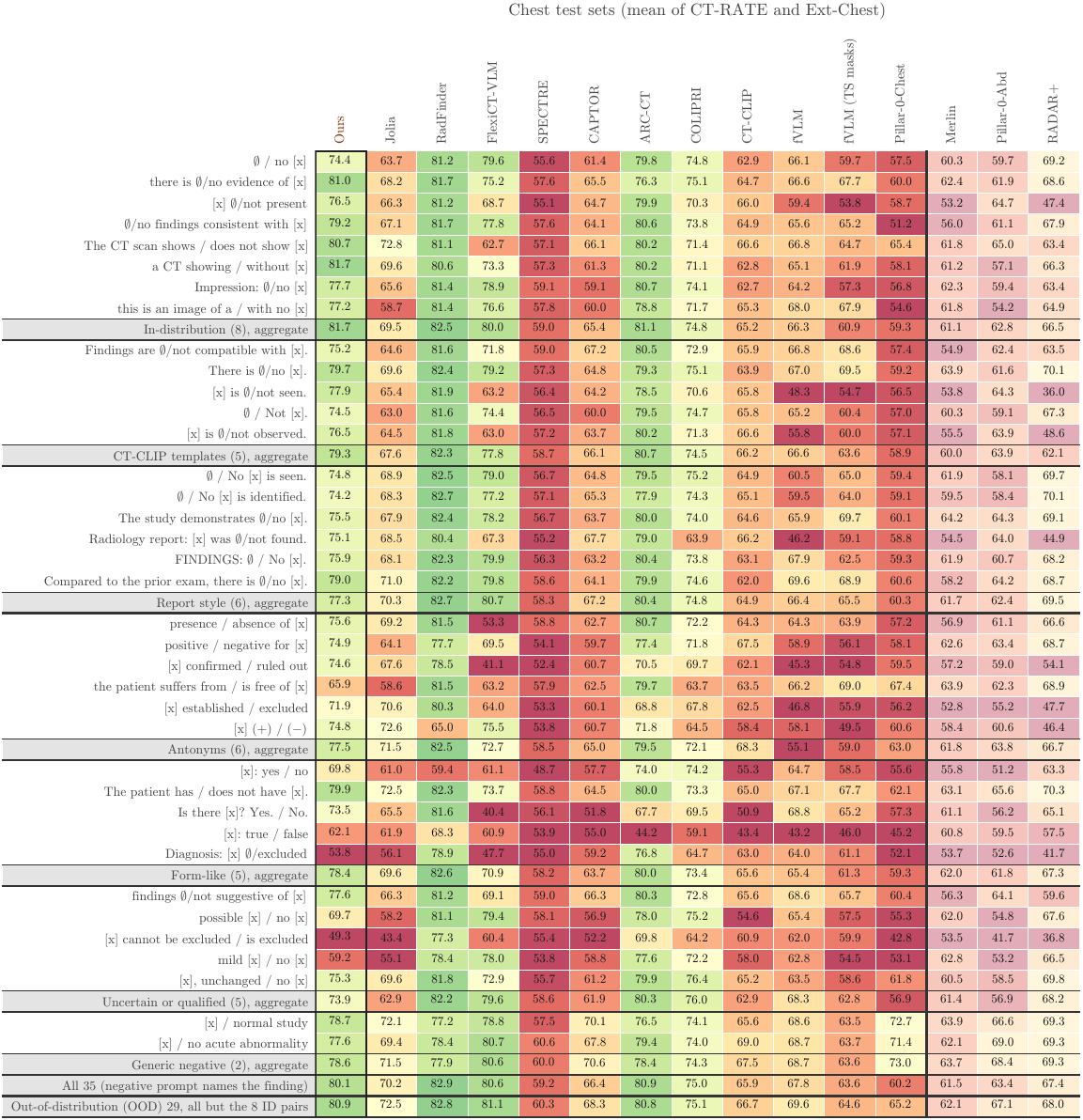}
  \caption{\textbf{Zero-shot classification with short prompts, chest test sets.} As \cref{fig:zs_pairs_abd}, mean of \ctrate and \extchest.}
  \label{fig:zs_pairs_chest}
\end{figure*}

\begin{figure*}[!htbp]
  \centering
  \includegraphics[width=\textwidth]{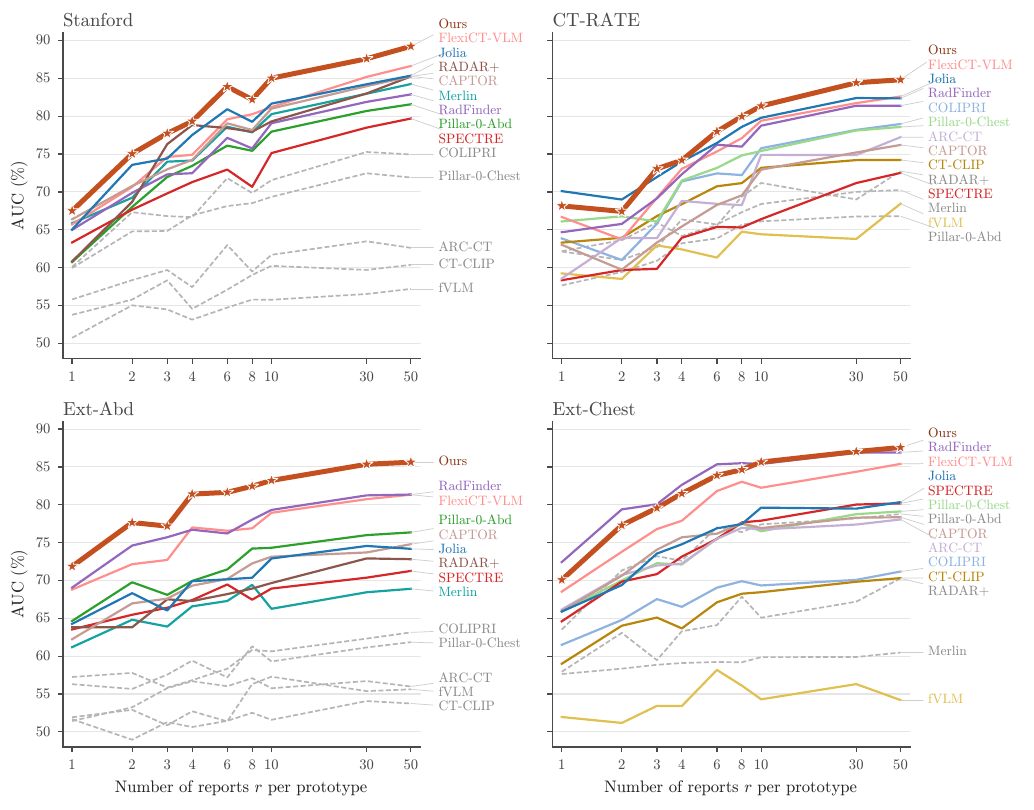}
  \caption{\textbf{Zero-shot classification with long prompts.} AUC (\%) against the number of reports $r$ per prototype, on the four test sets, on the same axis. Grey dashed curves: models whose released checkpoint was trained on the other anatomical region only.}
  \label{fig:zs_long}
\end{figure*}

With short prompts, the score of a finding is its mean cosine similarity with the positive prompts minus that with the negative ones, on normalised embeddings and without temperature. The aggregate of a set of pairs scores each finding this way over all its pairs, as in \cref{sec:zs}.
\JoliaTwo is trained on eight of these pairs through $\mathcal{L}_{\text{OSL}}$ (\cref{app:osl}). We therefore also score 29 pairs that it never sees, in six more families (\cref{tab:zs_templates}). These are CT-CLIP templates, report style, antonyms, form-like pairs, uncertain or qualified findings, and generic negatives.
We write S-OOD for these 29 pairs, the zero-shot short of the main text, and S-ID for the eight pairs seen in training. \cref{tab:zs_summary} gives both, and zero-shot long, on each test set.
On these out-of-distribution (OOD) pairs, the aggregate of \JoliaTwo over the four test sets goes from 83.2 to 82.7, compared with 81.5 for RadFinder and 81.4 for FlexiCT-VLM (\cref{fig:zs_pairs_abd,fig:zs_pairs_chest}).
\JoliaTwo is first on the abdominal sets, ARC-CT on \ctrate and RadFinder on \extchest.
\JoliaTwo drops on uncertain prompts, such as ``[label] cannot be excluded'' / ``[label] is excluded'' (51.1 / 49.3 on the abdominal / chest sets, compared with 66.8 / 77.3 for RadFinder), possibly because $\mathcal{L}_{\text{OSL}}$ contrasts a finding with its negation and never with an uncertain statement.
On the form-like pair ``[label]: yes'' / ``[label]: no'', it reaches 72.9 / 69.8, compared with 54.4 / 59.4 for RadFinder.
With long prompts (\cref{fig:zs_long}), \JoliaTwo is first on \stanford and \extabdo for any number of reports $r$ per prototype, on \ctrate from $r=3$ (\Jolia leads below), and on \extchest from $r=10$ (RadFinder leads below).

\subsection{Image-text retrieval}
\label{app:retrieval}
\cref{tab:retrieval,tab:retrieval_ext} give the Recall@1, 5 and 10 in both directions.
\JoliaTwo has the best recall at every rank, in both directions and on the four test sets.
The gap is largest at Recall@1: in image-to-text, \JoliaTwo reaches 86.3 on \stanford and 60.7 on \ctrate, compared with 70.9 for FlexiCT-VLM and 48.6 for \Jolia with [CLS], the next best.
On the external sets, the recall of every model drops, and \JoliaTwo keeps the highest Recall@1 (20.3 compared with 11.1 for the next model on \extabdo, 18.7 compared with 11.9 on \extchest).
\begin{table*}[!htbp]

    \centering
    \footnotesize
    \caption{\textbf{Image-text retrieval results.} Recall@1, @5, @10 for image-to-text (I$\rightarrow$T) and text-to-image (T$\rightarrow$I). Mean over disjoint pools of 100 test studies.}
    \label{tab:retrieval}
        \setlength{\tabcolsep}{5pt}
    \begin{adjustbox}{max width=\linewidth}   %
    \begin{tabular}{l ccc ccc ccc ccc}
        \toprule
    & \multicolumn{6}{c}{\textbf{Chest (\ctrate)}} & \multicolumn{6}{c}{\textbf{Abdomen (\stanford)}} \\
        \cmidrule(lr){2-7} \cmidrule(lr){8-13}
        & \multicolumn{3}{c}{I$\rightarrow$T} & \multicolumn{3}{c}{T$\rightarrow$I} & \multicolumn{3}{c}{I$\rightarrow$T} & \multicolumn{3}{c}{T$\rightarrow$I} \\
        \cmidrule(lr){2-4} \cmidrule(lr){5-7} \cmidrule(lr){8-10} \cmidrule(lr){11-13}
        \textbf{Model} & @1 & @5 & @10 & @1 & @5 & @10 & @1 & @5 & @10 & @1 & @5 & @10 \\
        \midrule
            CT-CLIP~\cite{hamamci2024ctclip} & 9.07 & 28.43 & 45.50 & 9.21 & 27.07 & 41.64 & 1.43 & 7.02 & 12.53 & 2.10 & 8.12 & 14.31 \\
            fVLM~\cite{fvlm} & 2.36 & 10.29 & 18.50 & 3.71 & 13.43 & 22.57 & 1.12 & 5.86 & 11.55 & 1.45 & 6.84 & 13.08 \\
            COLIPRI~\cite{colipri} & 32.71 & 62.50 & 76.86 & 35.93 & 69.07 & 81.00 & 4.92 & 18.35 & 31.14 & 8.76 & 27.92 & 42.65 \\
            SPECTRE~\cite{spectre} & 23.29 & 52.29 & 64.86 & 25.50 & 53.07 & 65.43 & 39.12 & 72.59 & 84.06 & 39.92 & 74.04 & 85.14 \\
            Merlin~\cite{blankemeier2024merlin} & 4.36 & 17.57 & 28.29 & 5.36 & 20.14 & 30.57 & 57.29 & 86.57 & 94.16 & 53.06 & 84.76 & 92.51 \\
            Pillar-0-Abd~\cite{pillar0} & 3.14 & 11.00 & 19.07 & 3.00 & 11.86 & 20.43 & 46.69 & 78.47 & 88.39 & 45.06 & 75.41 & 85.47 \\
            Pillar-0-Chest~\cite{pillar0} & 12.43 & 33.50 & 47.29 & 13.29 & 37.07 & 50.86 & 3.75 & 14.43 & 25.24 & 5.47 & 20.24 & 32.35 \\
            FlexiCT-VLM~\cite{flexictvlm} & 41.29 & 71.43 & 82.29 & 40.00 & 65.29 & 73.71 & 70.90 & 93.71 & 97.57 & 71.00 & 92.37 & 96.31 \\
            ARC-CT~\cite{arcct} & 28.57 & 62.57 & 76.64 & 23.21 & 59.21 & 75.36 & 1.92 & 8.33 & 15.43 & 2.49 & 9.63 & 18.41 \\
            CAPTOR~\cite{captor} & 22.43 & 53.36 & 68.29 & 21.21 & 50.36 & 64.86 & 59.82 & 92.63 & 98.12 & 51.49 & 87.69 & 95.75 \\
            RadFinder~\cite{radfinder} & 18.57 & 51.93 & 69.93 & 22.57 & 55.29 & 72.71 & 39.59 & 77.61 & 89.16 & 57.71 & 89.22 & 95.47 \\
            RADAR+~\cite{radar} & 4.36 & 16.93 & 28.71 & 3.71 & 18.93 & 31.21 & 15.73 & 51.76 & 70.71 & 16.06 & 52.27 & 71.41 \\
            M3D-CLIP~\cite{m3d} & 2.00 & 8.00 & 13.64 & 1.93 & 7.43 & 12.64 & 2.04 & 8.67 & 15.82 & 1.86 & 9.00 & 15.88 \\
        \midrule
            \Jolia~\cite{jolia} - [CLS] & 48.64 & 79.57 & 88.00 & 50.50 & 79.36 & 88.07 & 60.12 & 89.92 & 96.43 & 62.27 & 91.22 & 97.31 \\
            \Jolia~\cite{jolia} & 47.29 & 79.36 & 88.00 & 53.14 & 80.29 & 88.21 & 65.02 & 91.98 & 97.25 & 69.43 & 93.75 & 97.92 \\
        \midrule
            \JoliaTwo - [CLS] & 57.71 & 82.57 & 90.71 & 58.21 & 83.14 & 90.43 & 81.20 & 97.04 & 99.06 & 83.29 & 97.75 & 99.20 \\
            \textbf{\JoliaTwo} & \textbf{60.71} & \textbf{84.43} & \textbf{91.86} & \textbf{60.43} & \textbf{84.50} & \textbf{91.36} & \textbf{86.33} & \textbf{98.20} & \textbf{99.45} & \textbf{88.63} & \textbf{98.39} & \textbf{99.37} \\
        \bottomrule
    \end{tabular}
    \end{adjustbox}
\end{table*}

\begin{table*}[!htbp]

    \centering
    \footnotesize
    \caption{\textbf{Image-text retrieval on the external sets \extabdo and \extchest.} Recall@1, @5, @10 for image-to-text (I$\rightarrow$T) and text-to-image (T$\rightarrow$I), mean over disjoint pools of 100 test studies. No model saw these sets during training.}
    \label{tab:retrieval_ext}
        \setlength{\tabcolsep}{5pt}
    \begin{adjustbox}{max width=\linewidth}   %
    \begin{tabular}{l ccc ccc ccc ccc}
        \toprule
        & \multicolumn{6}{c}{\textbf{Abdomen (\extabdo)}} & \multicolumn{6}{c}{\textbf{Chest (\extchest)}} \\
        \cmidrule(lr){2-7} \cmidrule(lr){8-13}
        & \multicolumn{3}{c}{I$\rightarrow$T} & \multicolumn{3}{c}{T$\rightarrow$I} & \multicolumn{3}{c}{I$\rightarrow$T} & \multicolumn{3}{c}{T$\rightarrow$I} \\
        \cmidrule(lr){2-4} \cmidrule(lr){5-7} \cmidrule(lr){8-10} \cmidrule(lr){11-13}
        \textbf{Model} & @1 & @5 & @10 & @1 & @5 & @10 & @1 & @5 & @10 & @1 & @5 & @10 \\
        \midrule
            CT-CLIP~\cite{hamamci2024ctclip} & 1.00 & 6.86 & 12.57 & 2.29 & 9.00 & 15.00 & 3.14 & 11.19 & 19.89 & 3.54 & 12.65 & 21.03 \\
            fVLM~\cite{fvlm} & 1.14 & 5.00 & 11.43 & 1.29 & 7.00 & 11.86 & 1.35 & 5.22 & 10.95 & 2.22 & 8.70 & 14.73 \\
            COLIPRI~\cite{colipri} & 2.43 & 12.14 & 19.57 & 2.57 & 11.86 & 18.71 & 5.86 & 17.03 & 26.49 & 8.35 & 22.30 & 31.05 \\
            SPECTRE~\cite{spectre} & 8.57 & 23.71 & 31.43 & 8.00 & 22.43 & 33.71 & 9.16 & 22.32 & 32.14 & 9.51 & 23.97 & 34.22 \\
            Merlin~\cite{blankemeier2024merlin} & 6.29 & 23.00 & 35.00 & 8.00 & 21.71 & 33.00 & 3.24 & 10.27 & 16.78 & 4.00 & 12.59 & 19.14 \\
            Pillar-0-Abd~\cite{pillar0} & 10.71 & 26.86 & 38.43 & 10.57 & 26.86 & 38.00 & 4.81 & 16.38 & 26.89 & 4.35 & 15.49 & 27.00 \\
            Pillar-0-Chest~\cite{pillar0} & 2.14 & 9.43 & 16.29 & 2.29 & 8.29 & 14.86 & 7.68 & 21.22 & 32.57 & 6.95 & 20.81 & 30.86 \\
            FlexiCT-VLM~\cite{flexictvlm} & 11.14 & 30.43 & 43.57 & 13.29 & 32.43 & 44.86 & 11.46 & 27.51 & 38.73 & 12.76 & 29.00 & 38.81 \\
            ARC-CT~\cite{arcct} & 2.29 & 8.86 & 15.57 & 2.00 & 8.29 & 14.71 & 8.32 & 21.76 & 31.22 & 7.76 & 20.86 & 30.78 \\
            CAPTOR~\cite{captor} & 7.71 & 24.00 & 34.86 & 6.43 & 21.29 & 34.86 & 8.81 & 22.78 & 32.97 & 8.14 & 21.11 & 32.70 \\
            RadFinder~\cite{radfinder} & 11.00 & 32.29 & 43.86 & 10.43 & 34.29 & 44.71 & 9.49 & 27.14 & 39.19 & 12.14 & 31.51 & 44.08 \\
            RADAR+~\cite{radar} & 4.43 & 13.57 & 24.43 & 3.71 & 13.57 & 22.14 & 2.03 & 7.78 & 14.32 & 2.92 & 10.16 & 16.35 \\
            M3D-CLIP~\cite{m3d} & 1.57 & 7.57 & 12.43 & 1.57 & 5.71 & 10.86 & 3.08 & 11.68 & 20.65 & 3.51 & 13.54 & 23.30 \\
        \midrule
            \Jolia~\cite{jolia} - [CLS] & 10.29 & 22.57 & 33.57 & 7.86 & 22.00 & 33.86 & 11.92 & 27.22 & 37.84 & 10.76 & 26.68 & 36.46 \\
            \Jolia~\cite{jolia} & 7.86 & 22.00 & 33.71 & 9.57 & 23.29 & 33.86 & 8.81 & 22.22 & 32.22 & 10.97 & 24.70 & 34.32 \\
        \midrule
            \JoliaTwo - [CLS] & 20.14 & 42.86 & 54.57 & 16.29 & 36.00 & 49.29 & 18.05 & 36.46 & 47.78 & 17.41 & 34.78 & 44.92 \\
            \textbf{\JoliaTwo} & \textbf{20.29} & \textbf{45.57} & \textbf{58.57} & \textbf{18.29} & \textbf{40.57} & \textbf{50.71} & \textbf{18.68} & \textbf{37.78} & \textbf{48.51} & \textbf{17.92} & \textbf{36.24} & \textbf{46.51} \\
        \bottomrule
    \end{tabular}
    \end{adjustbox}
\end{table*}

\subsection{Summary by data type and region}
\label{app:global_split}
\begin{figure*}[!htbp]
  \centering
  \includegraphics[width=0.9\textwidth]{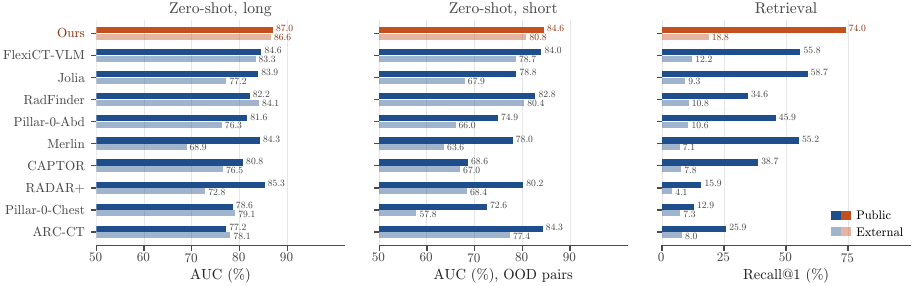}
  \caption{\textbf{Summary by data type, paired bars.} As \cref{fig:global_summary}, split by data type instead of region: the mean over the public sets (\stanford and CT-RATE, full colour) and over the external sets (\extabdo and \extchest, light). The number at the end of each bar is its value. A model trained on one region is averaged over the sets of its region only, so Pillar-0-Chest and ARC-CT use CT-RATE and \extchest, and Merlin, Pillar-0-Abd and RADAR+ use \stanford and \extabdo.}
  \label{fig:global_pairs_type}
\end{figure*}

\begin{figure*}[!htbp]
  \centering
  \includegraphics[width=0.9\textwidth]{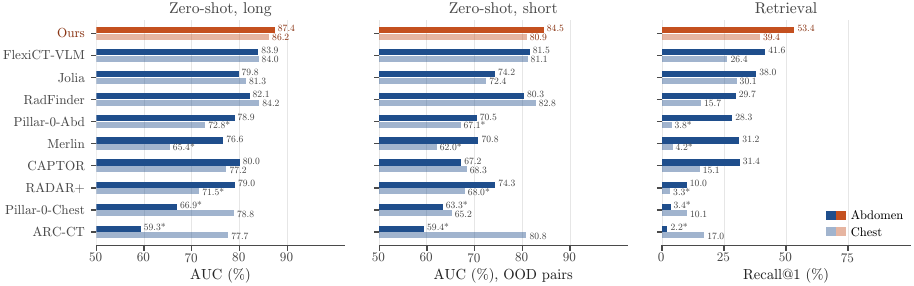}
  \caption{\textbf{Summary by anatomical region, paired bars.} \cref{fig:global_summary} at full width: the mean over the abdominal sets (\stanford and \extabdo, full colour) and over the chest sets (CT-RATE and \extchest, light). A model whose released checkpoint was trained on one region only is shown on the other region as well, with a star after the value. The number at the end of each bar is its value.}
  \label{fig:global_pairs_region}
\end{figure*}

\cref{fig:global_pairs_type} splits the summary figure (\cref{fig:global_summary}) by data type instead of anatomical region, and \cref{fig:global_pairs_region} shows it by region at full width, with a star on the models evaluated outside their region. Both have two bars per model.
The models are the ten of \cref{fig:global_summary}, in the same order. The public sets are \stanford and CT-RATE, the external ones \extabdo and \extchest. The numbers are the per-set values of \cref{tab:zs_summary,tab:retrieval,tab:retrieval_ext}, with short prompts on the 29 out-of-distribution pairs.
\JoliaTwo is first on the public and on the external sets in the three panels. By region (\cref{fig:global_pairs_region}), it is first everywhere except in zero-shot classification with short prompts on chest CT, where RadFinder leads (82.8 compared with 80.9).
In retrieval, the Recall@1 of every model drops on the external sets, from 74.0 to 18.8 for \JoliaTwo.

\section{Qualitative results}
\label{app:qualitative}
\begin{figure*}[!htbp]
  \centering
  \includegraphics[width=\textwidth]{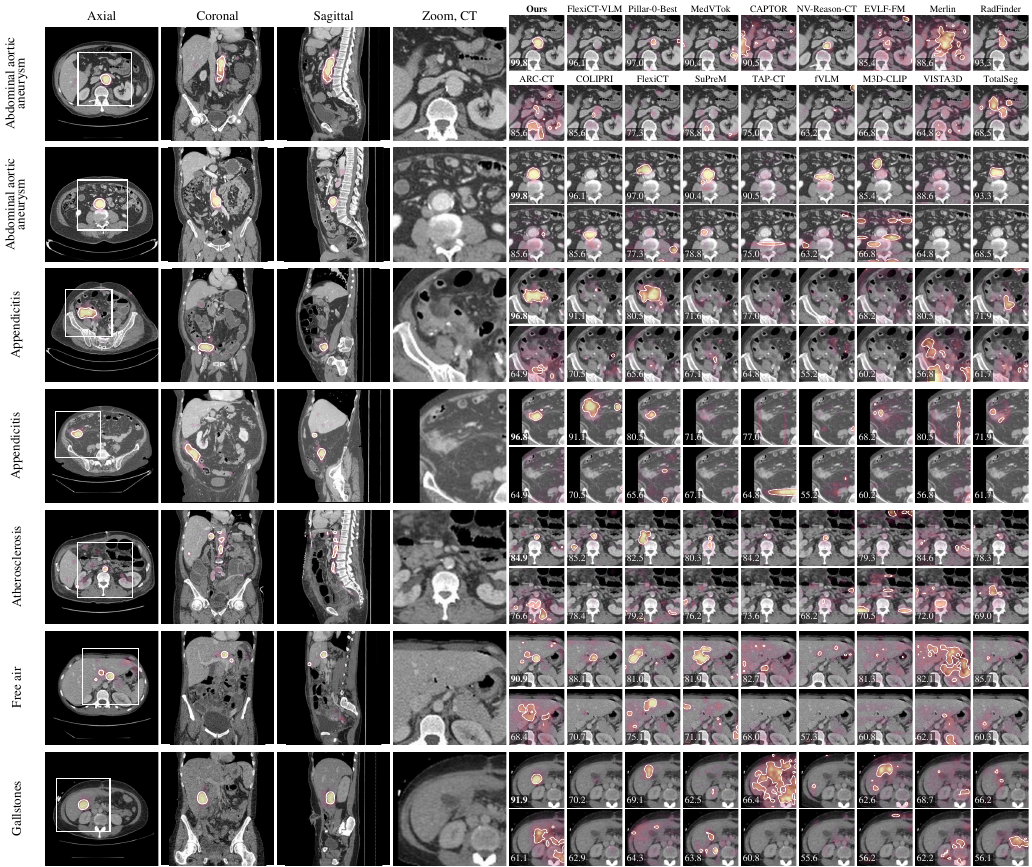}
  \caption{\textbf{Evidence of every model on the same case.} Left: gradient $\times$ input of the finding logit of \JoliaTwo (``Ours'') on the slices through its maximum on the structure of the finding, with the zoom region as a white box. Right: every model on the same axial slice and zoom, each on its own scale, as a heat map with a white line at 50\% of the maximum; the maximum of another model may lie on another slice. Number: AUC on this finding over the test set, not a score of the case. Curia and SPECTRE are not shown. Selected cases, \stanford test.}
  \label{fig:find_grid_1}
\end{figure*}

\begin{figure*}[!htbp]
  \centering
  \includegraphics[width=\textwidth]{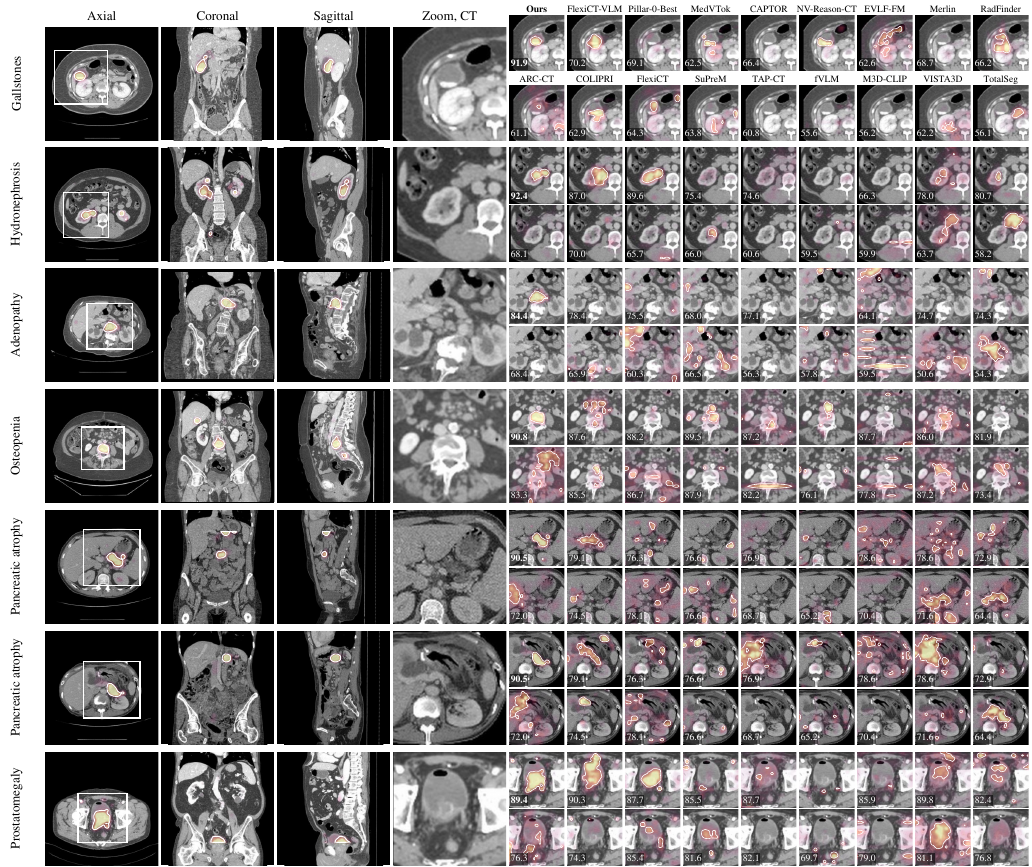}
  \caption{\textbf{Evidence of every model on the same case (continued).} Stanford test. See \cref{fig:find_grid_1}.}
  \label{fig:find_grid_2}
\end{figure*}

\begin{figure*}[!htbp]
  \centering
  \includegraphics[width=\textwidth]{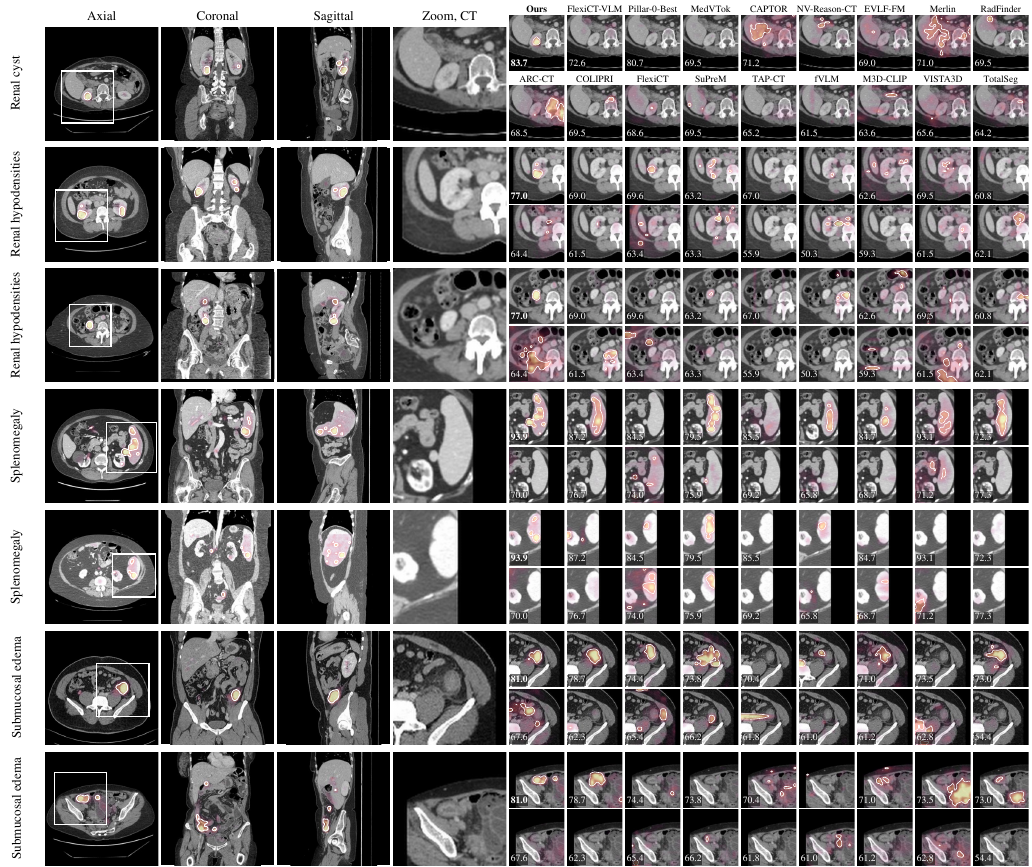}
  \caption{\textbf{Evidence of every model on the same case (continued).} Stanford test. See \cref{fig:find_grid_1}.}
  \label{fig:find_grid_3}
\end{figure*}

\begin{figure*}[!htbp]
  \centering
  \includegraphics[width=\textwidth]{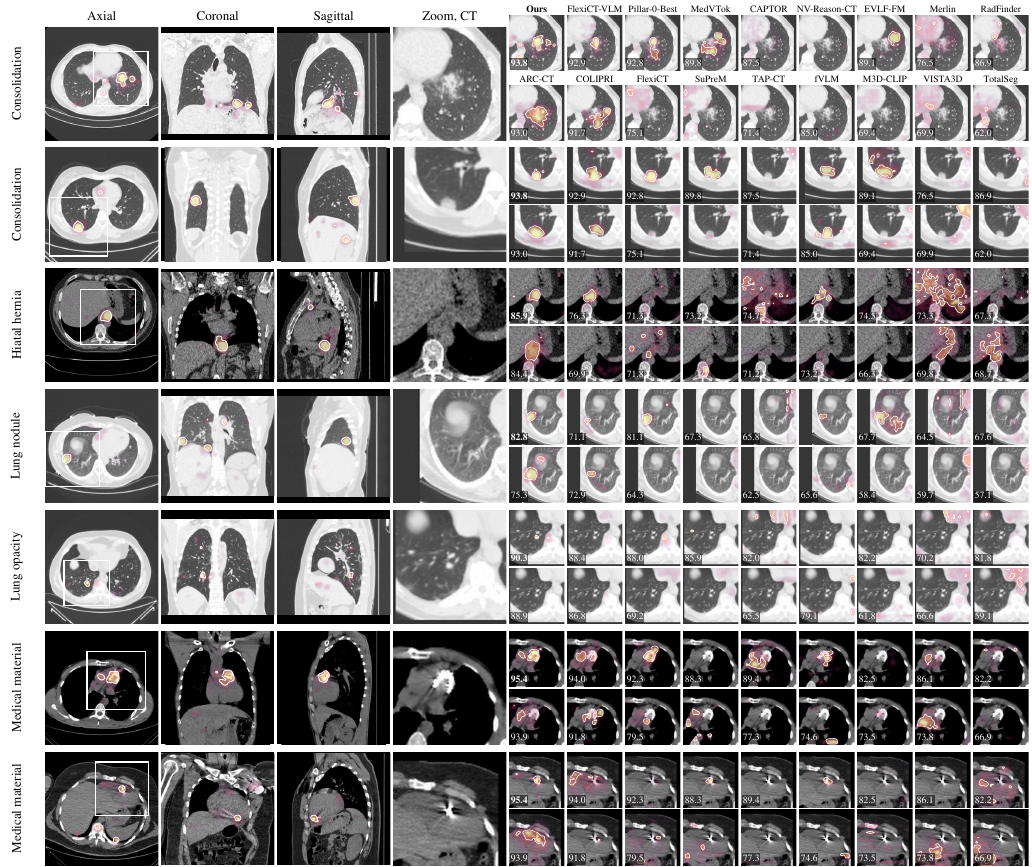}
  \caption{\textbf{Evidence of every model on the same case (continued).} \ctrate validation set. See \cref{fig:find_grid_1}.}
  \label{fig:find_grid_4}
\end{figure*}

\begin{figure*}[!htbp]
  \centering
  \includegraphics[width=\textwidth]{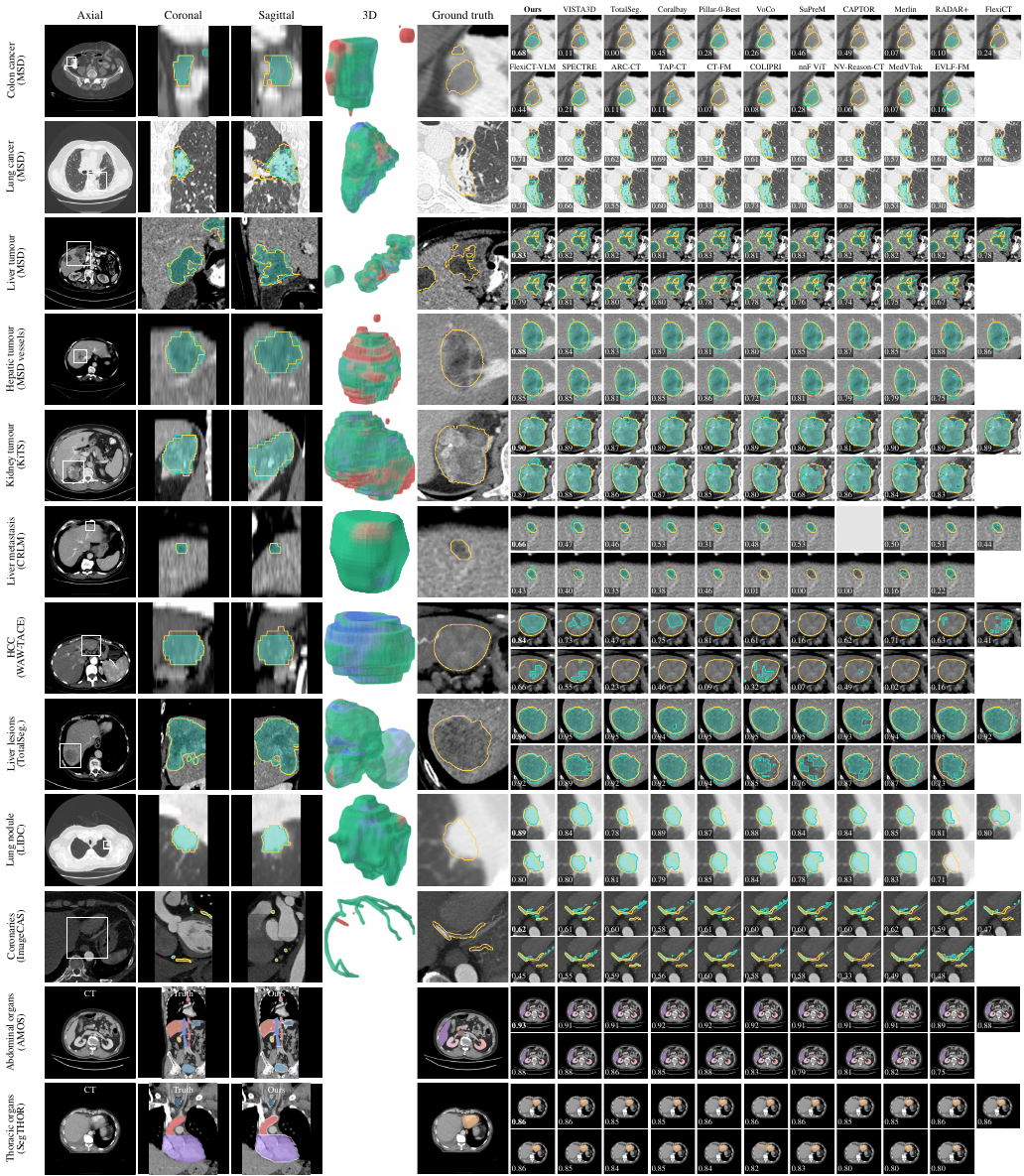}
  \caption{\textbf{Segmentation, one case per dataset.} Left: planes through the zoom and 3D surface of \JoliaTwo (``Ours''), coloured by the distance to the ground truth (blue inside, green within 1\,mm, red outside), then the ground truth. Right: every model on the same zoom, frozen encoder with a trained decoder (\cref{app:dense}); prediction in teal, ground truth in yellow. Number: 3D Dice of the case over the whole volume. Grey: prediction not exported. Curia is not shown. Selected cases; means over all cases are in \cref{tab:dense}. Organ datasets: the coronal slices show the ground truth and \JoliaTwo, the zoom is the whole axial slice, and the number is the mean Dice over organs. ImageCAS: 10\,mm maximum intensity projection.}
  \label{fig:seg_grid_all}
\end{figure*}

The cases of this section were selected to illustrate the behaviour of the models and do not replace the means of the main text.
\cref{fig:find_grid_1,fig:find_grid_2,fig:find_grid_3,fig:find_grid_4} show where each model reads a finding, and \cref{fig:seg_grid_all} one segmentation case per dataset, with every model.

\paragraph{Attribution maps.}
An attribution map shows the voxels that contribute most to the score of a model for a finding.
The maps of \cref{fig:qualitative_main,fig:find_grid_1,fig:find_grid_2,fig:find_grid_3,fig:find_grid_4} are gradient $\times$ input attributions~\cite{simonyan2014saliency,shrikumar2016notjust,ancona2018towards}.
For each model, the score of a finding is the logit of its linear head (\cref{sec:res_findings}), on the frozen encoder. We use the saved head of the evaluation and check that it gives the test probabilities of the evaluation on each case.
With $x$ the input of the network after the preprocessing of the model and $\ell$ the logit, the map is $\lvert x \odot \partial \ell / \partial x \rvert$, summed over the input channels and smoothed.
No model receives any localisation signal for findings, and the maps use no mask.
Each map is then placed back in the original scan. For \JoliaTwo, we invert its preprocessing exactly. For the other models, an automatic registration finds the permutation and flips of the axes, then a scale and an offset per axis, that maximise the correlation between the input of the model and the scan.
The slices pass through the strongest point of the map of \JoliaTwo on the structure of the finding, given by its TotalSegmentator mask~\cite{wasserthal2023totalsegmentator}. Each map is scaled to its own maximum.
Gradient $\times$ input is noisy, can miss a feature once the response of the network saturates~\cite{shrikumar2016notjust}, and a map can look plausible without reflecting what the model uses~\cite{adebayo2018sanity}. These maps show where the evidence lies on a few cases and do not measure localisation.

\FloatBarrier

\subsection{Dense features}
\label{app:pca}
\begin{figure*}[!htbp]
    \centering
    \setlength{\tabcolsep}{1pt}
    \renewcommand{\arraystretch}{1.1}
    \newlength{\imgwidth}
    \setlength{\imgwidth}{\dimexpr (\textwidth - 14\tabcolsep) / 8 \relax}
    \newcommand{\modelhdr}[1]{\footnotesize\shortstack{#1}}

    \begin{tabular}{@{} *{8}{c} @{}}
        \modelhdr{CT} &
        \modelhdr{\ours{}} &
        \modelhdr{\Jolia} &
        \modelhdr{\FlexiCT} &
        \modelhdr{\RadFinder} &
        \modelhdr{\PillarZero} &
        \modelhdr{\CAPTOR} &
        \modelhdr{\Merlin} \\[2pt]

        \includegraphics[width=\imgwidth]{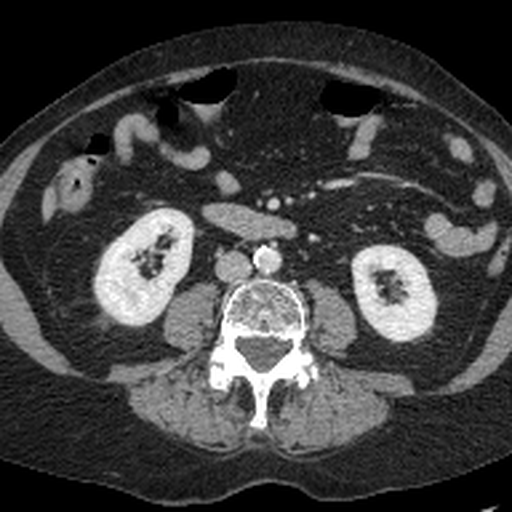} &
        \includegraphics[width=\imgwidth]{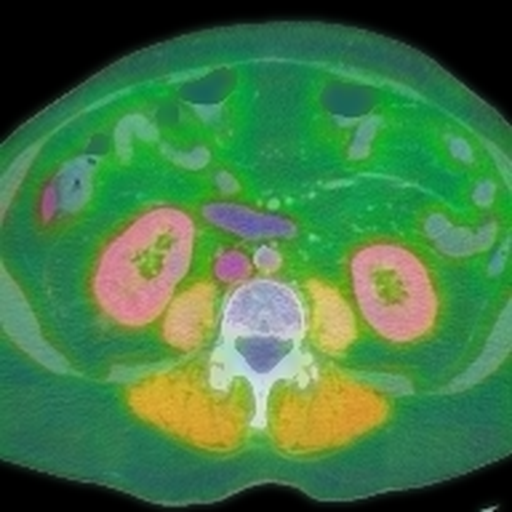} &
        \includegraphics[width=\imgwidth]{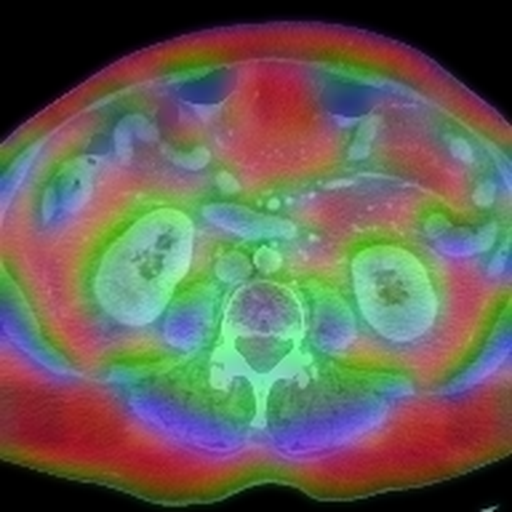} &
        \includegraphics[width=\imgwidth]{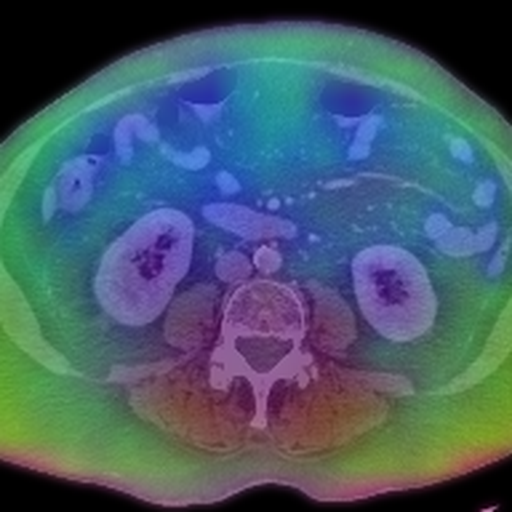} &
        \includegraphics[width=\imgwidth]{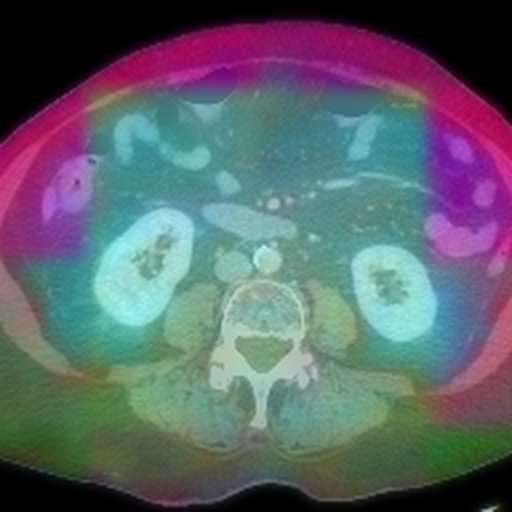} &
        \includegraphics[width=\imgwidth]{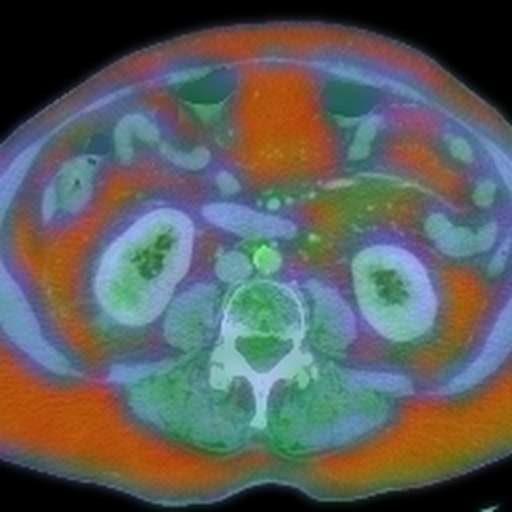} &
        \includegraphics[width=\imgwidth]{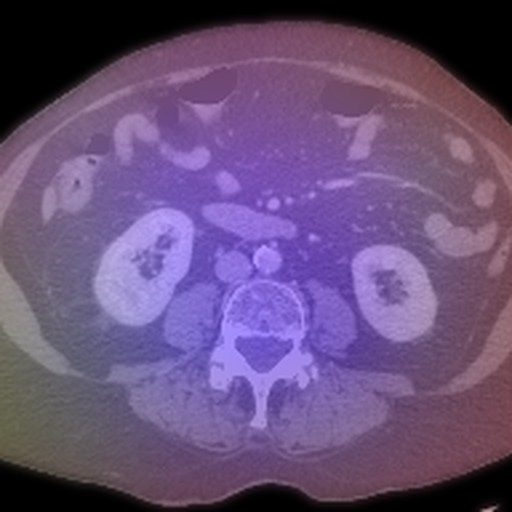} &
        \includegraphics[width=\imgwidth]{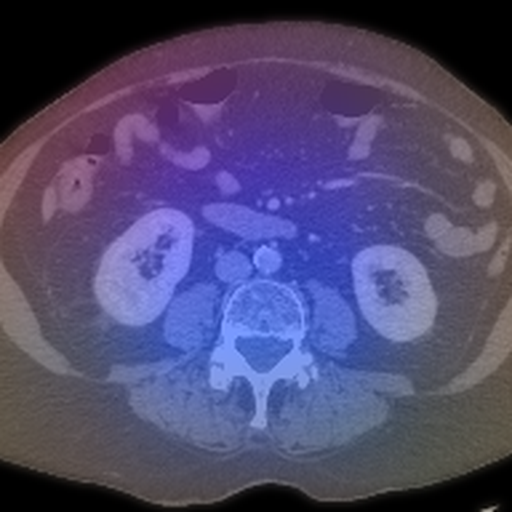} \\

        \includegraphics[width=\imgwidth]{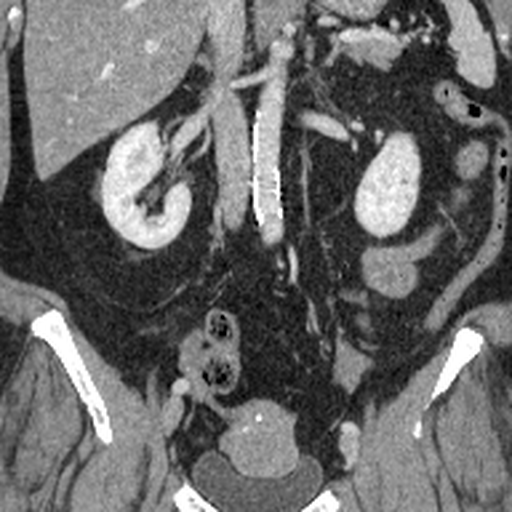} &
        \includegraphics[width=\imgwidth]{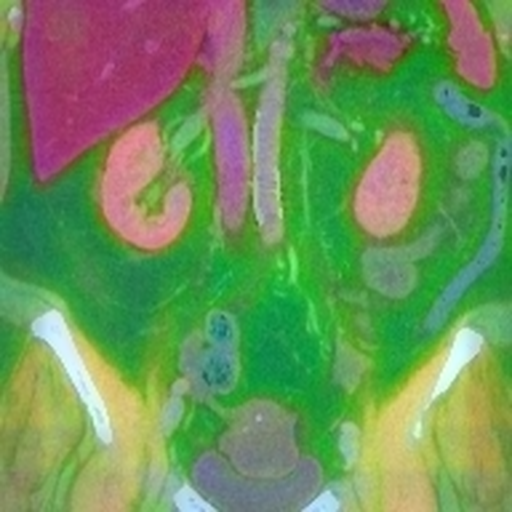} &
        \includegraphics[width=\imgwidth]{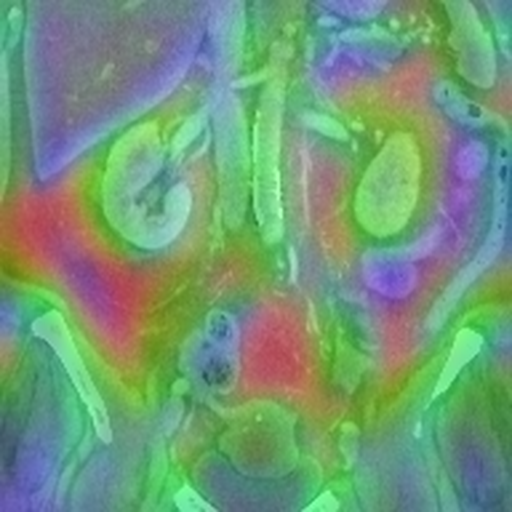} &
        \includegraphics[width=\imgwidth]{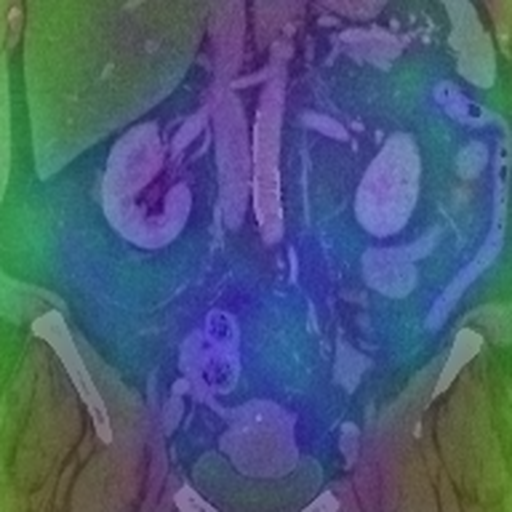} &
        \includegraphics[width=\imgwidth]{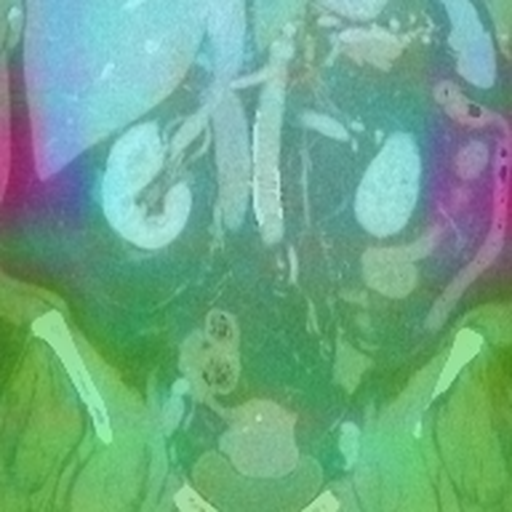} &
        \includegraphics[width=\imgwidth]{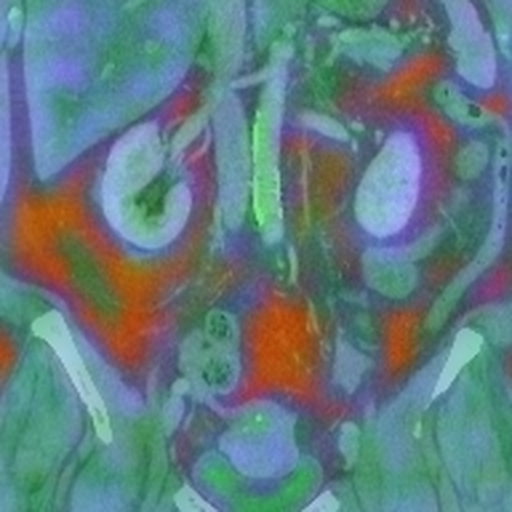} &
        \includegraphics[width=\imgwidth]{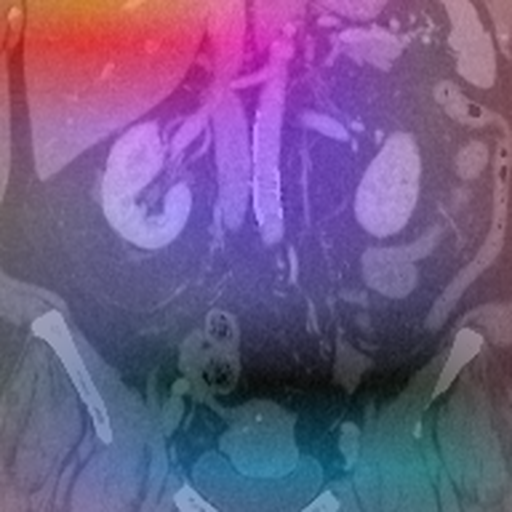} &
        \includegraphics[width=\imgwidth]{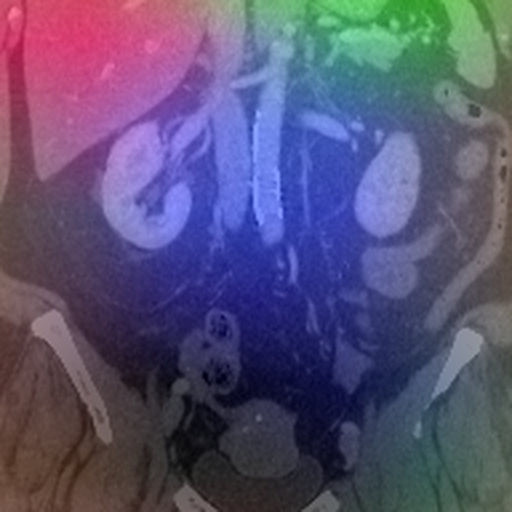} \\

        \includegraphics[width=\imgwidth]{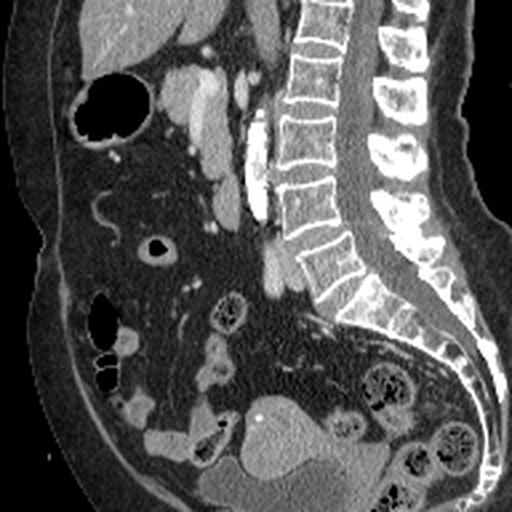} &
        \includegraphics[width=\imgwidth]{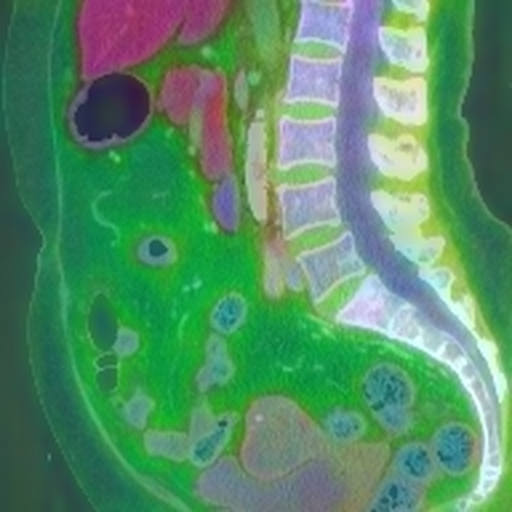} &
        \includegraphics[width=\imgwidth]{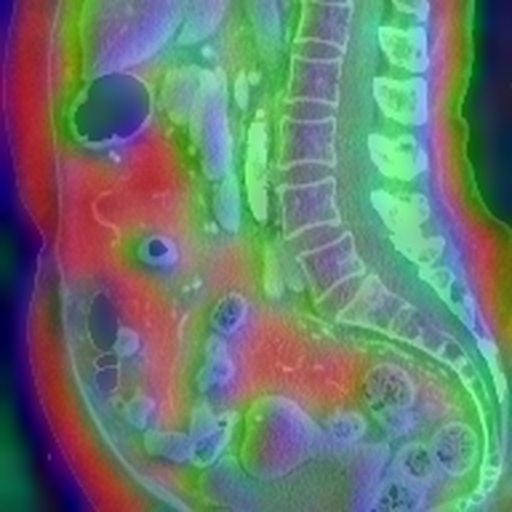} &
        \includegraphics[width=\imgwidth]{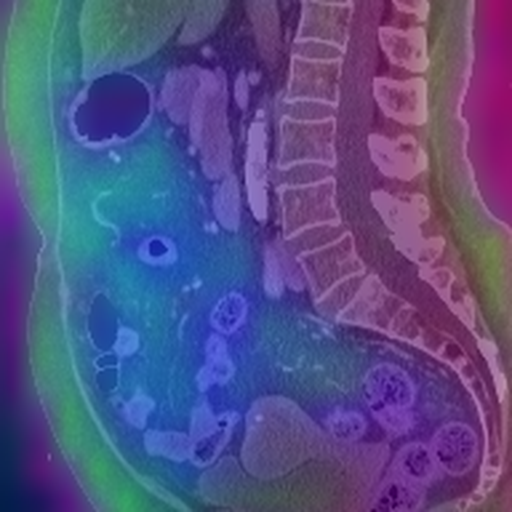} &
        \includegraphics[width=\imgwidth]{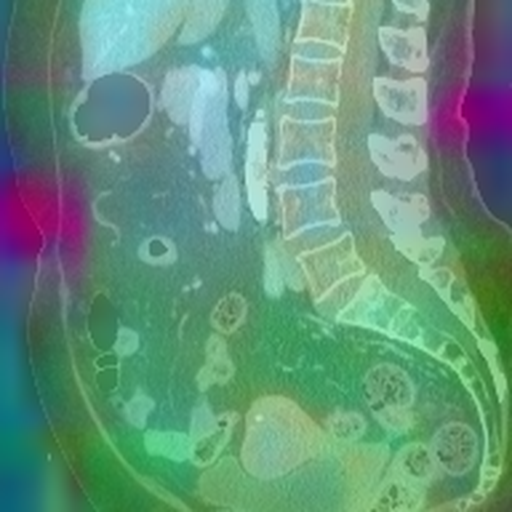} &
        \includegraphics[width=\imgwidth]{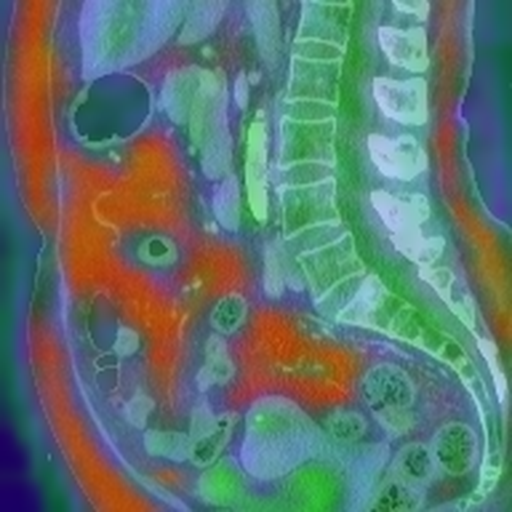} &
        \includegraphics[width=\imgwidth]{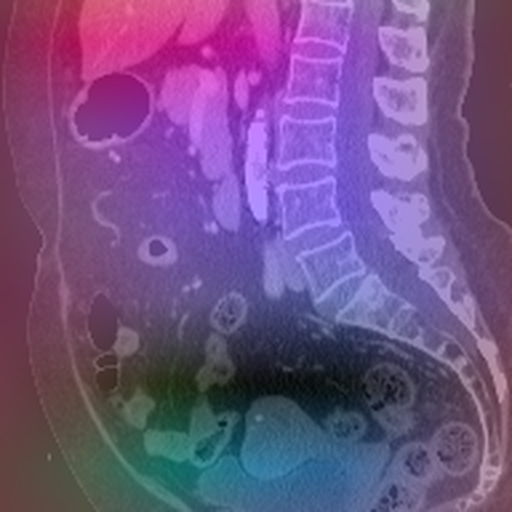} &
        \includegraphics[width=\imgwidth]{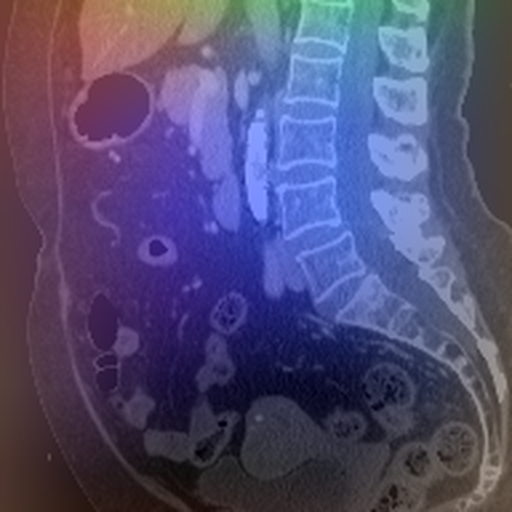} \\

        \includegraphics[width=\imgwidth]{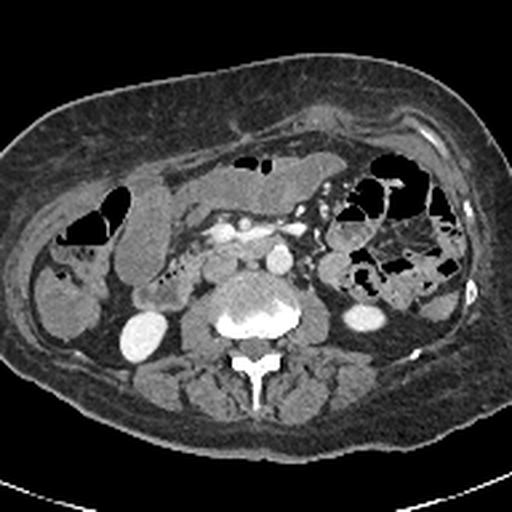} &
        \includegraphics[width=\imgwidth]{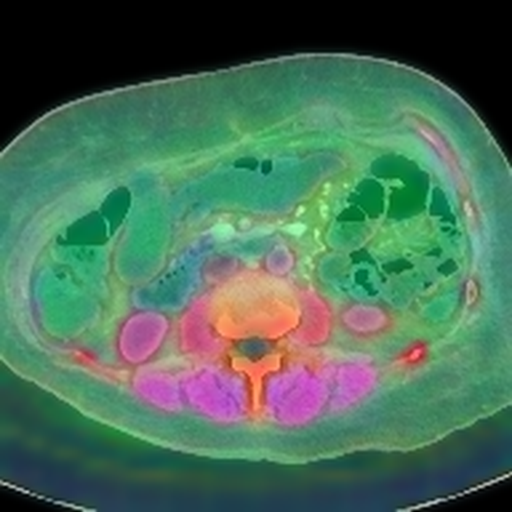} &
        \includegraphics[width=\imgwidth]{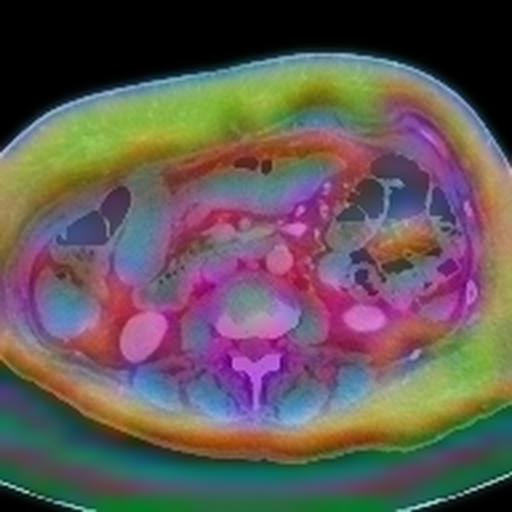} &
        \includegraphics[width=\imgwidth]{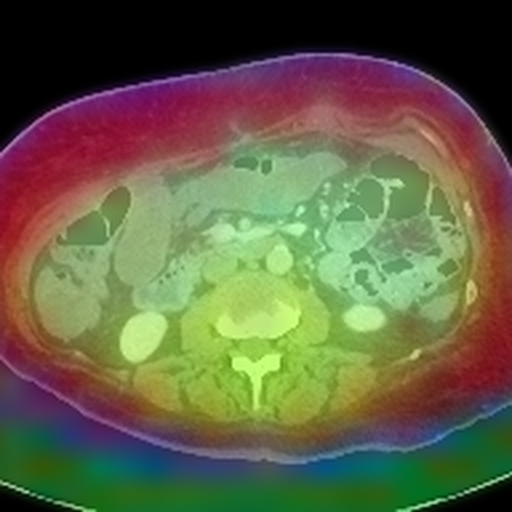} &
        \includegraphics[width=\imgwidth]{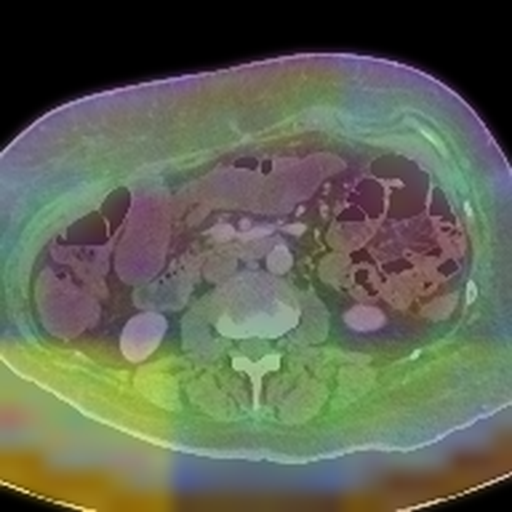} &
        \includegraphics[width=\imgwidth]{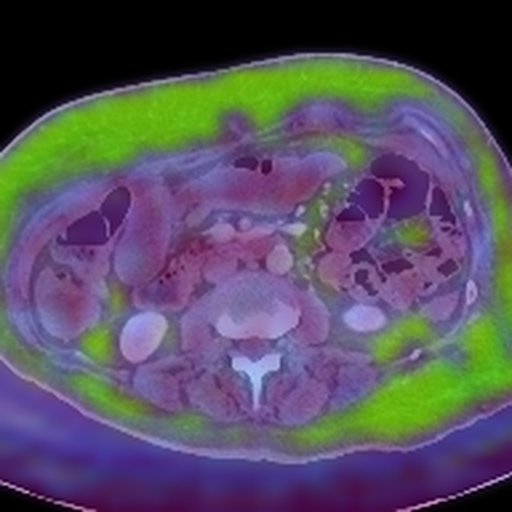} &
        \includegraphics[width=\imgwidth]{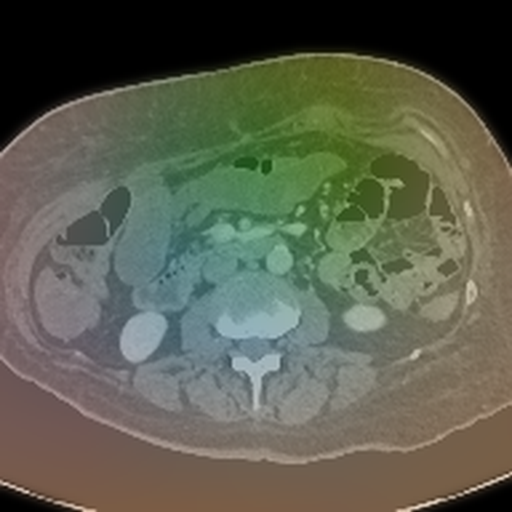} &
        \includegraphics[width=\imgwidth]{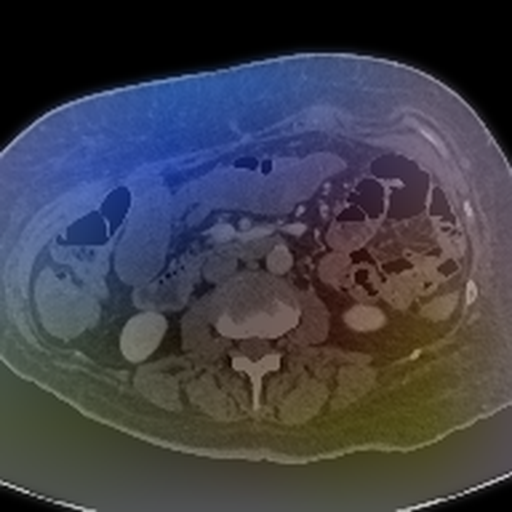} \\

        \includegraphics[width=\imgwidth]{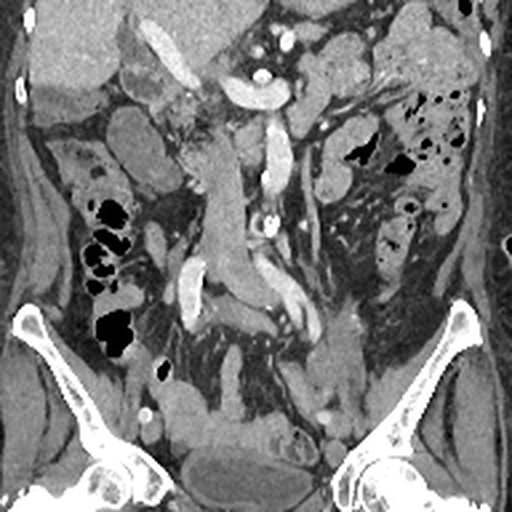} &
        \includegraphics[width=\imgwidth]{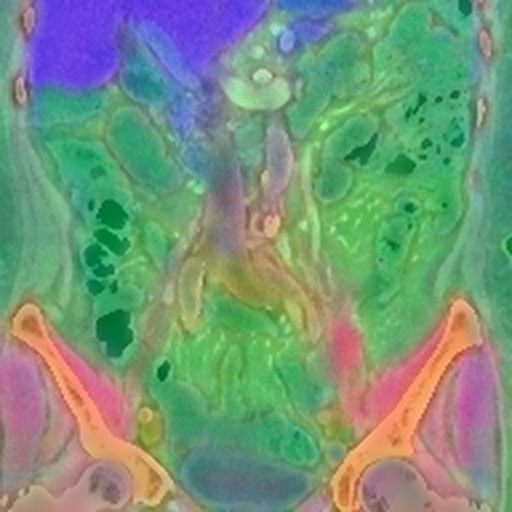} &
        \includegraphics[width=\imgwidth]{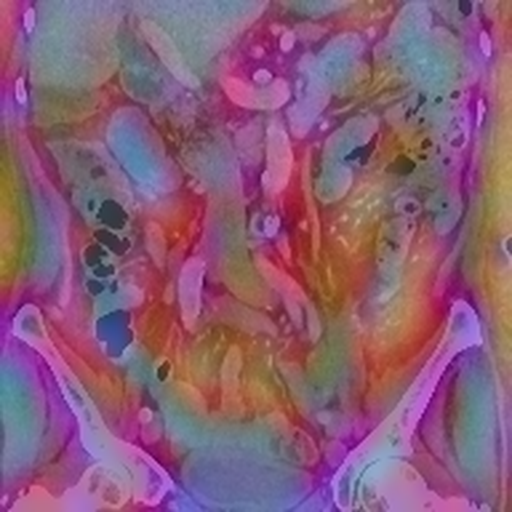} &
        \includegraphics[width=\imgwidth]{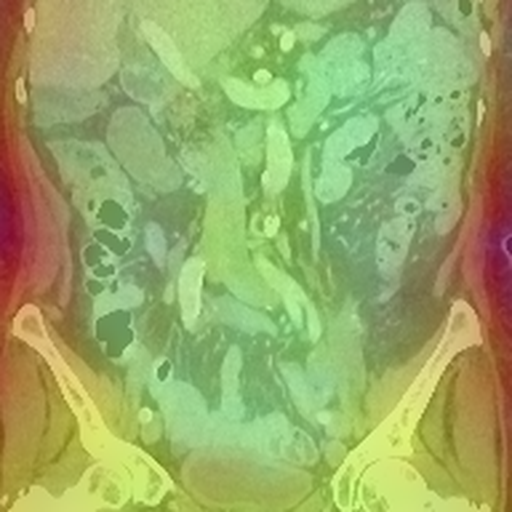} &
        \includegraphics[width=\imgwidth]{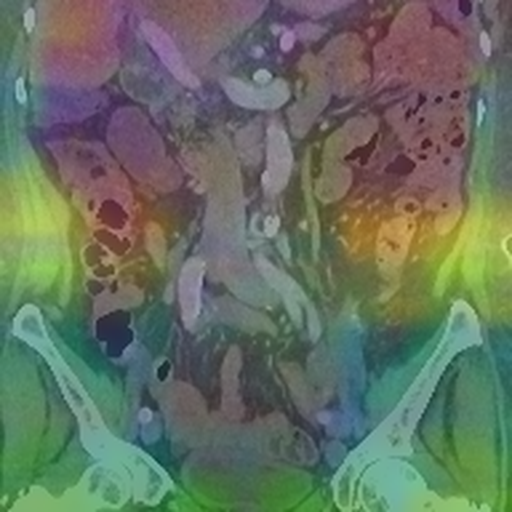} &
        \includegraphics[width=\imgwidth]{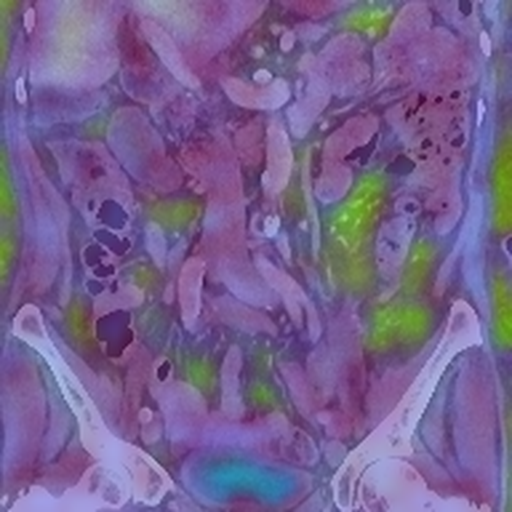} &
        \includegraphics[width=\imgwidth]{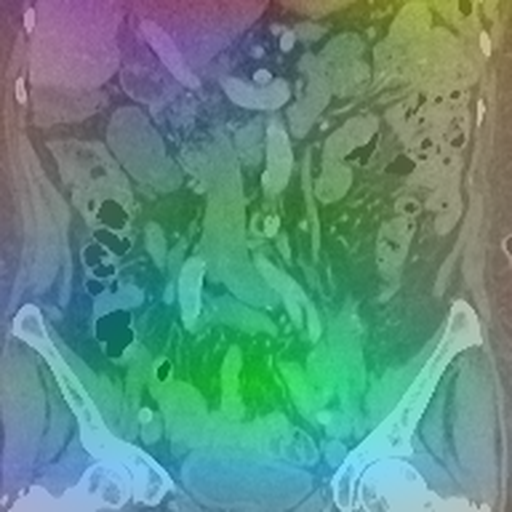} &
        \includegraphics[width=\imgwidth]{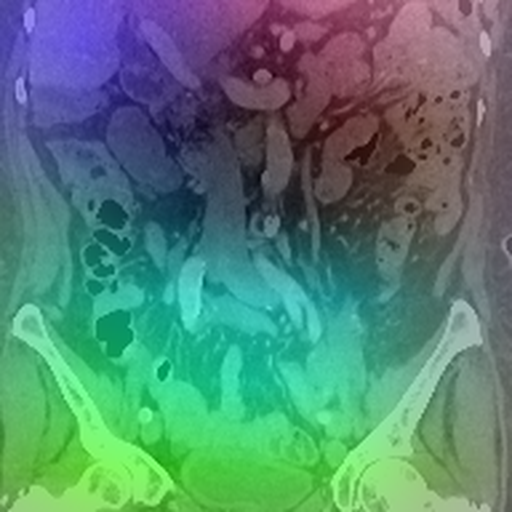} \\

        \includegraphics[width=\imgwidth]{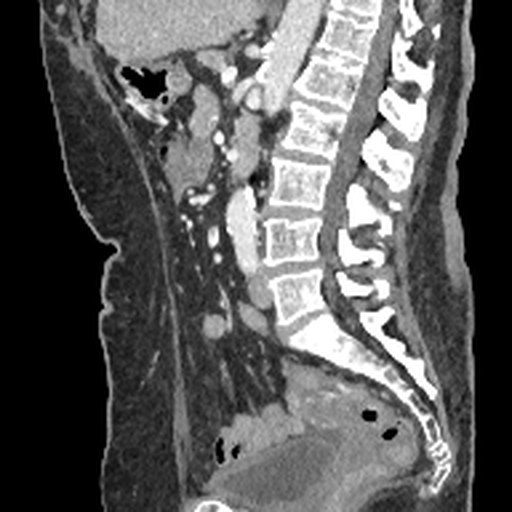} &
        \includegraphics[width=\imgwidth]{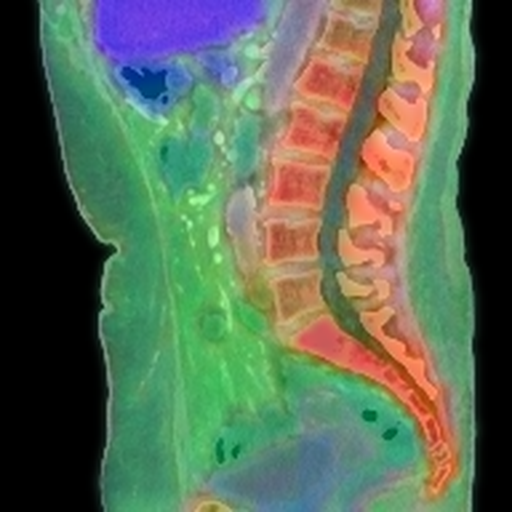} &
        \includegraphics[width=\imgwidth]{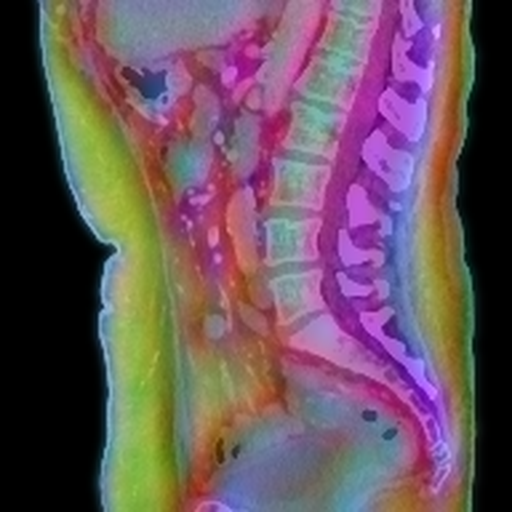} &
        \includegraphics[width=\imgwidth]{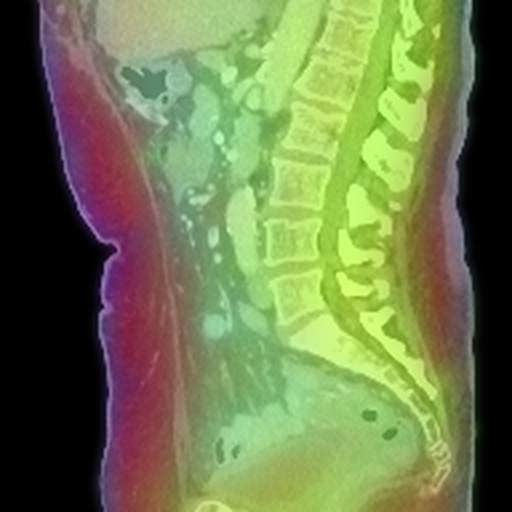} &
        \includegraphics[width=\imgwidth]{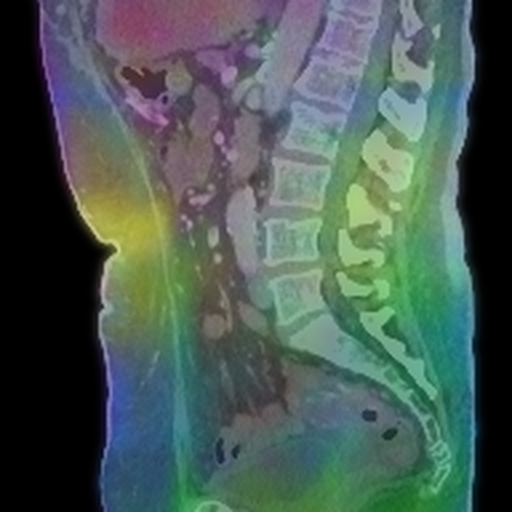} &
        \includegraphics[width=\imgwidth]{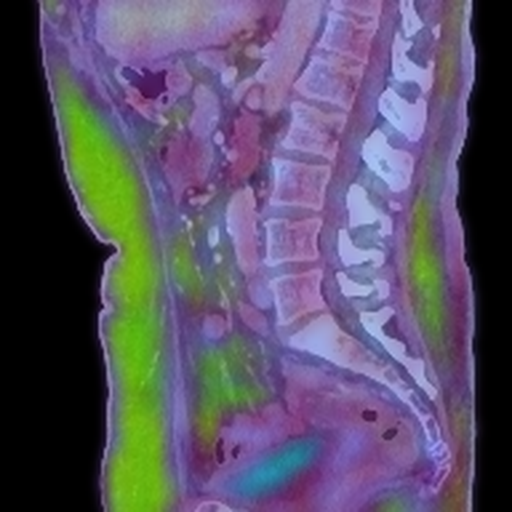} &
        \includegraphics[width=\imgwidth]{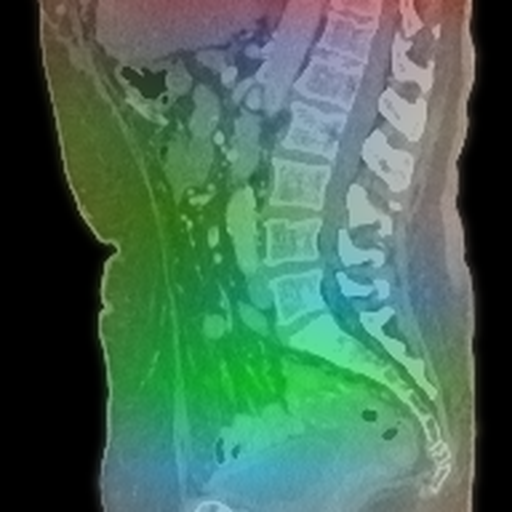} &
        \includegraphics[width=\imgwidth]{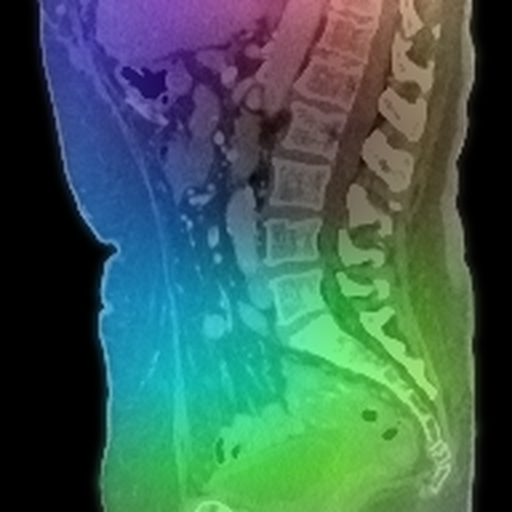} \\

    \end{tabular}
    \caption{\textbf{Principal component analysis of the dense features.} Per volume and model, PCA of the finest patch-token map (PC1 dropped, PC2--4 mapped to RGB between the 2nd and 98th percentiles). Each competitor's token grid is registered onto the same $288$\,mm crop as ours, so that every cell shows the same slice and field of view, overlaid on the CT (first column). Two abdominal volumes of \stanford, axial, coronal and sagittal slices. Colours are not comparable across cells; only the spatial structure is.}
    \label{fig:pca}
\end{figure*}

\begin{figure*}[!htbp]
    \centering
    \setlength{\tabcolsep}{1pt}
    \renewcommand{\arraystretch}{1.1}
    \setlength{\imgwidth}{\dimexpr (\textwidth - 14\tabcolsep) / 8 \relax}
    \providecommand{\modelhdr}[1]{\footnotesize\shortstack{#1}}  %

    \begin{tabular}{@{} *{8}{c} @{}}
        \modelhdr{CT} &
        \modelhdr{\ours{}} &
        \modelhdr{\Jolia} &
        \modelhdr{\FlexiCT} &
        \modelhdr{\RadFinder} &
        \modelhdr{\PillarZero} &
        \modelhdr{\COLIPRI} &
        \modelhdr{\Merlin} \\[2pt]

        \includegraphics[width=\imgwidth]{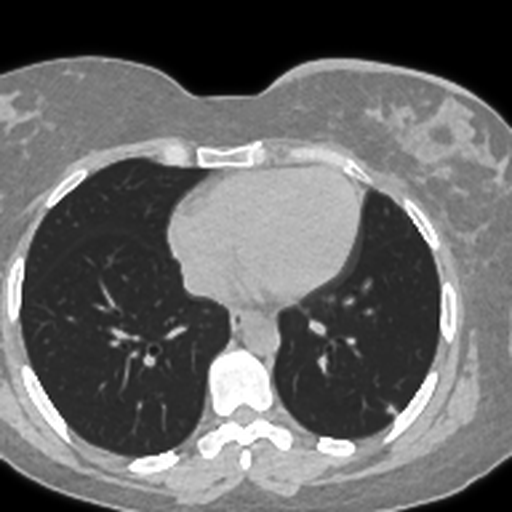} &
        \includegraphics[width=\imgwidth]{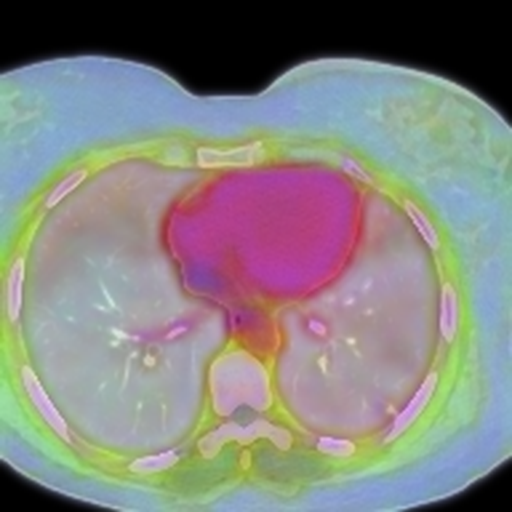} &
        \includegraphics[width=\imgwidth]{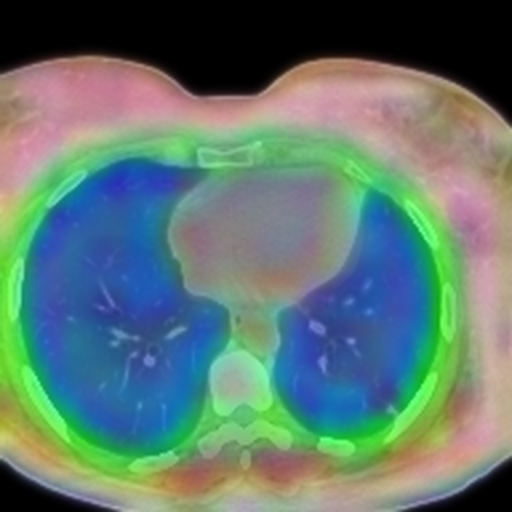} &
        \includegraphics[width=\imgwidth]{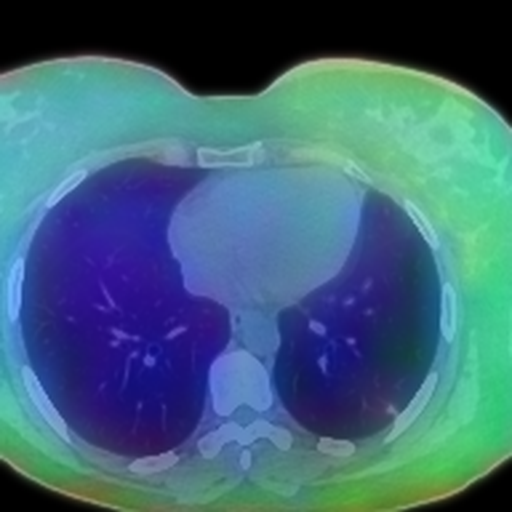} &
        \includegraphics[width=\imgwidth]{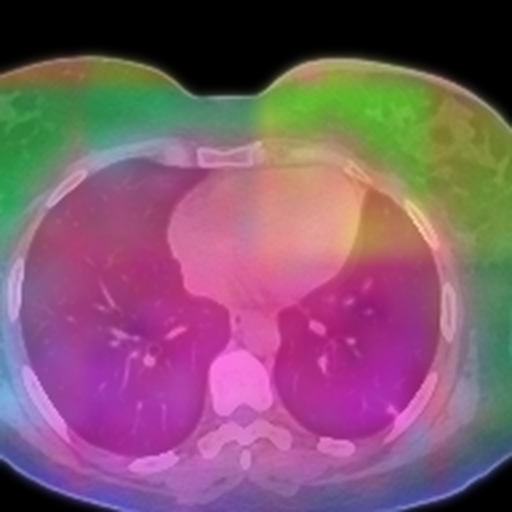} &
        \includegraphics[width=\imgwidth]{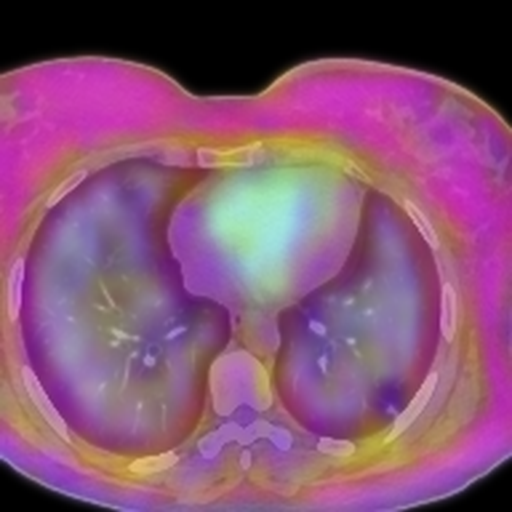} &
        \includegraphics[width=\imgwidth]{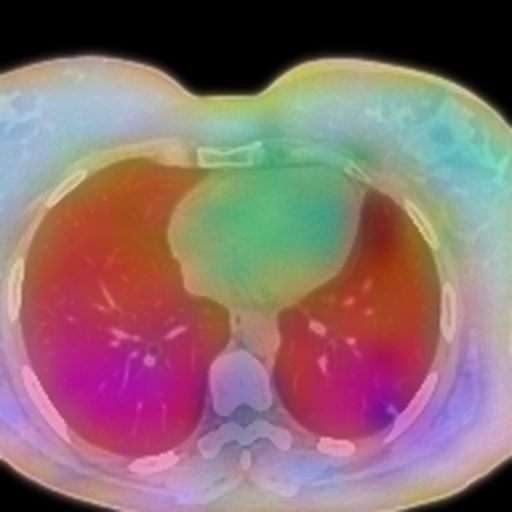} &
        \includegraphics[width=\imgwidth]{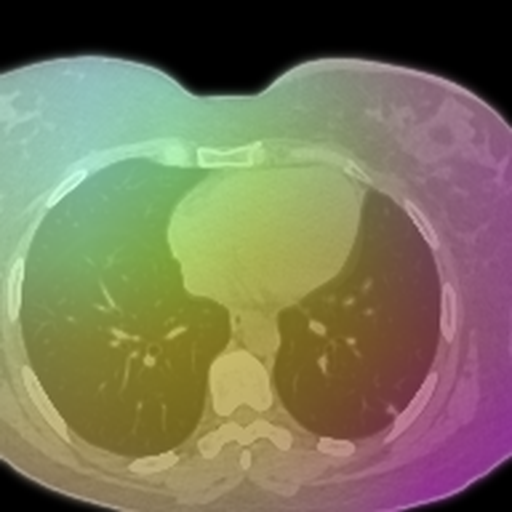} \\

        \includegraphics[width=\imgwidth]{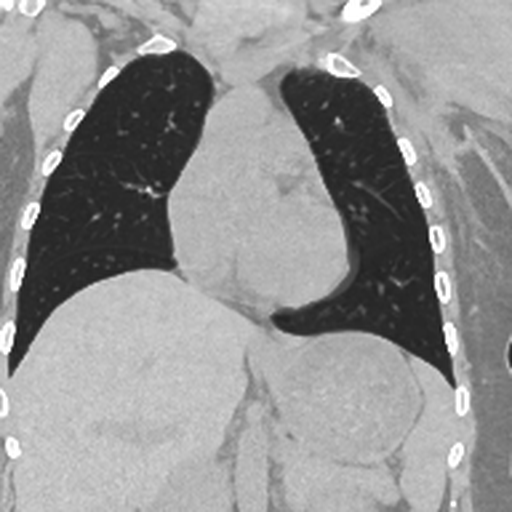} &
        \includegraphics[width=\imgwidth]{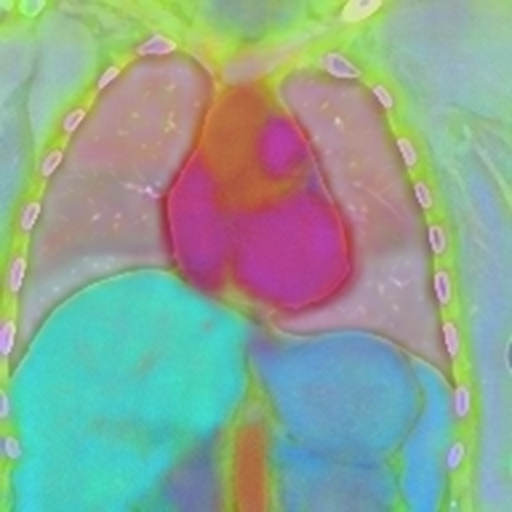} &
        \includegraphics[width=\imgwidth]{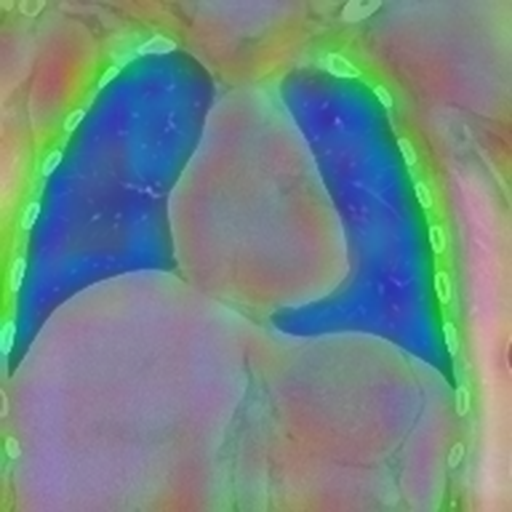} &
        \includegraphics[width=\imgwidth]{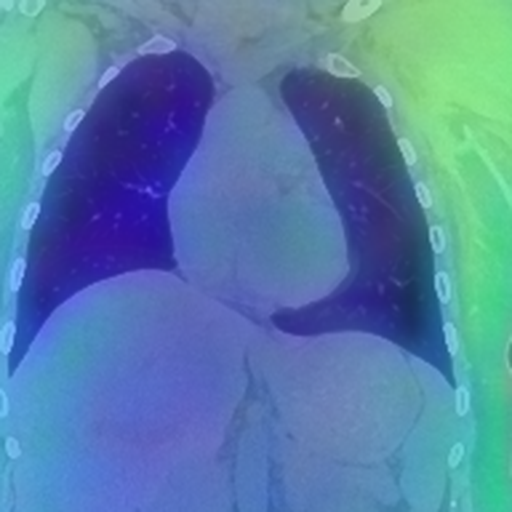} &
        \includegraphics[width=\imgwidth]{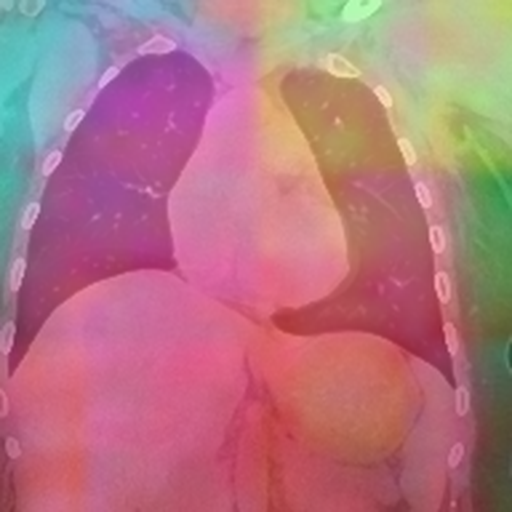} &
        \includegraphics[width=\imgwidth]{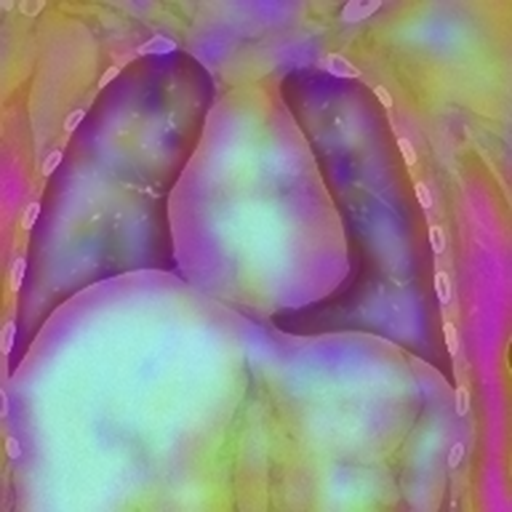} &
        \includegraphics[width=\imgwidth]{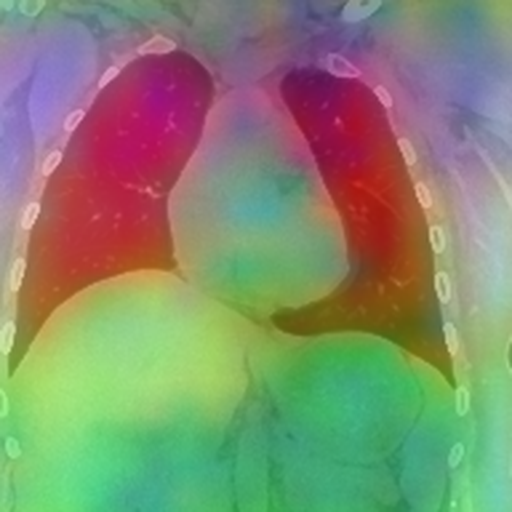} &
        \includegraphics[width=\imgwidth]{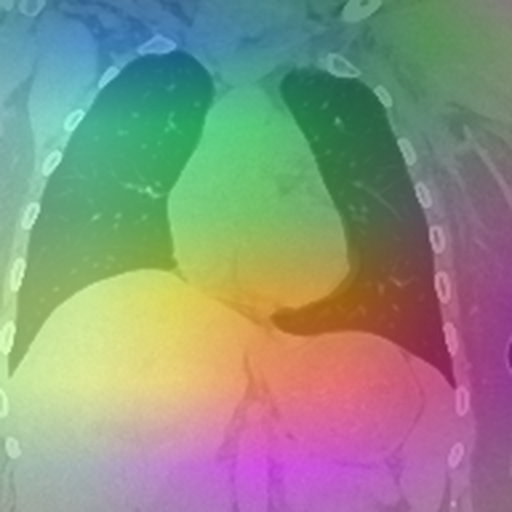} \\

        \includegraphics[width=\imgwidth]{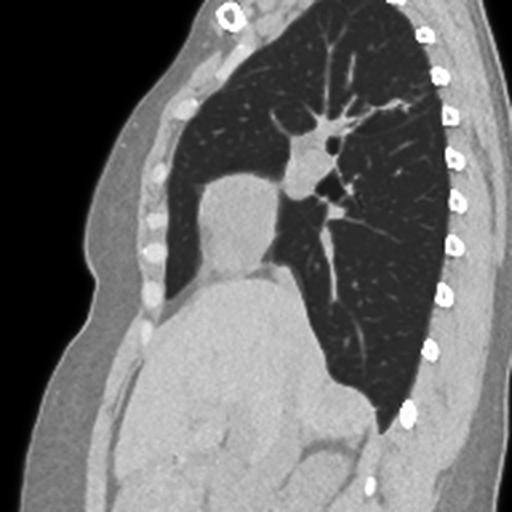} &
        \includegraphics[width=\imgwidth]{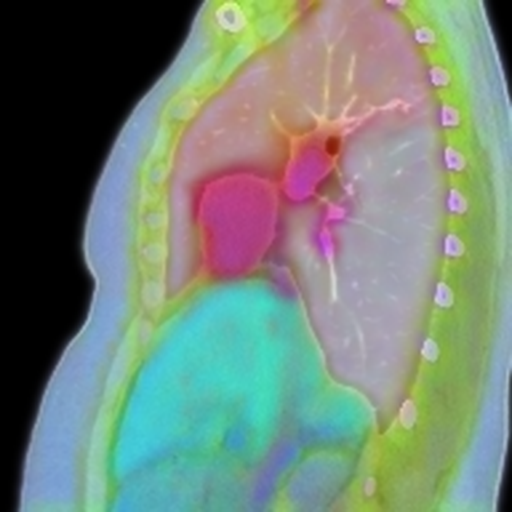} &
        \includegraphics[width=\imgwidth]{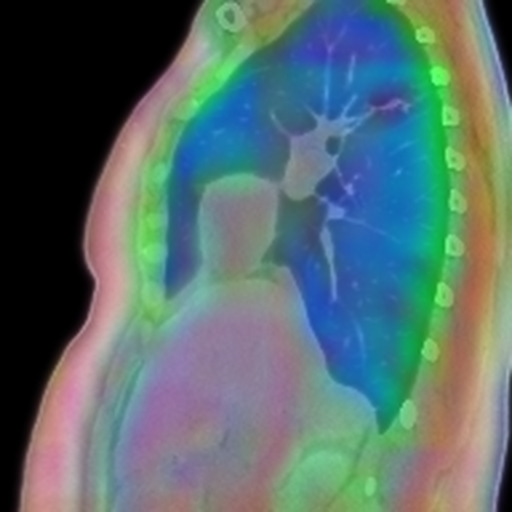} &
        \includegraphics[width=\imgwidth]{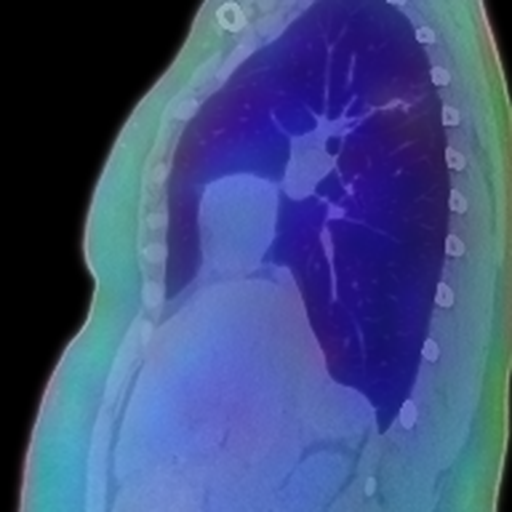} &
        \includegraphics[width=\imgwidth]{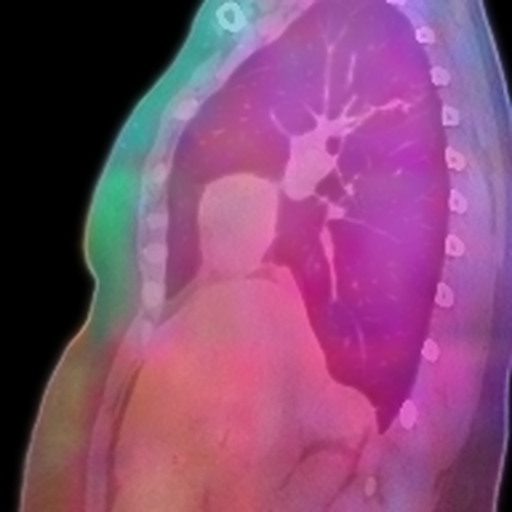} &
        \includegraphics[width=\imgwidth]{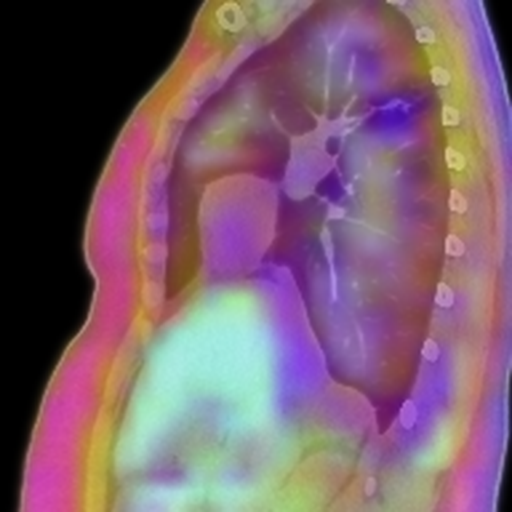} &
        \includegraphics[width=\imgwidth]{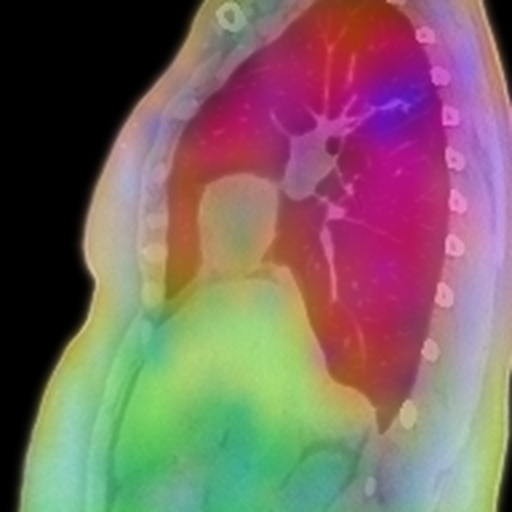} &
        \includegraphics[width=\imgwidth]{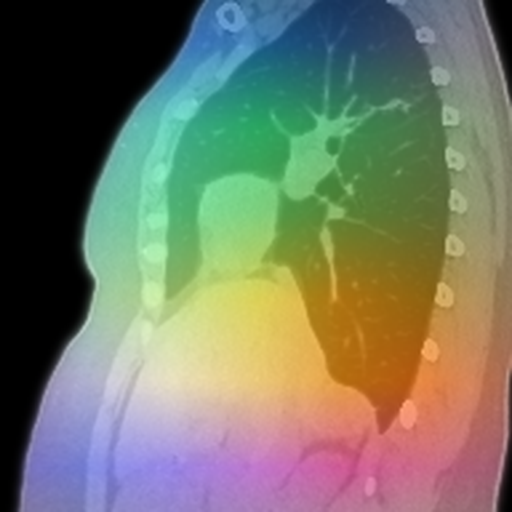} \\

        \includegraphics[width=\imgwidth]{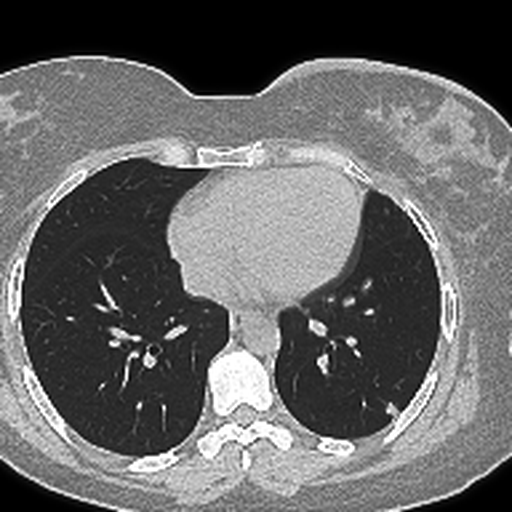} &
        \includegraphics[width=\imgwidth]{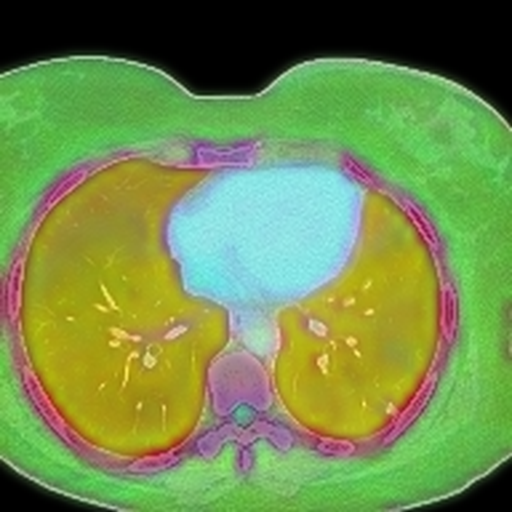} &
        \includegraphics[width=\imgwidth]{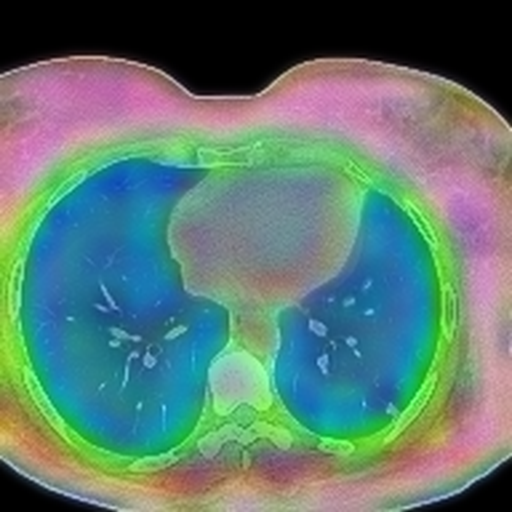} &
        \includegraphics[width=\imgwidth]{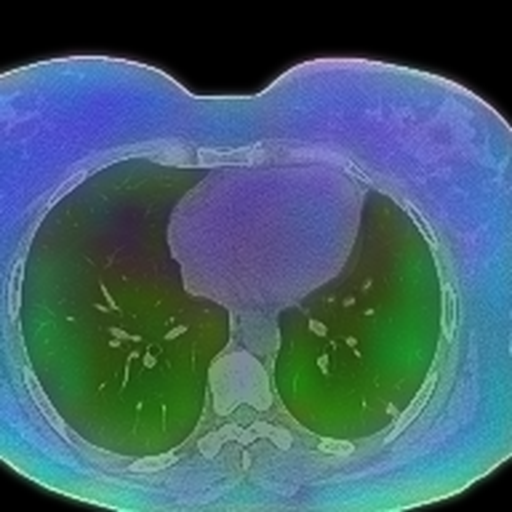} &
        \includegraphics[width=\imgwidth]{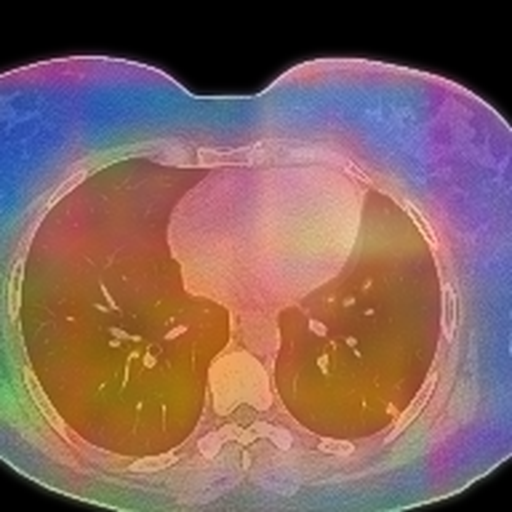} &
        \includegraphics[width=\imgwidth]{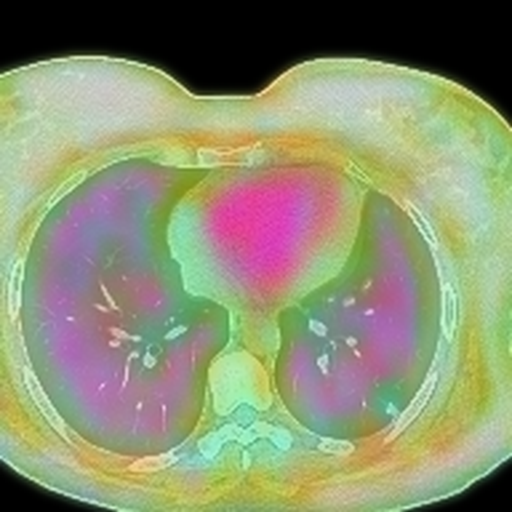} &
        \includegraphics[width=\imgwidth]{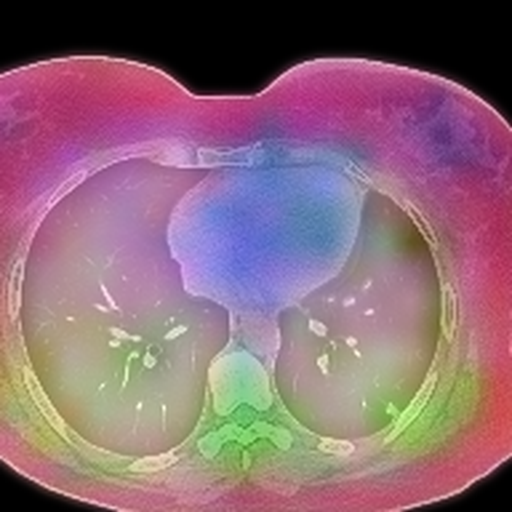} &
        \includegraphics[width=\imgwidth]{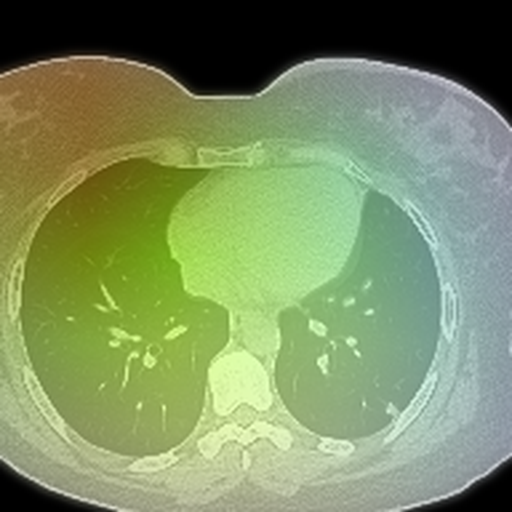} \\

        \includegraphics[width=\imgwidth]{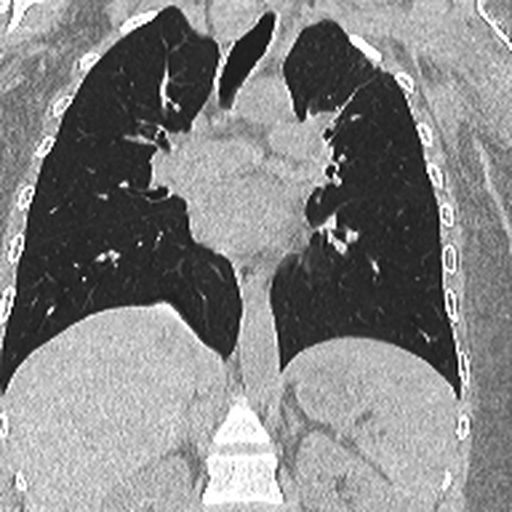} &
        \includegraphics[width=\imgwidth]{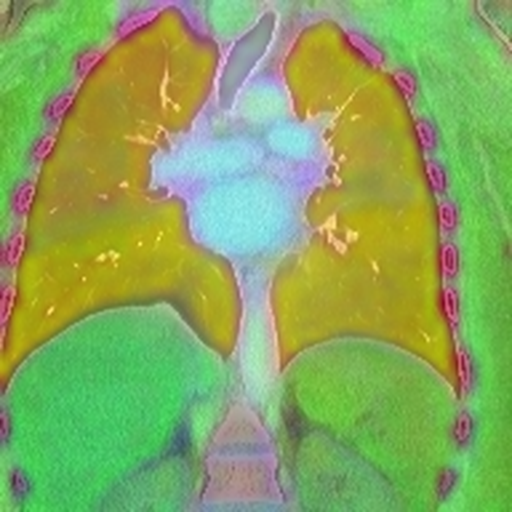} &
        \includegraphics[width=\imgwidth]{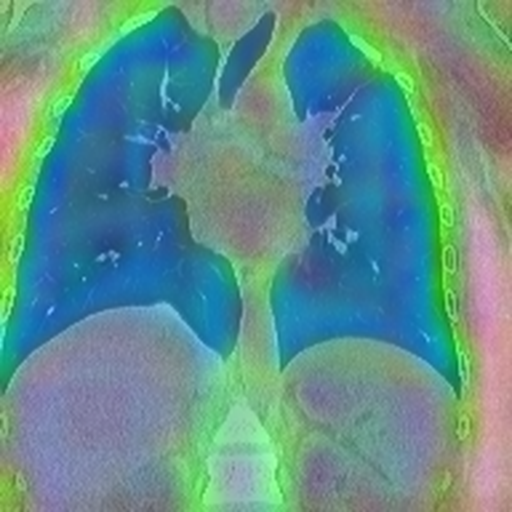} &
        \includegraphics[width=\imgwidth]{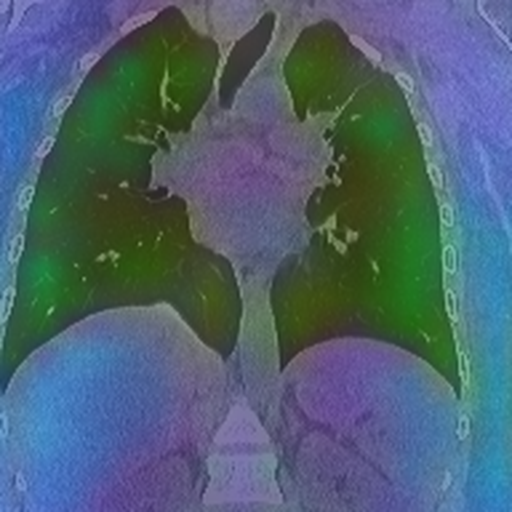} &
        \includegraphics[width=\imgwidth]{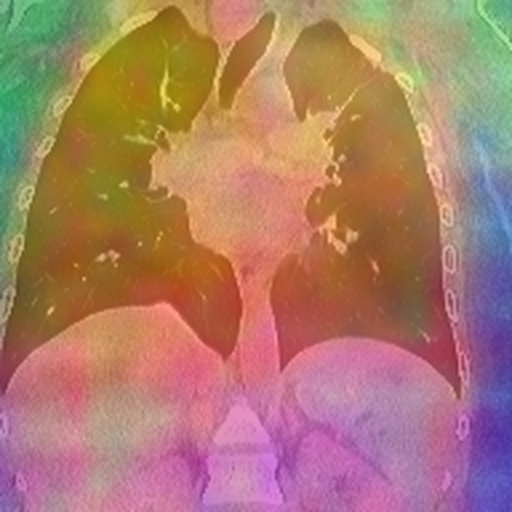} &
        \includegraphics[width=\imgwidth]{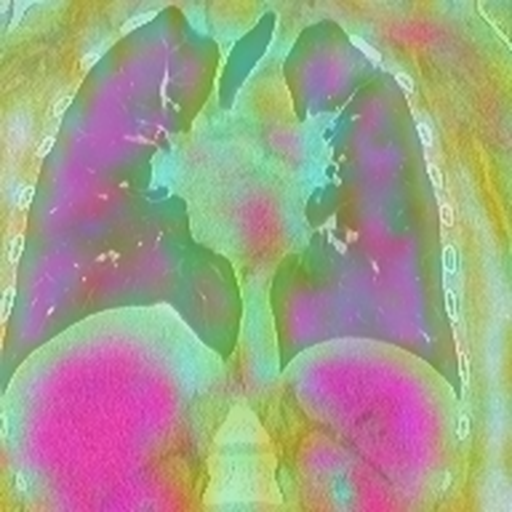} &
        \includegraphics[width=\imgwidth]{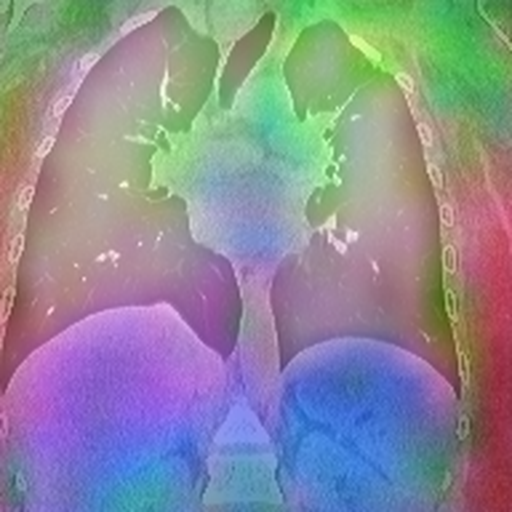} &
        \includegraphics[width=\imgwidth]{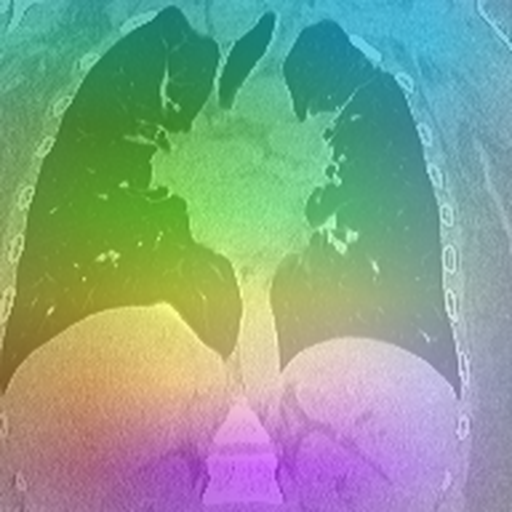} \\

        \includegraphics[width=\imgwidth]{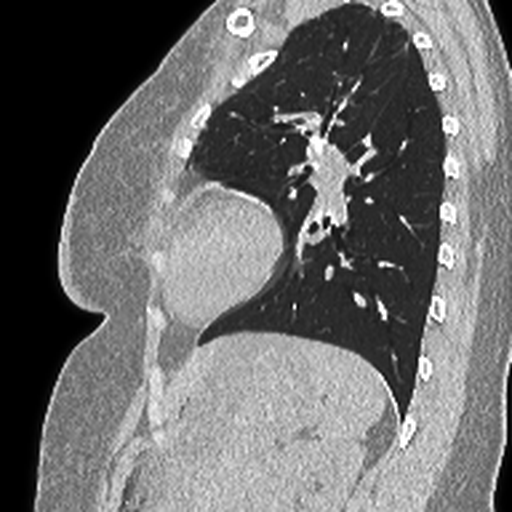} &
        \includegraphics[width=\imgwidth]{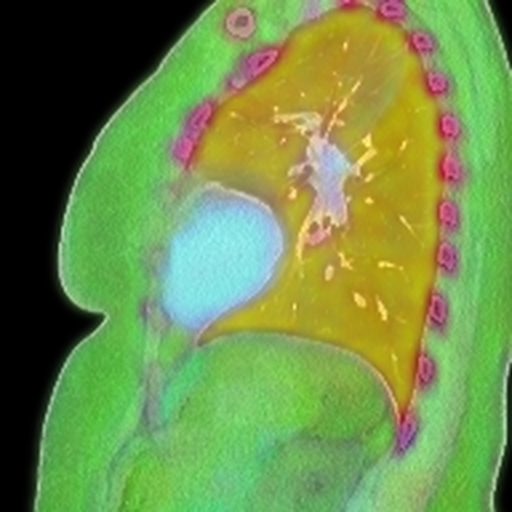} &
        \includegraphics[width=\imgwidth]{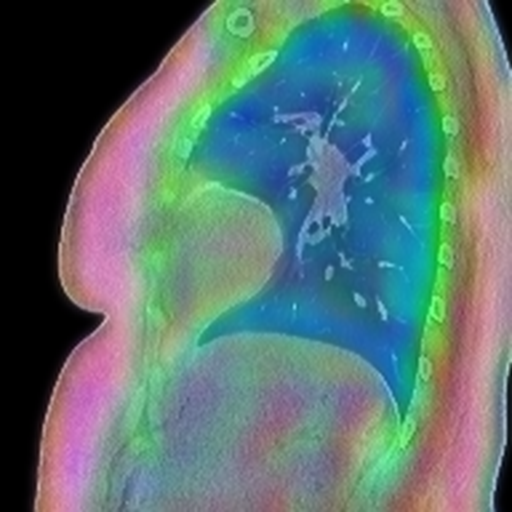} &
        \includegraphics[width=\imgwidth]{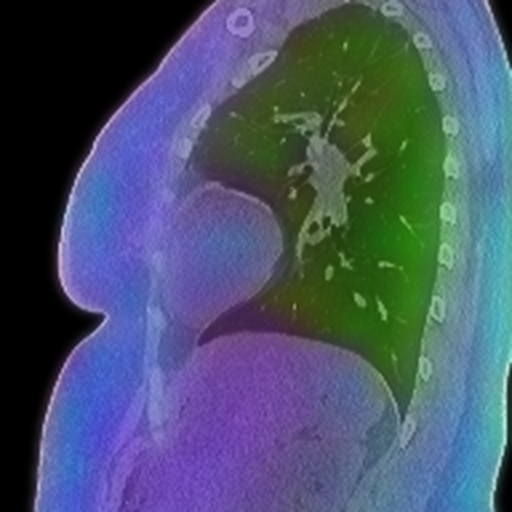} &
        \includegraphics[width=\imgwidth]{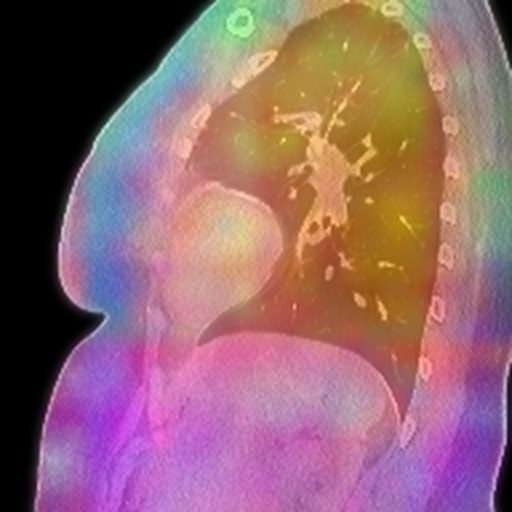} &
        \includegraphics[width=\imgwidth]{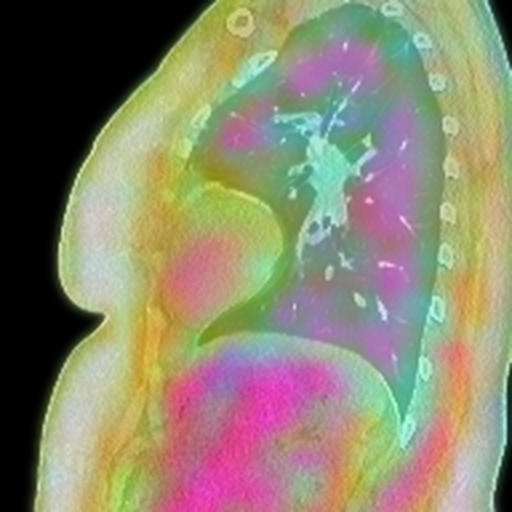} &
        \includegraphics[width=\imgwidth]{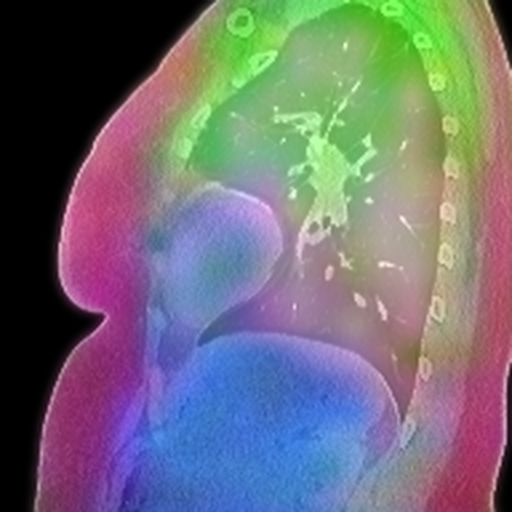} &
        \includegraphics[width=\imgwidth]{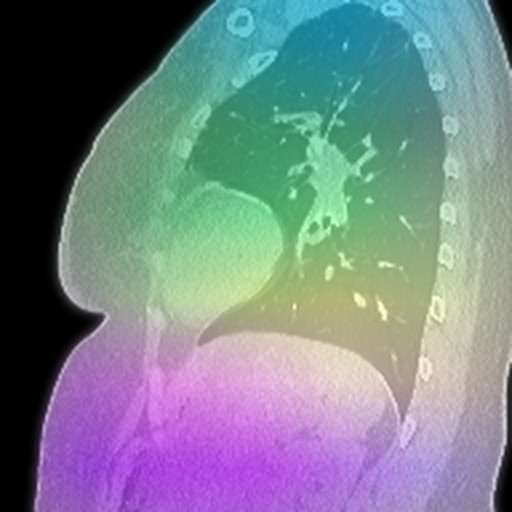} \\

    \end{tabular}
    \caption{\textbf{Principal component analysis of the dense features, chest.} Same recipe and layout as \cref{fig:pca} on two chest volumes of \ctrate (axial, coronal and sagittal slices; CT shown with a blended soft-tissue / lung window); \COLIPRI, trained on chest CT only, takes the place of \CAPTOR. Colours are not comparable across cells; only the spatial structure is.}
    \label{fig:pca_chest}
\end{figure*}

\cref{fig:pca,fig:pca_chest} show the dense features of each model with a principal component analysis.
In the maps of \JoliaTwo, each organ has a nearly uniform colour with a sharp edge at its boundary, as for the kidneys, the liver, the spine and the heart.
The maps of CAPTOR (COLIPRI on the chest volumes) and Merlin vary smoothly across the volume and separate few structures, and those of \Jolia and Pillar-0 follow edges and textures more than organs.
These are a few selected volumes, not a measure.

\section{Ablations}
\label{app:ablations}

\subsection{Setup of the ablation runs}
\label{app:ablation_setup}
The ablation runs of \cref{sec:ablations} use crops of $192^3$ voxels at 1.5\,mm and 8 GPUs, with a batch size per GPU of 12 in the first two phases, 6 for \DenseBT and 4 for patch-level segmentation, and they have no $\mathcal{L}_{\text{OSL}}$. The last three rows of \cref{tab:ablation} add $\mathcal{L}_{\text{OSL}}$, then 1\,mm volumes, then the final cooldown of the learning rate.
Each run has 40k steps per objective: 40k for most runs with one objective, including \MAE and \MAEBT on the backbones other than \HieraCT, 80k for \Conquer with \MAE or \MAEBT, 120k for \Conquer with \iBOT and \DenseBT, and 160k for \Method.
Longer runs change the AUC little: from 40k to 105k steps on \HieraCT, \Conquer goes from 84.88 to 84.30, \MAE from 74.13 to 73.50 and \MAEBT from 78.57 to 79.52.

\paragraph{Components alone.}
Added alone to \Conquer (84.88 AUC), patch-level segmentation reaches 86.59, \OneMM 86.46 and finding-label supervision 85.27 (\cref{tab:ablation_detail_findings}).
These gains partly overlap: patch-level segmentation adds 1.71 AUC alone but 0.74 after query grounding, since both learn from the masks.
On the Dice, patch-level segmentation alone raises \Conquer from 61.82 to 62.26, but adds 2.2 after query grounding (\cref{tab:ablation_detail_seg,tab:ablation}).
\OneMM alone on \Conquer raises the AUC by 1.6 and the Dice by 3.7, close to its gains on top of the masks (+1.5 and +4.3).

\subsection{Pretraining methods and backbones}
\label{app:ablation_detail}
\cref{tab:ablation_detail_findings,tab:ablation_detail_seg} give, for each run, the AUC on each test set and the Dice on each of the 13 segmentation datasets, with their means.
They cover the pretraining methods on \HieraCT (\cref{sec:res_pretraining}), the five backbones (\cref{sec:res_pareto}) and the runs of the incremental ablation (\cref{sec:res_ladder}).
\begin{table}[!htbp]
  \centering
  \footnotesize
  \caption{\textbf{Finding classification of the ablation runs} (\cref{sec:res_pretraining,sec:res_pareto,sec:res_ladder}): AUC (\%) on each test set and mean. Bold: best per column for each backbone.}
  \label{tab:ablation_detail_findings}
  \setlength{\tabcolsep}{2pt}
  \begin{adjustbox}{max width=\linewidth}   %
  \begin{tabular}{@{}lccccc@{}}
    \toprule
    Pretraining & \extabdo & Stanf. & \extchest & CT-RATE & Mean \\
    \midrule
    \multicolumn{6}{@{}l}{\textit{\HieraCT}} \\
    \DINOvTwo & 62.11 & 66.71 & 77.47 & 70.32 & 69.15 \\
    \MAE & 66.78 & 72.55 & 80.01 & 77.19 & 74.13 \\
    \iBOT & 57.74 & 60.04 & 78.00 & 65.17 & 65.24 \\
    \IJEPA & 68.65 & 74.27 & 82.80 & 79.72 & 76.36 \\
    \MAEBT & 71.06 & 77.97 & 85.04 & 80.19 & 78.57 \\
    \iBOT + \DenseBT & 68.09 & 73.24 & 82.73 & 77.59 & 75.41 \\
    \DenseBT & 64.88 & 72.19 & 82.34 & 77.07 & 74.12 \\
    \VoCo & 61.93 & 67.77 & 78.64 & 74.95 & 70.82 \\
    \Franca & 61.86 & 67.40 & 78.00 & 72.38 & 69.91 \\
    \addlinespace[3pt]
    \CLIP & 76.69 & 82.38 & 89.46 & 86.61 & 83.78 \\
    \Conquer & 78.59 & 84.02 & 90.24 & 86.68 & 84.88 \\
    \Conquer + \MAE + finding labels & 71.77 & 80.21 & 88.02 & 84.49 & 81.12 \\
    \Conquer + \MAEBT + finding labels & 72.41 & 80.53 & 88.13 & 84.15 & 81.31 \\
    \Conquer + \iBOT + \DenseBT + finding labels & 79.59 & 84.94 & 90.06 & 87.13 & 85.43 \\
    \addlinespace[3pt]
    \Conquer + \OneMM & 80.65 & 86.56 & 90.97 & 87.67 & 86.46 \\
    \Conquer + patch-level segmentation & 82.34 & 86.34 & 90.79 & 86.88 & 86.59 \\
    \addlinespace[3pt]
    \Conquer + finding labels & 79.26 & 84.47 & 90.99 & 86.38 & 85.27 \\
    \Conquer + \Coverage + finding labels & 81.13 & 85.66 & 90.43 & 87.06 & 86.07 \\
    \Conquer + \Coverage + \SSL + finding labels & 81.80 & 86.19 & 91.41 & 87.35 & 86.68 \\
    \Method & 83.56 & 86.55 & 92.23 & 87.33 & 87.42 \\
    \Method + \osl & 82.61 & 86.71 & 91.95 & 86.86 & 87.03 \\
    \quad + \OneMM & 85.14 & 87.89 & 92.34 & 88.55 & 88.48 \\
    \quad + cooldown & \textbf{85.65} & \textbf{88.25} & \textbf{92.85} & \textbf{88.68} & \textbf{88.86} \\
    \midrule
    \multicolumn{6}{@{}l}{\textit{Base Hiera}} \\
    \Conquer & 77.47 & 82.86 & 88.34 & 83.94 & 83.15 \\
    \midrule
    \multicolumn{6}{@{}l}{\textit{ResEnc-L}} \\
    \MAE & 63.86 & 68.20 & 81.09 & 75.39 & 72.14 \\
    \MAEBT & 64.11 & 71.41 & 81.23 & 76.34 & 73.27 \\
    \CLIP & 77.53 & 82.01 & 89.27 & 86.12 & 83.73 \\
    \Conquer & 79.45 & 83.99 & 89.94 & 86.25 & 84.91 \\
    \Method & \textbf{82.89} & \textbf{86.00} & \textbf{90.68} & \textbf{87.73} & \textbf{86.83} \\
    \midrule
    \multicolumn{6}{@{}l}{\textit{ResNet}} \\
    \MAE & 62.67 & 69.25 & 80.21 & 75.27 & 71.85 \\
    \MAEBT & 63.42 & 71.78 & 81.72 & 76.09 & 73.25 \\
    \CLIP & 75.82 & 81.21 & 88.58 & 85.45 & 82.77 \\
    \Conquer & 77.03 & 82.65 & 88.82 & 85.35 & 83.46 \\
    \Method & \textbf{79.86} & \textbf{84.94} & \textbf{90.26} & \textbf{85.99} & \textbf{85.26} \\
    \midrule
    \multicolumn{6}{@{}l}{\textit{Atlas}} \\
    \MAE & 66.76 & 71.63 & 80.50 & 75.65 & 73.64 \\
    \MAEBT & 67.58 & 74.38 & 81.82 & 77.27 & 75.26 \\
    \CLIP & 76.07 & 81.67 & 88.51 & 86.32 & 83.14 \\
    \Conquer & 77.51 & 82.96 & \textbf{89.53} & \textbf{86.89} & 84.22 \\
    \Method & \textbf{77.87} & \textbf{84.18} & 89.07 & 86.80 & \textbf{84.48} \\
    \midrule
    \multicolumn{6}{@{}l}{\textit{Swin}} \\
    \MAE & 66.94 & 74.04 & 82.53 & 77.38 & 75.22 \\
    \MAEBT & 69.25 & 74.95 & 83.75 & 78.43 & 76.60 \\
    \CLIP & 73.73 & 80.32 & 86.32 & 82.16 & 80.63 \\
    \Conquer & 74.73 & 81.50 & 87.46 & 82.43 & 81.53 \\
    \Method & \textbf{78.38} & \textbf{83.80} & \textbf{89.10} & \textbf{84.52} & \textbf{83.95} \\
    \bottomrule
  \end{tabular}
  \end{adjustbox}
\end{table}

\begin{table*}[!htbp]
  \centering
  \scriptsize
  \caption{\textbf{Segmentation of the ablation runs} (\cref{sec:res_pretraining,sec:res_pareto,sec:res_ladder}): Dice (\%) on each of the 13 datasets of \cref{tab:dense} (lesion class only for the lesion datasets, frozen encoder) and mean. Bold: best per column for each backbone; one decimal: older evaluation.}
  \label{tab:ablation_detail_seg}
  \setlength{\tabcolsep}{2.6pt}
  \begin{adjustbox}{max width=\linewidth}   %
  \begin{tabular}{@{}lcccccccccccccc@{}}
    \toprule
    Pretraining & MSD & AMOS & TS & Lung & Colon & Liv.\ v. & KiTS & CRLM & WAW & TS liv. & LIDC & ImCAS & SegTHOR & Mean \\
    \midrule
    \multicolumn{15}{@{}l}{\textit{\HieraCT}} \\
    \DINOvTwo & 46.88 & 80.21 & 77.10 & 56.75 & 32.04 & 47.37 & 69.28 & 50.94 & 49.68 & 61.94 & 47.95 & 40.25 & 81.51 & 57.07 \\
    \MAE & 54.77 & 85.24 & 83.66 & 68.33 & 47.03 & 56.55 & \textbf{79.68} & 56.16 & 67.92 & 66.93 & 49.49 & 41.17 & 85.65 & 64.81 \\
    \iBOT & 55.91 & 80.05 & 75.53 & 62.67 & 31.42 & 51.78 & 63.94 & 50.79 & 53.26 & 60.95 & 44.60 & 40.61 & 82.95 & 58.04 \\
    \IJEPA & 52.65 & 82.39 & 80.42 & 59.85 & 22.76 & 51.43 & 71.96 & 55.78 & 49.37 & 63.39 & 47.46 & 48.85 & 83.30 & 59.20 \\
    \MAEBT & 51.66 & 85.80 & 86.09 & 68.10 & 46.62 & 51.18 & 79.50 & 52.25 & 61.95 & 69.28 & 47.00 & 42.95 & 85.68 & 63.70 \\
    \iBOT + \DenseBT & 54.89 & 84.29 & 81.91 & 64.10 & 35.40 & 53.13 & 69.81 & 52.19 & 52.87 & 64.17 & 46.90 & 42.64 & 85.65 & 60.61 \\
    \DenseBT & 50.31 & 84.14 & 84.84 & 62.76 & 33.99 & 51.99 & 69.96 & 48.84 & 58.41 & 64.78 & 47.34 & 37.73 & 84.76 & 59.99 \\
    \VoCo & 54.84 & 84.23 & 83.04 & 64.53 & 34.34 & 53.16 & 73.48 & 53.59 & 54.84 & 65.99 & 50.75 & 35.92 & 84.26 & 61.00 \\
    \Franca & 54.96 & 82.55 & 79.21 & 60.85 & 30.46 & 51.60 & 69.53 & 51.77 & 54.56 & 60.64 & 48.17 & 37.88 & 83.07 & 58.87 \\
    \addlinespace[3pt]
    \CLIP & 53.25 & 82.68 & 82.19 & 71.65 & 34.49 & 53.64 & 67.32 & 50.39 & 52.58 & 62.85 & 50.75 & 39.11 & 84.17 & 60.39 \\
    \Conquer & 51.02 & 83.66 & 81.78 & 72.87 & 37.14 & 53.14 & 63.26 & 54.15 & 58.32 & 65.31 & 52.00 & 45.86 & 85.14 & 61.82 \\
    \Conquer + \MAE + finding labels & 48.82 & 83.35 & 83.20 & 66.66 & 37.58 & 46.14 & 72.35 & 49.53 & 53.58 & 65.37 & 48.74 & 32.14 & 84.62 & 59.39 \\
    \Conquer + \MAEBT + finding labels & 49.72 & 83.76 & 84.01 & 63.81 & 39.42 & 57.04 & 70.25 & 53.46 & 60.33 & 64.57 & 53.77 & 36.38 & 84.98 & 61.65 \\
    \Conquer + \iBOT + \DenseBT + finding labels & 50.16 & 83.54 & 83.78 & 71.92 & 36.06 & 50.17 & 67.84 & 51.21 & 53.39 & 65.51 & 47.03 & 47.84 & 84.58 & 61.00 \\
    \addlinespace[3pt]
    \Conquer + \OneMM & 52.26 & 85.13 & 82.74 & 71.88 & 45.69 & 53.92 & 72.01 & 54.38 & 65.84 & 70.42 & 59.94 & 52.70 & 84.84 & 65.52 \\
    \Conquer + patch-level segmentation & 55.13 & 85.42 & 84.90 & 67.83 & 36.95 & 56.11 & 72.60 & 48.81 & 65.30 & 67.25 & 48.41 & 35.77 & 84.96 & 62.26 \\
    \addlinespace[3pt]
    \Conquer + finding labels & 51.36 & 82.35 & 83.62 & 65.33 & 42.01 & 49.37 & 69.71 & 51.01 & 51.51 & 65.36 & 47.36 & 47.84 & 84.88 & 60.90 \\
    \Conquer + \Coverage + finding labels & 53.76 & 84.61 & 84.25 & 68.53 & 40.61 & 49.53 & 71.66 & 48.30 & 60.28 & 67.67 & 51.43 & 46.61 & 85.09 & 62.49 \\
    \Conquer + \Coverage + \SSL + finding labels & 54.25 & 85.08 & 85.05 & 70.13 & 41.93 & 50.16 & 71.35 & 48.83 & 59.48 & 68.07 & 46.03 & 43.27 & 85.71 & 62.26 \\
    \Method & 54.90 & 84.82 & 86.31 & 66.24 & 43.20 & 52.44 & 75.28 & 50.27 & 67.48 & 68.62 & 49.47 & 53.83 & 85.45 & 64.49 \\
    \Method + \osl & 56.71 & 85.74 & \textbf{86.42} & 68.46 & 47.75 & 55.30 & 73.29 & 51.63 & 64.44 & 69.48 & 51.45 & 50.67 & 85.64 & 65.15 \\
    \quad + \OneMM & \textbf{60.34} & 86.41 & 86.15 & \textbf{74.66} & 48.25 & \textbf{59.77} & 77.37 & \textbf{57.36} & 72.75 & 73.75 & \textbf{60.17} & 59.27 & \textbf{86.02} & 69.41 \\
    \quad + cooldown & 60.13 & \textbf{86.79} & 86.32 & 74.22 & \textbf{49.12} & 59.65 & 77.76 & 56.78 & \textbf{73.71} & \textbf{73.98} & 59.82 & \textbf{60.01} & 85.96 & \textbf{69.56} \\
    \midrule
    \multicolumn{15}{@{}l}{\textit{Base Hiera}} \\
    \Conquer & 49.58 & 84.06 & 83.21 & 69.42 & 33.23 & 49.74 & 67.76 & 52.80 & 55.45 & 64.76 & 47.29 & 41.78 & 85.45 & 60.35 \\
    \midrule
    \multicolumn{15}{@{}l}{\textit{ResEnc-L}} \\
    \MAE & 51.07 & 84.27 & 82.22 & 62.81 & 32.17 & 55.23 & 73.06 & 55.89 & 62.10 & 65.45 & 46.33 & 48.36 & \textbf{85.62} & 61.89 \\
    \MAEBT & \textbf{56.82} & 83.95 & 81.40 & 64.37 & 35.02 & \textbf{55.91} & 72.67 & 53.47 & 60.18 & 65.20 & \textbf{51.30} & 47.75 & 84.89 & 62.53 \\
    \CLIP & 50.12 & 82.83 & 79.44 & 71.20 & 31.05 & 46.80 & 64.08 & 50.97 & 54.91 & 63.83 & 49.79 & 48.89 & 84.14 & 59.85 \\
    \Conquer & 49.22 & 83.10 & 78.68 & \textbf{71.22} & 37.24 & 51.62 & 66.72 & 52.63 & 55.24 & 64.14 & 49.56 & 44.56 & 84.14 & 60.62 \\
    \Method & 53.08 & \textbf{85.15} & \textbf{84.21} & 70.93 & \textbf{42.74} & 54.80 & \textbf{74.32} & \textbf{57.54} & \textbf{67.93} & \textbf{67.98} & 50.65 & \textbf{50.19} & 85.50 & \textbf{65.00} \\
    \midrule
    \multicolumn{15}{@{}l}{\textit{ResNet}} \\
    \MAE & 52.73 & 85.14 & 85.14 & 65.79 & 37.03 & \textbf{54.97} & 74.28 & 53.31 & 60.25 & 66.35 & 40.35 & 36.83 & \textbf{85.62} & 61.37 \\
    \MAEBT & 53.62 & 85.21 & 85.43 & \textbf{66.56} & 33.65 & 54.80 & 73.90 & 55.38 & 57.74 & 64.77 & 45.24 & \textbf{45.32} & 85.23 & 62.07 \\
    \CLIP & 52.96 & 83.39 & 82.38 & 60.49 & 34.84 & 53.52 & 69.43 & 51.01 & 56.20 & 63.10 & \textbf{51.17} & 34.42 & 84.58 & 59.81 \\
    \Conquer & \textbf{55.12} & 83.65 & 82.22 & 58.79 & 35.25 & 51.71 & 66.17 & 52.13 & 49.65 & 63.65 & 48.94 & 43.21 & 84.00 & 59.58 \\
    \Method & 52.14 & \textbf{85.25} & \textbf{86.40} & 61.29 & \textbf{41.21} & 52.91 & \textbf{76.89} & \textbf{55.50} & \textbf{61.31} & \textbf{68.07} & 48.81 & 44.79 & 85.51 & \textbf{63.08} \\
    \midrule
    \multicolumn{15}{@{}l}{\textit{Atlas}} \\
    \MAE & 51.95 & 84.39 & 82.14 & 65.03 & 30.41 & 54.70 & 73.56 & \textbf{56.26} & 59.84 & 62.76 & \textbf{47.11} & \textbf{51.62} & \textbf{85.26} & 61.93 \\
    \MAEBT & 54.51 & 85.06 & 84.61 & 64.22 & 33.48 & \textbf{56.66} & 73.59 & 54.16 & 61.12 & 68.13 & 47.08 & 42.91 & 85.06 & 62.35 \\
    \CLIP & 54.57 & 84.25 & 82.85 & 65.69 & 35.36 & 52.53 & 70.24 & 53.23 & 54.62 & 65.68 & 44.10 & 43.97 & 84.65 & 60.90 \\
    \Conquer & 53.07 & 84.36 & 82.91 & 65.67 & 32.47 & 53.48 & 68.7 & 54.0 & 60.7 & 64.45 & 46.58 & 36.75 & 84.70 & 60.60 \\
    \Method & \textbf{55.72} & \textbf{85.23} & \textbf{85.24} & \textbf{67.77} & \textbf{41.98} & 54.02 & \textbf{73.98} & 55.48 & \textbf{62.85} & \textbf{68.51} & 46.16 & 41.42 & 85.24 & \textbf{63.35} \\
    \midrule
    \multicolumn{15}{@{}l}{\textit{Swin}} \\
    \MAE & 51.66 & \textbf{85.05} & 83.95 & \textbf{65.98} & \textbf{40.45} & 50.45 & \textbf{76.65} & 53.91 & \textbf{57.43} & \textbf{65.69} & 45.42 & 39.15 & 85.84 & 61.66 \\
    \MAEBT & 51.21 & 84.85 & \textbf{84.49} & 65.58 & 31.48 & 52.86 & 76.21 & \textbf{54.18} & 55.58 & 64.43 & 44.76 & 51.18 & 85.60 & \textbf{61.72} \\
    \CLIP & 51.78 & 83.72 & 81.18 & 55.88 & 30.25 & 54.02 & 68.42 & 52.41 & 47.15 & 60.05 & \textbf{46.18} & 45.58 & 85.51 & 58.63 \\
    \Conquer & \textbf{54.18} & 83.19 & 79.95 & 55.62 & 26.03 & 51.56 & 69.47 & 53.03 & 50.17 & 62.53 & 44.72 & \textbf{53.28} & 85.55 & 59.18 \\
    \Method & 52.22 & 84.90 & 84.04 & 59.39 & 34.39 & \textbf{54.14} & 72.91 & 53.48 & 55.91 & 63.80 & 42.42 & 48.37 & \textbf{86.34} & 60.95 \\
    \bottomrule
  \end{tabular}
  \end{adjustbox}
\end{table*}

\subsection{Query localisation}
\label{app:localisation}
We measure the localisation of each organ query on the finest scale of the frozen encoder, with three metrics (\cref{tab:oqs_per_dataset}).
The pointing accuracy is the share of volumes where the attention maximum falls inside the organ, and the attention mass is the share of the attention on the tokens of the organ.
The attention Dice (Dice$_{\text{att}}$) is the Dice between the organ mask and as many of the most-attended voxels as the mask has voxels. It measures overlap, not a segmentation.
Without query grounding, the attention maximum of an organ query falls inside its organ in 5.2\% of cases for \Jolia.
Query grounding optimises the attention mass inside the mask, so a high rate is expected for the supervised queries, and \JoliaTwo reaches 91.5\%.
In \cref{fig:attention_maps}, the attention of \JoliaTwo fills each organ, whereas that of \Jolia is diffuse and mostly outside it.
\cref{tab:oqs_per_dataset} gives the localisation of the organ queries along the incremental ablation and per test set, and \cref{tab:oqs_per_organ} per organ.
Query grounding raises the pointing accuracy from 5.2\% to 90.9\%, with similar values on the four test sets. Self-supervision then lowers it to 88.5\% and the attention mass from 88.3\% to 82.8\%; \dts and \OneMM bring them back to 91.5\% and 86.2\%.
Per organ, the query of \JoliaTwo points inside its organ in at least 85.9\% of the volumes, except for the prostate (72.3\%) and the brain (55.6\%), which TotalSegmentator finds in 657 and 54 test volumes. Among the other organs, the attention Dice is lowest on thin structures, such as the common carotid arteries (27.7) and the adrenal glands (64.7).
\cref{fig:query_top5} gives the concentration of the attention of each query, including queries without a mask.

\begin{table*}[!htbp]
\centering
\footnotesize
\setlength{\tabcolsep}{3.5pt}
\begin{adjustbox}{max width=\linewidth}   %
\begin{tabular}{@{}l ccc | ccc | ccc | ccc | ccc@{}}
\toprule
 & \multicolumn{3}{c|}{Stanford} & \multicolumn{3}{c|}{\extabdo} & \multicolumn{3}{c|}{CT-RATE} & \multicolumn{3}{c|}{\extchest} & \multicolumn{3}{c}{Mean} \\
\cmidrule(lr){2-4} \cmidrule(lr){5-7} \cmidrule(lr){8-10} \cmidrule(lr){11-13} \cmidrule(lr){14-16}
Model & Point. & Mass & Dice$_{\text{att}}$ & Point. & Mass & Dice$_{\text{att}}$ & Point. & Mass & Dice$_{\text{att}}$ & Point. & Mass & Dice$_{\text{att}}$ & Point. & Mass & Dice$_{\text{att}}$ \\
\midrule
\Jolia~\cite{jolia} & 5.8 & 4.6 & 4.2 & 5.0 & 4.4 & 5.1 & 4.6 & 4.7 & 4.0 & 5.3 & 4.3 & 4.3 & 5.2 & 4.5 & 4.4 \\
\Conquer on \HieraCT & 5.8 & 5.0 & 4.7 & 6.3 & 5.8 & 4.9 & 4.7 & 5.8 & 3.3 & 4.0 & 5.3 & 4.1 & 5.2 & 5.4 & 4.2 \\
\quad + \Coverage & 88.6 & \textbf{86.1} & 70.0 & 93.0 & \textbf{89.5} & 73.4 & 89.7 & \textbf{87.3} & 72.0 & 92.1 & \textbf{90.2} & 72.7 & 90.9 & \textbf{88.3} & 72.0 \\
\quad + \SSL & 86.8 & 80.3 & 65.9 & 90.6 & 82.8 & 68.5 & 87.1 & 83.3 & 68.3 & 89.6 & 84.9 & 68.5 & 88.5 & 82.8 & 67.8 \\
\quad + \dts & 88.0 & 84.0 & 70.2 & 92.4 & 88.1 & 74.1 & 89.3 & 85.9 & 72.2 & 91.2 & 89.0 & 73.5 & 90.2 & 86.8 & 72.5 \\
\JoliaTwo & \textbf{90.2} & 83.6 & \textbf{74.8} & \textbf{93.8} & 87.5 & \textbf{77.5} & \textbf{89.8} & 85.4 & \textbf{74.6} & \textbf{92.4} & 88.2 & \textbf{77.1} & \textbf{91.5} & 86.2 & \textbf{76.0} \\
\bottomrule
\end{tabular}
\end{adjustbox}
\caption{\textbf{Organ-query localisation along the ablation, per test set.} Frozen encoder, finest scale, no training on top; 48 organ queries with a mask, mean over organs of the per-organ mean over volumes; 5{,}125 Stanford, 791 \extabdo, 1{,}455 CT-RATE and 3{,}727 \extchest test volumes (of the 5{,}137, 811, 1{,}564 and 3{,}851 of \cref{tab:datasets}). Point.: the attention maximum falls inside the organ (1.3\,\% for a uniform map). Mass: share of the attention on the tokens of the organ. Dice$_{\text{att}}$: attention Dice, the Dice between the organ mask and as many most-attended voxels as the mask has voxels; it measures overlap, not a segmentation. \Conquer on \HieraCT: \Conquer trained by us on \HieraCT, to which the next rows add one component each. \JoliaTwo also has $\mathcal{L}_{\text{OSL}}$, 1\,mm volumes and the cooldown.}
\label{tab:oqs_per_dataset}
\end{table*}

\begin{table*}[!htbp]
\centering
\footnotesize
\setlength{\tabcolsep}{3pt}
\begin{adjustbox}{max width=\linewidth}   %
\begin{tabular}{@{}lrr rrr @{\hspace{14pt}} lrr rrr@{}}
\toprule
Organ & $n$ & \Jolia & \multicolumn{3}{c}{\JoliaTwo} & Organ & $n$ & \Jolia & \multicolumn{3}{c}{\JoliaTwo} \\
\cmidrule(lr){4-6} \cmidrule(lr){10-12}
 & & Dice$_{\text{att}}$ & Point. & Mass & Dice$_{\text{att}}$ &  & & Dice$_{\text{att}}$ & Point. & Mass & Dice$_{\text{att}}$ \\
\midrule
\multicolumn{6}{@{}l}{\textit{Abdominal organs}} & \multicolumn{6}{@{}l}{\textit{Vessels}} \\
\quad liver & 11020 & 19.0 & 97.6 & 96.9 & 95.5 & \quad aorta & 11042 & 1.5 & 98.8 & 95.0 & 89.8 \\
\quad spleen & 10501 & 7.4 & 94.0 & 93.1 & 89.7 & \quad inferior vena cava & 11030 & 0.2 & 99.2 & 92.8 & 85.5 \\
\quad pancreas & 9316 & 0.5 & 95.5 & 88.1 & 81.4 & \quad superior vena cava & 4786 & 0.0 & 95.4 & 93.6 & 84.3 \\
\quad kidneys & 9254 & 14.1 & 96.1 & 96.0 & 89.6 & \quad portal vein & 9588 & 0.2 & 94.2 & 85.1 & 71.4 \\
\quad adrenal glands & 9729 & 0.0 & 94.6 & 87.3 & 64.7 & \quad splenic vein & 9588 & 0.1 & 94.3 & 85.2 & 71.4 \\
\quad gallbladder & 7292 & 0.0 & 90.6 & 85.0 & 77.5 & \quad hepatic vein & 11020 & 9.1 & 97.7 & 96.8 & 95.4 \\
\quad biliary & 11021 & 33.2 & 97.7 & 96.9 & 93.4 & \quad pulmonary vein & 5063 & 0.1 & 98.5 & 89.8 & 76.7 \\
\quad bladder & 5234 & 41.0 & 89.1 & 83.4 & 80.5 & \quad iliac artery & 6368 & 0.0 & 98.8 & 90.5 & 71.9 \\
\quad prostate & 657 & 0.1 & 72.3 & 61.3 & 55.6 & \quad iliac vein & 6355 & 0.0 & 98.8 & 90.4 & 80.2 \\
\quad \textit{mean} & & \textit{12.8} & \textit{91.9} & \textit{87.5} & \textit{80.9} & \quad common carotid artery & 2947 & 0.1 & 91.7 & 84.4 & 27.7 \\
\multicolumn{6}{@{}l}{\textit{Digestive tract}} & \quad subclavian artery & 2937 & 0.0 & 99.6 & 92.1 & 72.4 \\
\quad esophagus & 6074 & 0.1 & 97.3 & 91.3 & 78.9 & \quad brachiocephalic trunk & 2971 & 0.0 & 98.6 & 93.6 & 80.6 \\
\quad stomach & 10952 & 10.5 & 95.9 & 87.8 & 87.5 & \quad brachiocephalic vein & 3135 & 0.3 & 97.3 & 91.1 & 78.9 \\
\quad duodenum & 8324 & 0.1 & 93.9 & 80.8 & 74.7 & \quad \textit{mean} & & \textit{0.9} & \textit{97.2} & \textit{90.8} & \textit{75.9} \\
\quad small bowel & 8220 & 1.1 & 85.9 & 86.7 & 79.7 & \multicolumn{6}{@{}l}{\textit{Bone and muscle}} \\
\quad colon & 9825 & 0.8 & 94.6 & 88.3 & 83.4 & \quad spine & 10993 & 3.8 & 97.9 & 96.3 & 80.2 \\
\quad large bowel & 9825 & 0.7 & 94.5 & 88.2 & 83.3 & \quad ribs & 11027 & 0.5 & 97.1 & 94.7 & 67.6 \\
\quad gastrointestinal tract & 11060 & 8.1 & 95.8 & 93.9 & 87.8 & \quad sternum & 5476 & 1.3 & 98.9 & 94.8 & 84.9 \\
\quad \textit{mean} & & \textit{3.1} & \textit{94.0} & \textit{88.1} & \textit{82.2} & \quad clavicles & 2909 & 0.8 & 98.9 & 96.9 & 86.5 \\
\multicolumn{6}{@{}l}{\textit{Thorax and neck}} & \quad scapulae & 3219 & 0.5 & 97.2 & 93.6 & 78.2 \\
\quad heart & 5776 & 2.3 & 97.8 & 95.8 & 91.7 & \quad hip & 6378 & 0.5 & 99.0 & 96.7 & 91.2 \\
\quad atrial appendage & 4779 & 0.0 & 99.3 & 94.3 & 79.7 & \quad sacrum & 6311 & 0.0 & 98.2 & 94.3 & 84.6 \\
\quad lungs & 9326 & 7.8 & 95.7 & 94.0 & 84.8 & \quad femurs & 4429 & 0.1 & 95.3 & 94.8 & 85.9 \\
\quad pleura & 9326 & 2.3 & 95.8 & 93.9 & 84.9 & \quad bone & 11098 & 14.5 & 99.0 & 96.9 & 79.0 \\
\quad mediastinum & 6169 & 9.0 & 93.6 & 94.0 & 88.1 & \quad muscle & 9717 & 8.5 & 96.2 & 93.9 & 87.4 \\
\quad trachea & 4688 & 0.1 & 99.1 & 94.0 & 82.6 & \quad \textit{mean} & & \textit{3.1} & \textit{97.8} & \textit{95.3} & \textit{82.6} \\
\quad thyroid & 2599 & 0.0 & 97.3 & 91.8 & 76.3 & \multicolumn{6}{@{}l}{\textit{Nervous system}} \\
\quad \textit{mean} & & \textit{3.1} & \textit{96.9} & \textit{94.0} & \textit{84.0} & \quad brain & 54 & 35.9 & 55.6 & 65.0 & 63.0 \\
 &  &  &  &  &  & \quad spinal cord & 10328 & 0.0 & 96.1 & 93.3 & 84.0 \\
 &  &  &  &  &  & \quad \textit{mean} & & \textit{18.0} & \textit{75.8} & \textit{79.1} & \textit{73.5} \\
\bottomrule
\end{tabular}
\end{adjustbox}
\caption{\textbf{Organ-query localisation per organ}, \Jolia~\cite{jolia} against \JoliaTwo, pooled over the four test sets, with the mean of each category. $n$: volumes containing the organ. Setting and metrics as in \cref{tab:oqs_per_dataset}. \Jolia is far below on every organ, so only its Dice$_{\text{att}}$ is shown. The table scores 48 of the 55 organs mapped to TotalSegmentator classes (\cref{app:coverage}). The organs supervised through a neighbouring structure, such as the hepatic vein and the large bowel, are scored against {the mask of that structure}.}
\label{tab:oqs_per_organ}
\end{table*}

\providecommand\qtNameMapped{Supervised by a mask}
\providecommand\qtNameUnmapped{Never supervised}
\providecommand\qtMeanMapped{0.42}\providecommand\qtMeanMappedTwo{0.92}       %
\providecommand\qtMeanUnmapped{0.29}\providecommand\qtMeanUnmappedTwo{0.48}
\definecolor{qtJolia}{HTML}{1F4E8C}\definecolor{qtMapped}{HTML}{C4501F}
\pgfplotstableread[col sep=tab, comment chars={\#}]{figures/data/query_top5_mapped.tsv}\qtM
\pgfplotstableread[col sep=tab, comment chars={\#}]{figures/data/query_top5_unmapped.tsv}\qtU
\providecommand\qtdot[1]{\tikz[baseline=-0.6ex]\fill[#1] (0,0) circle (1.5pt);}
\providecommand\qtrows[1]{%
  \addplot[qtMapped!45, line width=1.1pt, quiver={u=\thisrow{joliatwo}-\thisrow{jolia}, v=0, every arrow/.append style={-}},
    restrict expr to domain={\thisrow{up}}{1:1}, unbounded coords=discard] table[x=jolia, y=y]{#1};
  \addplot[qtJolia!45, line width=1.1pt, quiver={u=\thisrow{joliatwo}-\thisrow{jolia}, v=0, every arrow/.append style={-}},
    restrict expr to domain={\thisrow{up}}{0:0}, unbounded coords=discard] table[x=jolia, y=y]{#1};
  \addplot[only marks, mark=*, mark size=1.3pt, qtJolia, mark options={draw=white, line width=0.2pt}] table[x=jolia, y=y]{#1};
  \addplot[only marks, mark=*, mark size=1.5pt, qtMapped, mark options={draw=white, line width=0.2pt}] table[x=joliatwo, y=y]{#1};}
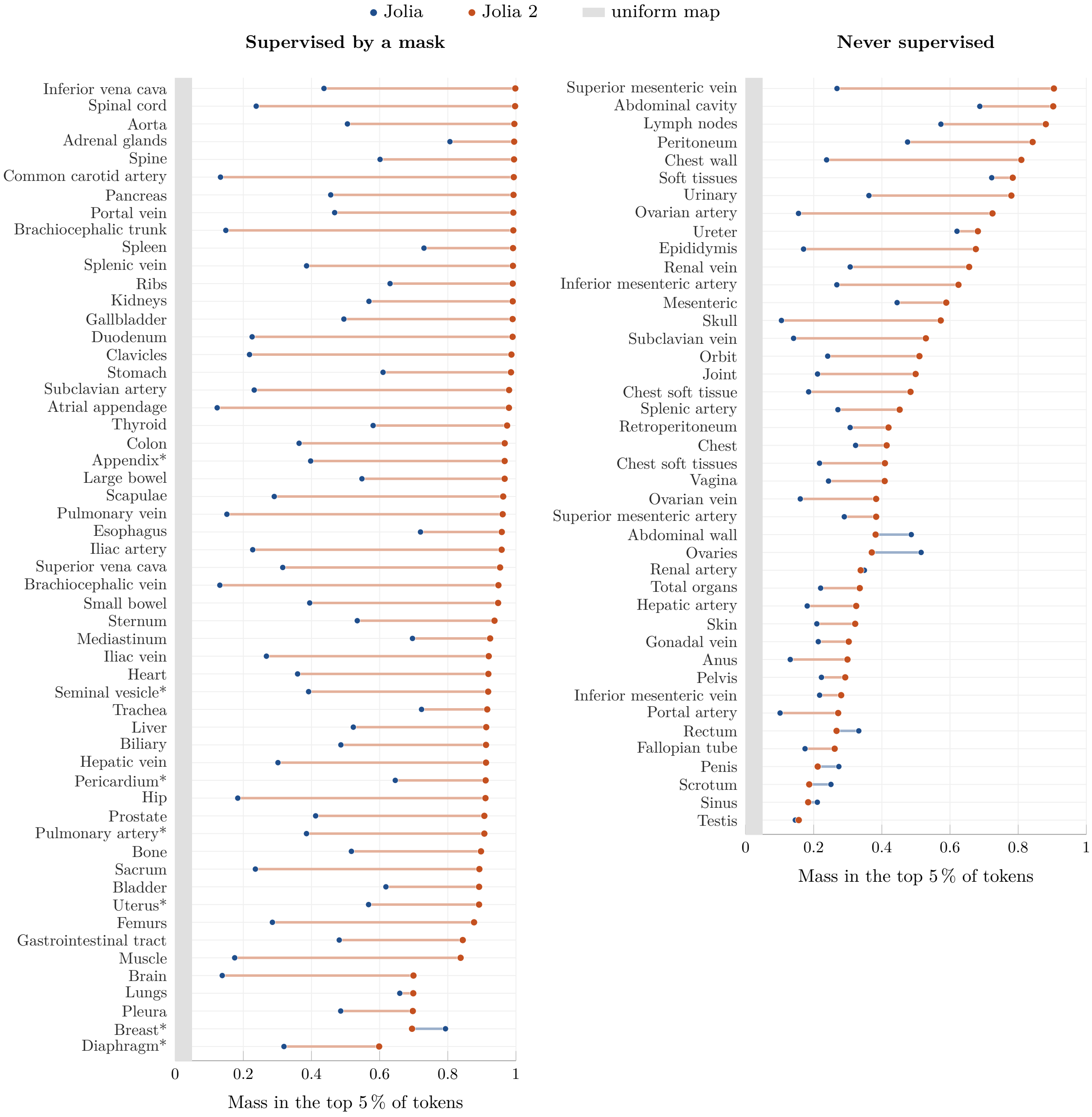
\begin{figure*}[!htbp]
  \centering
  {\footnotesize\qtdot{qtJolia}~\Jolia\qquad\qtdot{qtMapped}~\JoliaTwo\qquad\tikz[baseline=-0.6ex]\fill[black!12] (0,-0.6ex) rectangle (1.2em,0.6ex);~uniform map}\\[2pt]
  \begin{tikzpicture}
    \begin{groupplot}[group style={group size=2 by 1, horizontal sep=3.55cm},
        scale only axis, width=0.31\textwidth, height=15.2cm, xmin=0, xmax=1, xtick={0,0.2,...,1},
        xticklabel style={/pgf/number format/.cd, fixed, precision=1}, tick label style={font=\scriptsize},
        y tick label style={font=\scriptsize, text=black!85}, ytick style={draw=none},
        xlabel={Mass in the top 5\,\% of tokens}, xlabel style={font=\footnotesize},
        title style={font=\footnotesize, align=center, yshift=0.5ex},
        axis x line*=bottom, axis line style={black!35}, tick style={black!35},
        xmajorgrids, ymajorgrids, major grid style={black!6}, clip=false]
      \nextgroupplot[title={\textbf{\qtNameMapped}},
          ymin=-0.8, ymax=54.6, ytick={0,...,54}, yticklabels from table={\qtM}{name}, axis y line*=left, y axis line style={draw=none}]
        \fill[black!12] (axis cs:0,-0.8) rectangle (axis cs:0.05,54.6);
        \qtrows{\qtM}
      \nextgroupplot[title={\textbf{\qtNameUnmapped}}, height=11.7cm, yshift=1.75cm,
          ymin=-0.8, ymax=41.6, ytick={0,...,41}, yticklabels from table={\qtU}{name}, axis y line*=left, y axis line style={draw=none}]
        \fill[black!12] (axis cs:0,-0.8) rectangle (axis cs:0.05,41.6);
        \qtrows{\qtU}
    \end{groupplot}
  \end{tikzpicture}
  \caption{\textbf{How concentrated the organ-query attention is.}
    For each query, share of its attention held by the 5\,\% most-attended tokens, averaged over the test volumes of the four sets; a uniform map gives 0.05 (grey band). Rows are sorted by the value of \JoliaTwo.
    Left: queries supervised by a mask in training ($^{\ast}$: mask of a neighbouring structure, \cref{app:coverage}); right: queries never supervised (finding-type queries left out).
    Each segment joins \Jolia to \JoliaTwo; a blue segment marks a query whose concentration drops.
    A concentrated map is not necessarily well placed (\cref{fig:attention_maps}).}
  \label{fig:query_top5}
\end{figure*}

\subsection{Analysis of query grounding}
\label{app:mask_supervision}
These experiments complete \cref{sec:res_mask}. Their runs stop at 30k steps, compared with 40k for the runs of \cref{tab:ablation} without self-supervision and up to 180k for the others (\cref{app:ablation_setup}), so their values differ from those of \cref{tab:ablation}: without query grounding the AUC is 84.26, and with the organ masks of the 42 organs it is 86.08.
In \cref{fig:cov_nb_organs}, the supervised organs form nested sets of $N=7$, 14, 28 and 42 organs.
In \cref{fig:cov_target}, the mask is replaced by its 3D bounding box, by the 2D bounding box of each axial slice, or by a sphere of the equivalent radius of the organ around one voxel drawn inside the mask, either binary or Gaussian with $\sigma$ half the radius. Two controls keep only the volume of the organ: a box of that volume with random side ratios, centred on a random voxel of the body and redrawn at each step, or placed once per organ for the whole training.
The AUC is 84.26 without query grounding, 84.36 and 84.41 with the two random boxes, 85.44 to 85.56 with a point or a 3D box, 85.84 with the box per slice and 86.08 with the mask. The Dice is 60.53 without query grounding, 59.90 and 60.12 with the random boxes, 61.08 to 61.91 with a point or a 3D box, 61.93 with the box per slice and 62.77 with the mask.
\cref{tab:cov_scale_size} varies the token scales of the loss and the size of the supervised organs. For the size, the 42 organs are split by volume into 14 small, 13 medium and 15 large organs.
The scales $s_0$, $s_1$ and the three scales together give similar gains, 1.8 to 2.1 points, with $s_0$ alone slightly ahead. \Method keeps the three scales.
Supervising only the small, medium or large organs gives 1.3 to 1.5 points, below the 1.8 of all 42 organs.
\begin{table}[!htbp]
  \centering
  \footnotesize
  \caption{\textbf{Token scales of the query grounding loss (top) and size of the supervised organs (bottom).} \HieraCT, $N{=}42$ unless stated, 30k steps. Mean linear-probe AUC (\%) over the four test sets; $\Delta$: points against no query grounding (84.26).}
  \label{tab:cov_scale_size}
  \begin{adjustbox}{max width=\linewidth}   %
  \begin{tabular}{@{}lcc@{}}
    \toprule
    & Avg & $\Delta$ \\
    \midrule
    \multicolumn{3}{@{}l}{\textit{Token scales}} \\
    $s_0$ & 86.31 & +2.05 \\
    $s_1$ & 86.15 & +1.89 \\
    $s_0$, $s_1$, $s_2$ & 86.08 & +1.82 \\
    \midrule
    \multicolumn{3}{@{}l}{\textit{Supervised organs}} \\
    Small (14) & 85.72 & +1.46 \\
    Medium (13) & 85.73 & +1.47 \\
    Large (15) & 85.55 & +1.29 \\
    All (42) & 86.08 & +1.82 \\
    \bottomrule
  \end{tabular}
  \end{adjustbox}
\end{table}

\subsection{Readout}
\label{app:readout}
\begin{figure}[!htbp]
  \centering
  \includegraphics[width=0.62\textwidth]{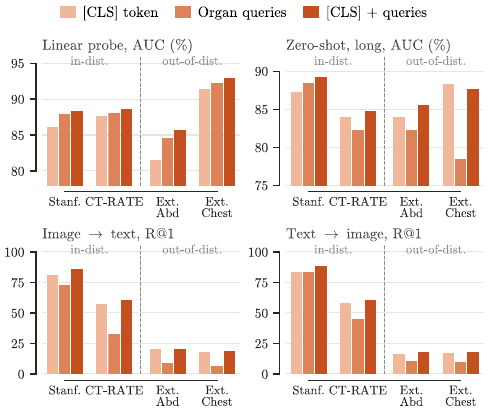}
  \caption{\textbf{Readout of \JoliaTwo.} [CLS] token, organ queries, or both (sum of the two similarities for retrieval). Zero-shot, long: prototypes built from $r=50$ reports. Retrieval: Findings section of the report, pools of $n=100$ studies, R@1.}
  \label{fig:readout}
\end{figure}

\cref{fig:readout} compares the [CLS] token, the organ queries (their organ tokens $c_k$, \cref{sec:prelim}) and both for \JoliaTwo in every evaluation, and \cref{tab:competitors,tab:transfer_readout_detail,tab:zs_summary,tab:retrieval_readout_detail} give the scores on each test set.
For finding classification, cross-center transfer and retrieval, [CLS] with the organ queries is the best readout on every test set.
In zero-shot classification, the organ queries alone are weaker on average. [CLS] alone is better with short prompts on the chest sets and on average, and both together are better with long prompts on three of the four test sets and on average. With [CLS] alone and the OOD pairs, \JoliaTwo reaches 84.3 on \extchest, compared with 82.7 for RadFinder (\cref{tab:zs_summary}).

\providecommand{\dup}[1]{{\scriptsize\textcolor{ForestGreen}{($+$#1)}}}
\providecommand{\ddown}[1]{{\scriptsize\textcolor{BrickRed}{($-$#1)}}}
\begin{table}[!htbp]
\centering
\footnotesize
\caption{\textbf{Cross-center transfer readout of \JoliaTwo.} AUC (\%, mean of 5 seeds) of a linear probe on the [CLS] token, the organ queries or both, trained on \stanford or \ctrate and tested on the same center (ID) and on \extabdo or \extchest (OOD). In parentheses: as in \cref{tab:cross_transfert}, against a probe with the same readout (\cref{tab:competitors}). Avg EXT: mean of the two OOD columns. Best readout in bold.}
\label{tab:transfer_readout_detail}
\setlength{\tabcolsep}{5pt}
\begin{adjustbox}{max width=\linewidth}   %
\begin{tabular}{ll cc cc c}
\toprule
 & & \multicolumn{2}{c}{\textbf{Abdomen}} & \multicolumn{2}{c}{\textbf{Chest}} & \\
\cmidrule(lr){3-4}\cmidrule(lr){5-6}
\textbf{Model} & \textbf{Eval method} & \stanford & \extabdo & \ctrate & \extchest & \textbf{Avg EXT} \\
\midrule
\textbf{\JoliaTwo} & [CLS] & 86.0 & 82.4 \dup{0.9} & 88.9 & 79.4 \ddown{12.0} & 80.9 \\
 & Queries & 88.1 & 84.4 \ddown{0.2} & 89.5 & 80.4 \ddown{11.8} & 82.4 \\
 & Queries + [CLS] & \textbf{89.1} & \textbf{85.5} \ddown{0.1} & \textbf{90.3} & \textbf{84.1} \ddown{8.7} & \textbf{84.8} \\
\bottomrule
\end{tabular}
\end{adjustbox}
\end{table}

\begin{table*}[!htbp]
\centering
\footnotesize
\caption{\textbf{Image--report retrieval readout of \JoliaTwo.} R@1 (\%, test) on the four test sets, scoring image--report pairs with the [CLS] token, the organ queries, or both (sum of the two cosine similarities). Findings section, mean over disjoint pools of $n=100$ studies. I$\rightarrow$T: image-to-report; T$\rightarrow$I: report-to-image. Best readout in bold. \extabdo and \extchest: external sets, unseen in pretraining.}
\label{tab:retrieval_readout_detail}
\setlength{\tabcolsep}{5pt}
\begin{adjustbox}{max width=\linewidth}   %
\begin{tabular}{ll cc cc cc cc cc}
\toprule
 & & \multicolumn{4}{c}{\textbf{Abdomen}} & \multicolumn{4}{c}{\textbf{Chest}} & \multicolumn{2}{c}{} \\
\cmidrule(lr){3-6}\cmidrule(lr){7-10}
 & & \multicolumn{2}{c}{\stanford} & \multicolumn{2}{c}{\extabdo} & \multicolumn{2}{c}{CT-RATE} & \multicolumn{2}{c}{\extchest} & \multicolumn{2}{c}{\textbf{Average}} \\
\cmidrule(lr){3-4}\cmidrule(lr){5-6}\cmidrule(lr){7-8}\cmidrule(lr){9-10}\cmidrule(lr){11-12}
\textbf{Model} & \textbf{Eval method} & I$\rightarrow$T & T$\rightarrow$I & I$\rightarrow$T & T$\rightarrow$I & I$\rightarrow$T & T$\rightarrow$I & I$\rightarrow$T & T$\rightarrow$I & I$\rightarrow$T & T$\rightarrow$I \\
\midrule
 \textbf{\JoliaTwo} & [CLS] & 81.2 & 83.3 & 20.1 & 16.3 & 57.7 & 58.2 & 18.1 & 17.4 & 44.3 & 43.8 \\
  & Queries & 73.2 & 83.5 & 8.7 & 10.3 & 33.0 & 45.0 & 6.2 & 9.6 & 30.3 & 37.1 \\
  & Queries + [CLS] & \textbf{86.3} & \textbf{88.6} & \textbf{20.3} & \textbf{18.3} & \textbf{60.7} & \textbf{60.4} & \textbf{18.7} & \textbf{17.9} & \textbf{46.5} & \textbf{46.3} \\
\bottomrule
\end{tabular}
\end{adjustbox}
\end{table*}

\makeatletter
\immediate\write\@auxout{\string\newlabel{lastfigure}{{\the\value{figure}}{}{}{}{}}}
\immediate\write\@auxout{\string\newlabel{lasttable}{{\the\value{table}}{}{}{}{}}}
\makeatother

\end{document}